\documentclass[lettersize, journal]{IEEEtran}
\usepackage{amsmath,amsfonts}
\usepackage{color}
\usepackage{booktabs} 
\usepackage{multirow} 
\usepackage{algorithmic}
\usepackage{algorithm}
\usepackage{array}
\usepackage[caption=false]{subfig}
\usepackage{bm}
\usepackage{textcomp}
\usepackage{stfloats}
\usepackage{url}
\usepackage{verbatim}
\usepackage{graphicx}
\usepackage{cite}
\usepackage{makecell}
\usepackage{upgreek}
\usepackage{hyperref}

\begin{document}

\title{Supervised Learning for Analog and RF Circuit Design: Benchmarks and Comparative Insights}


\author{%
  Asal Mehradfar$^{1}$, 
  Xuzhe Zhao$^{2}$,
  Yue Niu$^{1}$,
  Sara Babakniya$^{1}$,
  Mahdi Alesheikh$^{2}$,\\
  Hamidreza Aghasi$^{2}$, 
  Salman Avestimehr$^{1}$%
  \thanks{$^{1}$University of Southern California (USC), Los Angeles, CA 90089, USA (e-mail: mehradfa@usc.edu; yueniu@usc.edu; babakniy@usc.edu; avestime@usc.edu).}%
  \thanks{$^{2}$University of California, Irvine (UCI), Irvine, CA 92697, USA (e-mail: xuzhez3@uci.edu; maleshei@uci.edu; haghasi@uci.edu).}%
}

\maketitle

\begin{abstract}

Automating analog and radio-frequency (RF) circuit design using machine learning (ML) significantly reduces the time and effort required for parameter optimization. This study explores supervised ML-based approaches for designing circuit parameters from performance specifications across various circuit types, including homogeneous and heterogeneous designs. By evaluating diverse ML models, from neural networks like transformers to traditional methods like random forests, we identify the best-performing models for each circuit. Our results show that simpler circuits, such as low-noise amplifiers, achieve exceptional accuracy with mean relative errors as low as 0.3\% due to their linear parameter-performance relationships. In contrast, complex circuits, like power amplifiers and voltage-controlled oscillators, present challenges due to their non-linear interactions and larger design spaces. For heterogeneous circuits, our approach achieves an 88\% reduction in errors with increased training data, with the receiver achieving a mean relative error as low as 0.23\%, showcasing the scalability and accuracy of the proposed methodology. Additionally, we provide insights into model strengths, with transformers excelling in capturing non-linear mappings and k-nearest neighbors performing robustly in moderately linear parameter spaces, especially in heterogeneous circuits with larger datasets. This work establishes a foundation for extending ML-driven design automation, enabling more efficient and scalable circuit design workflows.

\end{abstract}



\section{Introduction}\label{sec:intro}

\IEEEPARstart{A}{nalog}
and radio-frequency (RF) circuit design serves as a cornerstone of modern electronic systems, enabling advancements in communication, sensing, and signal processing applications. These circuits are integral to technologies such as 5G wireless communications \cite{Hong2021Role}, automotive radar systems \cite{Dokhanchi2019AutomotiveRadar}, subsurface imaging \cite{SubsurfaceImaging}, low-cost IoT devices\cite{IoTDevices}, large-scale quantum computing\cite{Quantum} and terahertz (THz) spectroscopy \cite{THzSpectroscopy}. However, as the semiconductor industry confronts the limits of Moore’s Law \cite{IEEE97_Moore}, the design of analog circuits has become increasingly challenging. 
Unlike digital circuits, which benefit from highly automated and scalable design methodologies \cite{10031431, 869353, elfadel2019machine}, analog circuits demand extensive manual effort due to intricate trade-offs between design metrics such as power consumption, bandwidth, gain and noise\cite{gielen2000computer, Els21_Analog, razavi_RFIC}. This complexity is further exacerbated by the need to select suitable circuit topologies and optimize parameters to meet tight performance specifications, a process that often spans weeks or months and requires substantial domain expertise \cite{razavi_RFIC, ICCAD19_BagNet}. These challenges underscore the critical need for innovative approaches to streamline analog circuit design.

The application of machine learning (ML) techniques in analog circuit design has gained significant momentum in recent years. By learning the complex relationships between design specifications (e.g., power consumption, bandwidth) and circuit parameters (e.g., transistor sizes, capacitance), ML models have shown potential to reduce the exhaustive parameter sweeps required for circuit optimization \cite{ICCAD19_BagNet, DATE_AutoCkt, IEEE_Angel}.  

Early works such as BagNet \cite{ICCAD19_BagNet} and AutoCkt \cite{DATE_AutoCkt} demonstrated the potential of neural networks and reinforcement learning in synthesizing circuit parameters, reducing manual design effort. However, these methods often focus on simple, proof-of-concept circuits like single-stage amplifiers and fail to scale effectively to real-world, multi-block systems. 

More recent advancements have targeted complex analog systems. AnGeL \cite{IEEE_Angel} proposed a semi-supervised learning framework that integrates neural networks to improve data efficiency and accelerate circuit parameter optimization. Similarly, GCN-RL \cite{DAC_GCNRL} utilized graph neural networks (GNNs) to capture the topological relationships within circuits and reinforcement learning to optimize transistor sizing, enabling better generalization across circuit types. 

In addition, \cite{ICML_Analog} introduced a framework for designing analog circuits to meet threshold specifications using a combination of supervised learning and neural network-based optimization. Their approach focuses on ensuring that generated designs meet critical performance metrics, providing a practical solution for the accurate synthesis of analog circuits. 

Despite these advances, many methods remain limited to specific circuit types or lack comprehensive datasets, emphasizing the need for standardized benchmarks for robust model evaluation and scalability testing.

To address these challenges, AICircuit \cite{Mehradfar2024AICircuit} introduced a comprehensive benchmark dataset for ML-assisted analog circuit design. This dataset encompasses homogeneous circuits, such as amplifiers and mixers, as well as heterogeneous systems like transmitters and receivers, which consist of cascaded circuit blocks with distinct functionalities. AICircuit evaluated multiple ML algorithms, including traditional regressors (e.g., random forests, support vector regression) and modern deep learning models (e.g., transformers, multi-layer perceptrons), revealing their strengths and limitations in learning complex mappings between design metrics and circuit parameters.

Despite its contributions, the initial AICircuit work left several key questions unanswered. For example, while it demonstrated the feasibility of ML models for analog circuit design, it did not thoroughly analyze model-specific performance across different circuit types. Moreover, actionable insights into selecting the most suitable ML algorithm for various design scenarios were limited. To bridge these gaps, this paper builds upon AICircuit with the following key contributions:
\begin{itemize}
    \item \textbf{Exploration of ML Model Dynamics:} Comprehensive descriptions of the ML models are provided, including their architectures, training protocols, and performance nuances for specific circuit categories.
    \item \textbf{Enhanced Evaluations:} Aggregated and performance-specific comparisons are presented, supported by additional tables and plots, offering a holistic view of each model's strengths and weaknesses.
    \item \textbf{Guidelines for Algorithm Selection:} We provide observations on the suitability of different ML methods for distinct circuit types, empowering researchers and practitioners to make informed decisions when adopting ML in analog design workflows.
\end{itemize}

By extending the scope of AICircuit, this work aims to establish a more comprehensive framework for AI-driven analog and radio frequency (RF) circuit design, bridging the gap between academic research and practical applications. This work not only serves as a guide for researchers in ML-assisted circuit design but also lays the groundwork for adopting these methodologies in industry-scale analog design processes. Through this analysis, we hope to provide a roadmap for leveraging ML techniques to streamline and enhance the design process for both homogeneous and heterogeneous circuits. To enable reproducibility, all datasets and corresponding code are available at \url{https://github.com/AvestimehrResearchGroup/AICircuit}.

\section{Problem Statement}\label{sec:problem}

The design of analog and radio-frequency (RF) circuits is a challenging process that often requires iterative parameter tuning to meet desired performance specifications. Traditional approaches, such as Bayesian optimization, have been explored to optimize circuit performance by guiding the parameter search using probabilistic models that balance exploration and exploitation \cite{bayesian2020, bayesianZhang}. While Bayesian optimization can be effective for low-dimensional problems, it suffers from significant limitations when applied to circuit design. The iterative sampling process is computationally expensive, requiring substantial resources to evaluate the design space. Additionally, Bayesian optimization struggles with the complex and non-linear relationships inherent in circuit parameter spaces, making it impractical for real-world applications.

Reinforcement learning (RL) has also been explored as a means of automating circuit design by iteratively learning parameter updates through a reward system based on achieving performance targets \cite{rlzhao}. However, RL-based methods suffer from significant drawbacks: they require vast amounts of training data, are computationally intensive, and demand considerable resources to converge. Furthermore, RL's reliance on iterative exploration of the design space makes it time-consuming, particularly in circuit design tasks where data generation through simulation is expensive.

In contrast to these previous works, supervised ML algorithms offer a more efficient alternative by leveraging pre-collected datasets to learn a direct mapping from performance specifications to circuit parameters, thereby eliminating the need for iterative searches and reducing computational overhead with greater scalability.

To build upon this, we adopt a new perspective on the problem by reversing the conventional circuit design flow. Instead of iteratively searching for circuit parameters to optimize performance metrics, we directly model the relationship between performance specifications and circuit parameters. Specifically, given a desired performance vector $\bm{x}$ (e.g., power consumption, bandwidth, gain), we predict a corresponding set of circuit parameters $\bm{y}$ (e.g., resistances, capacitances, transistor widths) using a machine learning model $\mathcal{M}$. This approach bypasses the need for computationally expensive optimization techniques and allows for efficient design:
\begin{equation}
    \bm{y} = \mathcal{M}(\bm{x}),
    \label{eq:eq1}
\end{equation}

The challenge of this problem lies in the fact that there may exist multiple sets of circuit parameters $\bm{y}$ that satisfy the same performance specifications $\bm{x}$. However, in real-world scenarios, the primary goal is not to identify all possible solutions but rather to find a single set of parameters that meets the desired performance requirements. This simplifies the design process and ensures the practicality of the solution for circuit designers.

In short, our approach leverages supervised machine learning to simplify the circuit design process significantly:
\begin{itemize}
    \item \textbf{Reverse Design Flow:} ML-assisted design predicts circuit parameters from performance specifications, avoiding the need for parameter sweeps or optimization routines.
    \item \textbf{Diverse Models:} We explore various ML models, including multilayer perceptrons (MLPs), transformers, and traditional methods such as decision trees, to learn the mapping from specifications to circuit parameters.
    \item \textbf{Circuit Diversity:} We investigate the performance of these models on a range of circuits, from fundamental homogeneous blocks such as voltage amplifiers to heterogeneous systems like wireless transmitters.
\end{itemize}

By framing the problem as a direct mapping from performance to parameters, we aim to overcome the limitations of traditional optimization-based methods and provide a robust and scalable approach to analog and RF circuit design.

\subsection{Analytical ML Perspective}

Let $(x,y)\sim P(x,y)$ represent i.i.d. samples from a joint distribution over inputs and outputs. The supervised learning objective aims to solve
\begin{equation}
\theta^{*} = \arg\min_{\theta} \mathbb{E}_{(x,y)\sim P(x,y)}[\ell(f_{\theta}(x), y)],
\label{eq:supervised_obj}
\end{equation}
where $f_{\theta}(x)$ represents the ML model parameterized by $\theta$, mapping input $x$ to its predicted output, and $\ell:\mathcal{Y}\times\mathcal{Y}\to\mathbb{R}_{\geq0}$ denotes the loss function. The function $\ell$ measures the prediction error between $f_{\theta}(x)$ and the true output $y$. The presence of explicitly paired $(x,y)$ samples ensures that each observed instance contributes a low-variance gradient signal $\nabla_{\theta}\ell(f_{\theta}(x),y)$, directly guiding $f_{\theta}$ toward a function that minimizes expected error under $P(x,y)$.


In contrast, unsupervised learning optimizes criteria that depend solely on the input distribution \( P(x) \), without reference to a ground-truth target \( y \). Such criteria, often expressed as minimizing divergences (e.g., \( D(P(x) \| P_\theta(x)) \)) or optimizing surrogate objectives, do not inherently guarantee that the discovered structures align with any specific predictive task. Without a direct mapping between \( x \) and \( y \), parameter updates in unsupervised learning cannot explicitly reduce task-relevant errors, resulting in models that capture irrelevant patterns or converge to local minima that lack practical significance.

RL similarly lacks direct sample-level targets, instead seeking to maximize expected cumulative returns $\mathbb{E}_{\pi_{\theta}}\left[\sum_{t}\gamma^{t}R(s_{t},a_{t})\right]$. The resulting optimization landscape is shaped by sequential dependencies and often sparse or delayed rewards, producing high-variance gradient estimates and complicating convergence. Moreover, credit assignment across time and action sequences introduces structural complexity absent in the supervised setting.

Therefore, the supervised approach leverages a known joint distribution $P(x,y)$ and a well-defined per-sample error $\ell(f_{\theta}(x), y)$, yielding stable, directly interpretable gradient signals and tractable optimization. This fundamental alignment of the learning objective with the underlying task’s data distribution and explicit targets provides a clear analytical advantage over both unsupervised and RL frameworks.  

\subsection{Application to Circuit Schematic Design}

Consider the problem of sizing parameters in a circuit schematic. Let \(x\) represent the performance metrics (e.g., gain, bandwidth, power consumption) and \(y\) represent the design parameters (e.g., transistor dimensions, bias voltages). By performing a parametric sweep over \(y\), we obtain a dataset of pairs \(\{(x_i, y_i)\}\), effectively mapping each performance metric set to its associated parameter settings. The goal is to predict \(y\) (design parameters) given a performance specification \(x_{\text{target}}\), which naturally aligns with supervised learning.

Specifically, let \(f_\theta(x)\) represent a model that predicts the circuit’s parameter settings \(y\) given performance metrics \(x\). 
This framework leverages the known performance metrics \(x_{\text{target}}\). Each sampled \((x, y)\) pair, obtained from circuit simulations, provides a direct gradient signal that adjusts \(\theta\) to optimize configurations yielding parameters \(y\) that achieve the desired performance \(x_{\text{target}}\). The data generation process (parameter sweeps) ensures $(x,y)$ pairs reflect the circuit’s underlying physics and variability, granting a stable and interpretable mapping from performance metrics to parameters. 

In contrast, unsupervised learning would only see the inputs and outputs without a specific target. The model might find correlations or clusters in the design parameter space, but these patterns do not necessarily indicate optimal design performance. Without a target specification, identifying which configurations in \(x\)-\(y\) space correspond to the desired operational point remains ambiguous.

RL would treat circuit sizing as a sequential decision-making process, evaluating returns based on rewards. However, parameter sizing for circuit design is often a one-shot problem, where a given \(x\) (performance metrics) and \(y\) (design parameters) pair is fully observable from a single simulation, and there is no temporal sequence of actions. Introducing RL would add unnecessary complexity, requiring policy exploration, handling delayed or sparse rewards, and incurring a higher computational cost. Since the environment (circuit simulations) is static and deterministic, directly fitting a model to map \(x\) to \(y\) via supervised learning is more computationally efficient and theoretically direct.

Thus, for circuit parameter sizing with performance targets, supervised learning provides a structured, data-driven route to optimal solutions, bypassing the ambiguities of unsupervised criteria and the complexity of RL-based exploration strategies.

\section{Dataset}\label{sec:data}

This section describes the dataset collection process, which encompasses both homogeneous and heterogeneous circuits. To provide a comprehensive understanding of the data, we first define the characteristics of homogeneous and heterogeneous circuits. Subsequently, we detail the procedure used to generate the dataset for both categories. 

\subsection{Datasets for Homogeneous Circuits}
\begin{figure*}[!t]
	\centering
        \raisebox{15pt}{\subfloat[CSVA]{\includegraphics[width=0.2\textwidth]{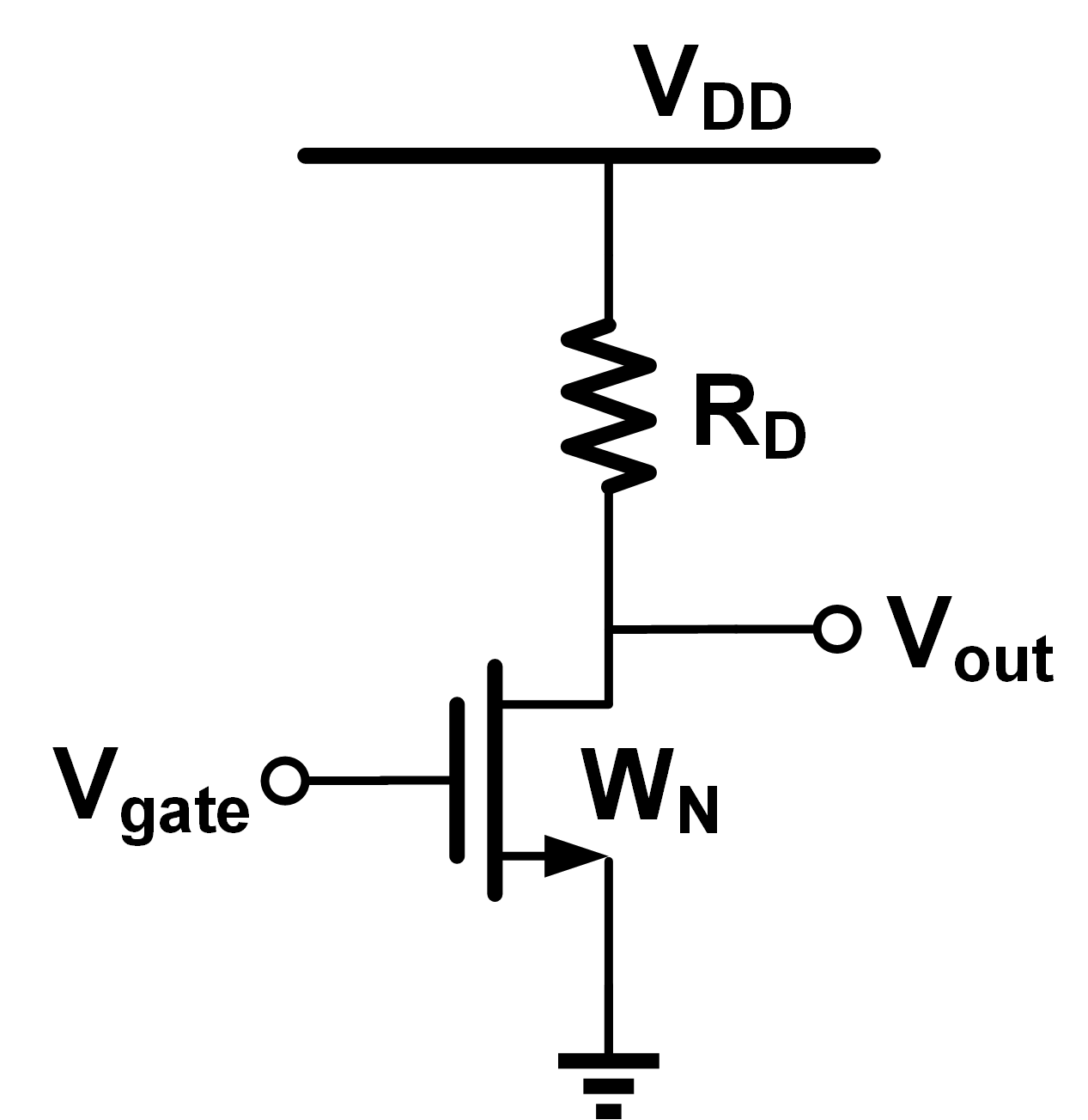}}}
        \hfil
        \raisebox{10pt}{\subfloat[CVA]{\includegraphics[width=0.21\textwidth]{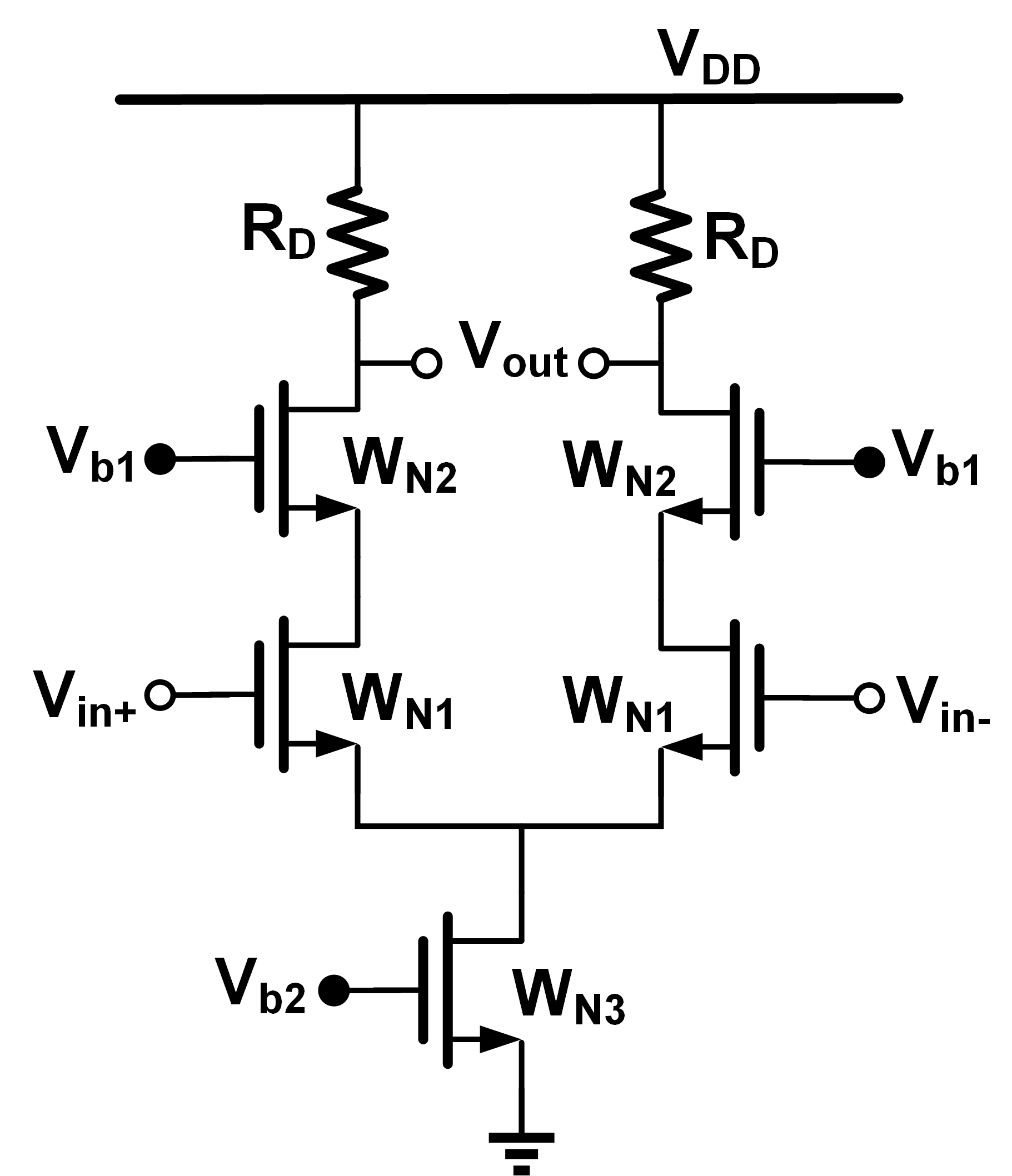}}}
        \hfil
        \subfloat[TSVA]{\includegraphics[width=0.28\textwidth]{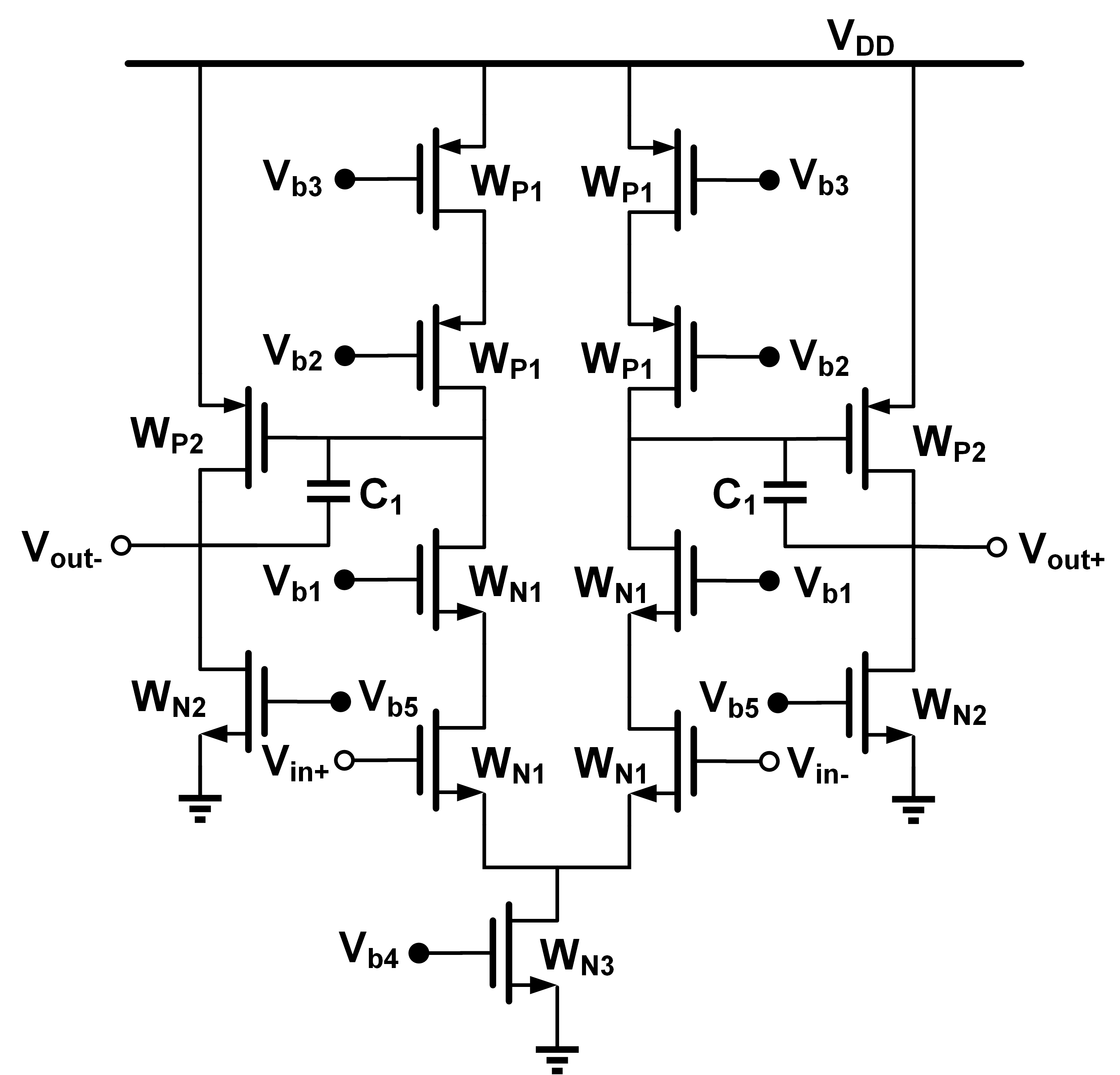}}
        \hfil
        \raisebox{10pt}{\subfloat[LNA]{\includegraphics[width=0.281\textwidth]{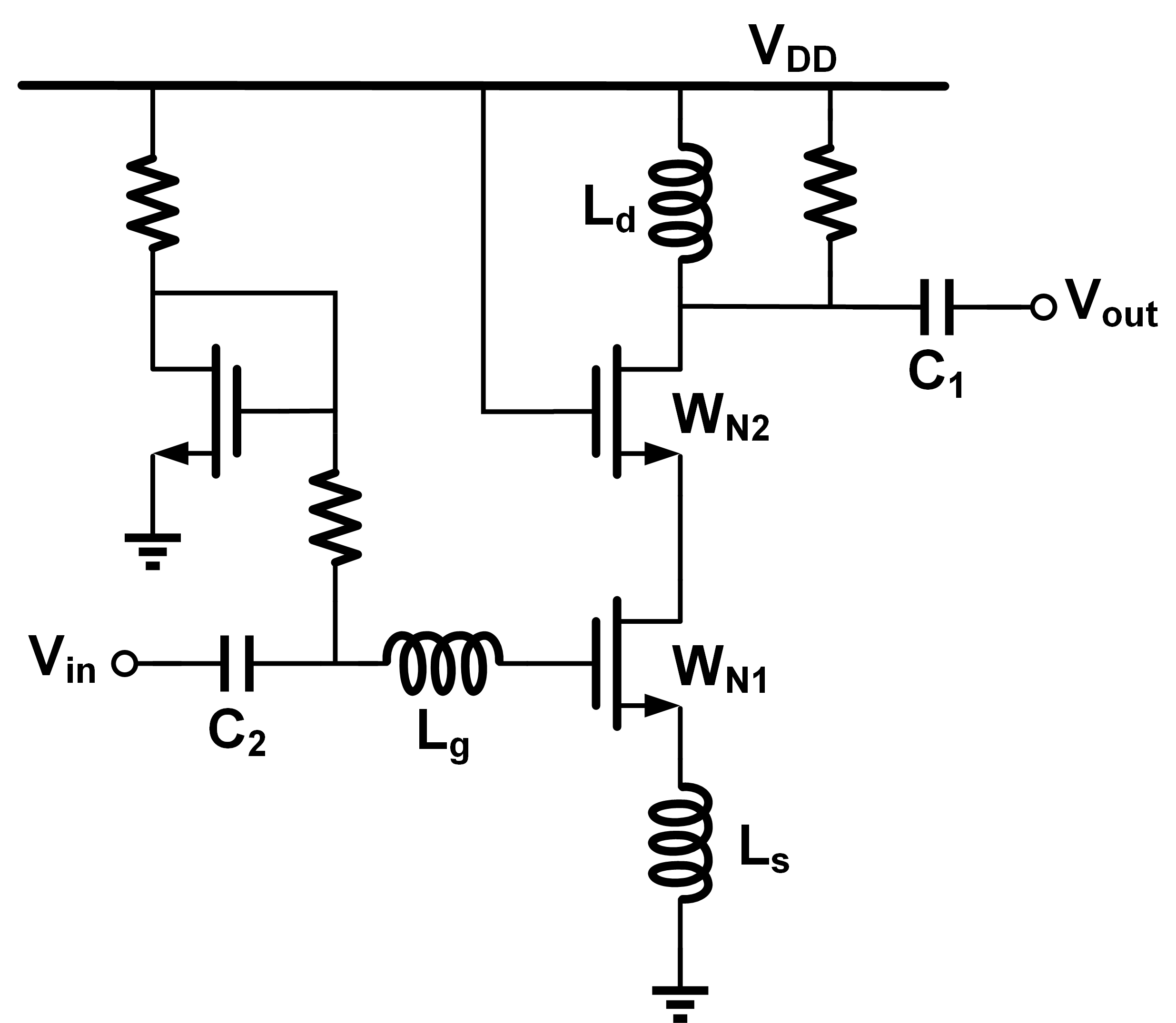}}}
        \hfil
        \subfloat[Mixer]
        {\includegraphics[width=0.26\textwidth]{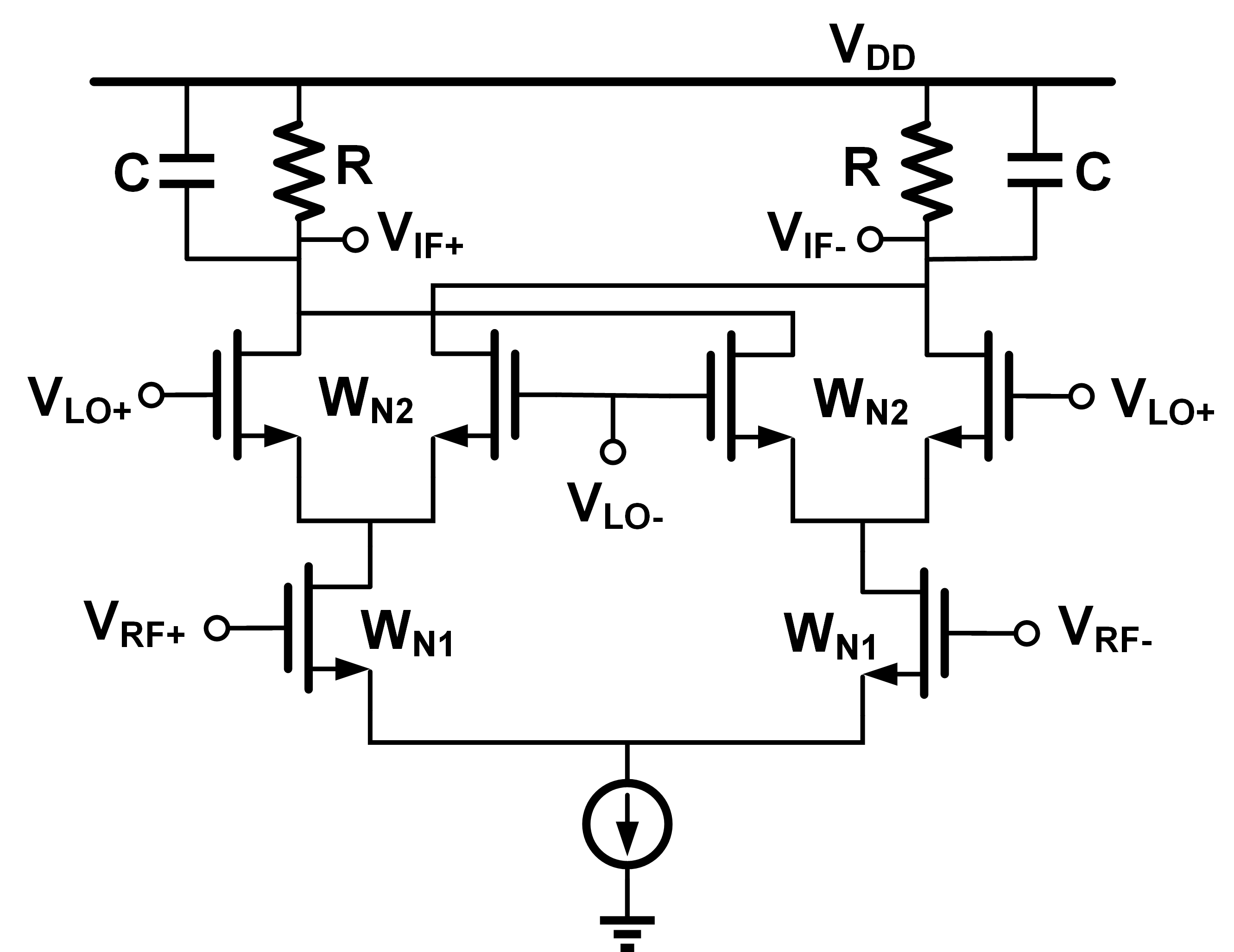}}
        \hfil
        \subfloat[VCO]
        {\includegraphics[width=0.29\textwidth]{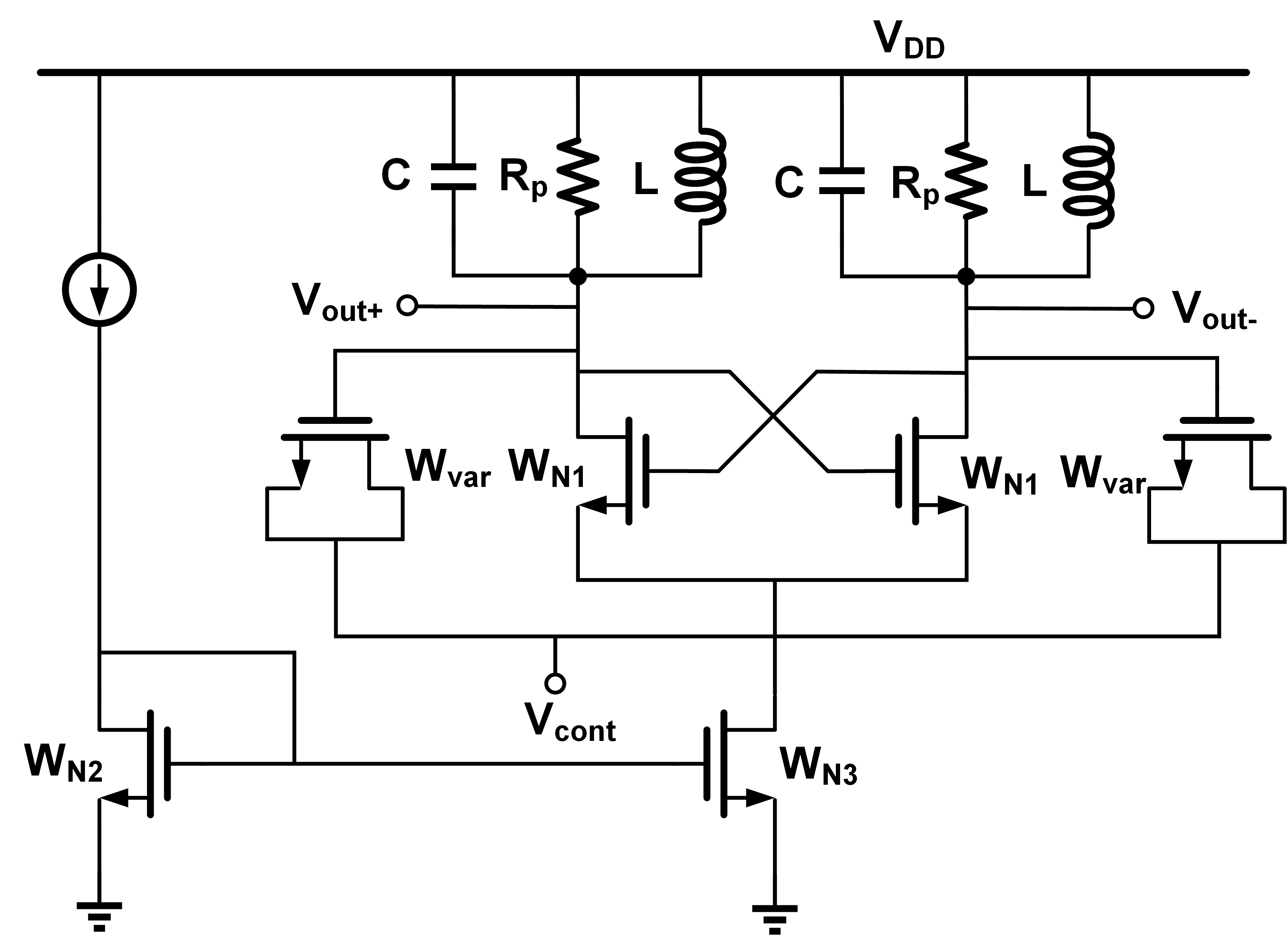}}
        \hfil
        \raisebox{27pt}{\subfloat[PA]
        {\includegraphics[width=0.45\textwidth]{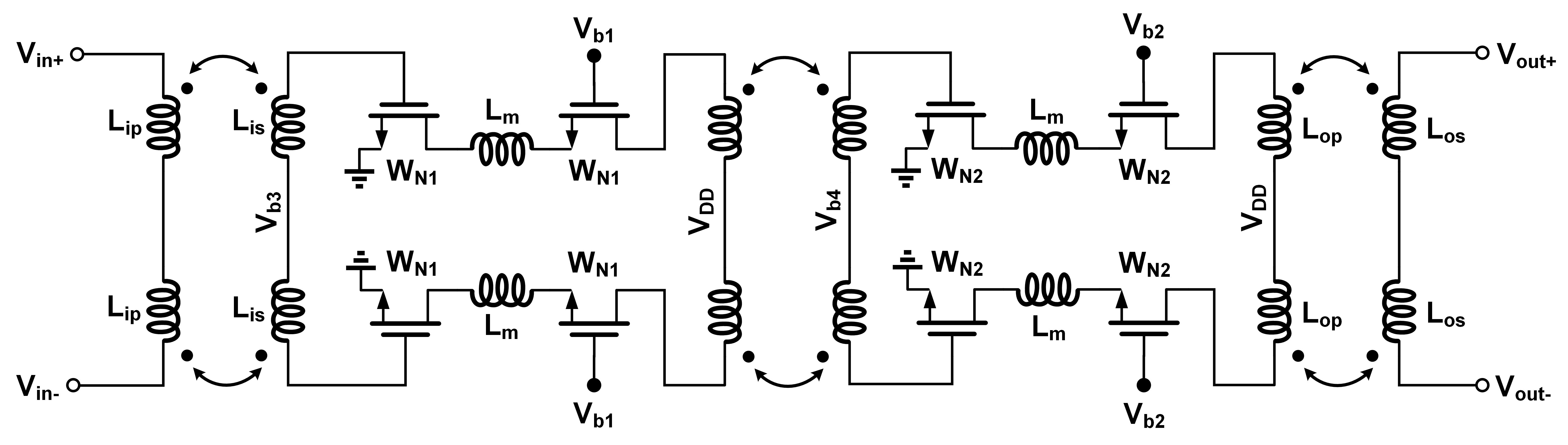}}}
        \hfil
        \caption{\centering Schematics of homogeneous circuits: (a) Common-Source Voltage Amplifier (CSVA), (b) Cascode Voltage Amplifier (CVA), (c) Two-Stage Voltage Amplifier (TSVA), (d) Low-Noise Amplifier (LNA), (e) Mixer, (f) Voltage-Controlled Oscillator (VCO), and (g) Power Amplifier (PA).}
        \label{fig:homogeneous_schematics}
\end{figure*}

Homogeneous circuits consist of a single type of circuit block, such as a common-source voltage amplifier (CSVA). These circuits are designed to perform a specific function, making them simpler in structure and easier to analyze compared to heterogeneous circuits. 
To capture a diverse range of use cases, the dataset includes seven commonly used analog and RF circuits: common-source voltage amplifier (CSVA), cascode voltage amplifier (CVA), two-stage voltage amplifier (TSVA), low-noise amplifier (LNA), mixer, voltage-controlled oscillator (VCO), and power amplifier (PA). These circuits, whose schematics are shown in Figure \ref{fig:homogeneous_schematics}, represent fundamental building blocks for complex systems. 

Among these circuits, the CSVA, CVA, and TSVA serve as essential components for voltage amplification in analog and feedback systems~\cite{razavi_design_2017}. CSVA amplifies the input signal at the gate and generates an output at the drain. CVA, which combines common-source and common-gate stages, offers enhanced gain and bandwidth suitable for high-frequency applications. TSVA improves output swings and gain by employing a two-stage amplification structure.

\begin{table}[!htb]
    \caption{Homogeneous circuits and the chosen parameters for each circuit. The sweeping range of selected design parameters is written in the form of $[\text{beg}, \text{increment}, \text{end}]$.}
    \label{tab:homogeneous_params}
    \centering 
    \resizebox{0.48\textwidth}{!}{
    \begin{tabular}{l|ccc}
        \toprule
        \textbf{Homogeneous Circuit} & \textbf{Parameter} & \textbf{Description} & \textbf{Sweeping Range} \\ \midrule
        \multirow{4}{*}{\makecell[l]{CSVA \\ \textbf{specs}: \\ dc power \textbar\ bandwidth \textbar\ gain}} & $\text{V}_{\text{DD}}$ & $\text{supply voltage}$ & $\text{1.2:0.1:1.8 (V)}$ \\
                             & $\text{V}_{\text{gate}}$ & $\text{gate voltage}$ & $\text{0.6:0.05:0.9 (V)}$\\
                             & $\text{R}_{\text{D}}$& $\text{load resistor}$ & $\text{0.5:0.1:3 (k\(\Omega\))}$\\
                             & $\text{W}_{\text{N}}$& $\text{width of nmos}$ & $\text{3:1:10 (\(\upmu\)m)}$\\ \midrule
        \multirow{4}{*}{\makecell[l]{CVA \\ \textbf{specs}: \\ dc power \textbar\ bandwidth \textbar\ gain}} & $\text{R}_{\text{D}}$& $\text{load resistor}$ & $\text{0.5:0.1:2 (k\(\Omega\))}$\\
           & $\text{W}_{\text{N1}}$& $\text{width of nmos}$ & $\text{6:1:17 (\(\upmu\)m)}$\\
           & $\text{W}_{\text{N2}}$& $\text{width of nmos}$ & $\text{5:1:12 (\(\upmu\)m)}$\\
           & $\text{W}_{\text{N3}}$& $\text{width of nmos}$ & $\text{4.5:0.5:9 (\(\upmu\)m)}$\\ \midrule
        \multirow{6}{*}{\makecell[l]{TSVA \\ \textbf{specs}: \\ dc power \textbar\ bandwidth \textbar\ gain}} & $\text{C}_{\text{1}}$& $\text{miller capacitor}$ & $\text{150:50:250 (fF)}$\\
           & $\text{W}_{\text{P1}}$& $\text{width of pmos}$ & $\text{10:1:18 (\(\upmu\)m)}$\\
           & $\text{W}_{\text{P2}}$& $\text{width of pmos}$ & $\text{7.5:5:22.5 (\(\upmu\)m)}$\\
           & $\text{W}_{\text{N1}}$& $\text{width of nmos}$ & $\text{10:1:18 (\(\upmu\)m)}$\\
           & $\text{W}_{\text{N2}}$& $\text{width of nmos}$ & $\text{7.5:5:22.5 (\(\upmu\)m)}$\\
           & $\text{W}_{\text{N3}}$& $\text{width of nmos}$ & $\text{16:2:24 (\(\upmu\)m)}$\\ \midrule
        \multirow{7}{*}{\makecell[l]{LNA \\ \textbf{specs}: \\ dc power \textbar\ bandwidth \\ \textbar\ power gain \textbar\ $\text{S}_{\text{11}}$ \textbar\ noise figure}} & $\text{C}_{\text{1}}$& $\text{output capacitor}$ & $\text{300:100:600 (fF)}$\\
           & $\text{C}_{\text{2}}$& $\text{input capacitor}$ & $\text{200:100:500 (fF)}$\\
           & $\text{L}_{\text{d}}$& $\text{drain inductor}$ & $\text{3:1:5 (nH)}$\\
           & $\text{L}_{\text{g}}$& $\text{gate inductor}$ & $\text{8.4:1:11.4 (nH)}$\\
           & $\text{L}_{\text{s}}$& $\text{source inductor}$ & $\text{0.6:0.1:0.8 (nH)}$\\
           & $\text{W}_{\text{N1}}$& $\text{width of nmos}$ & $\text{25:1.25:30 (\(\upmu\)m)}$\\ 
           & $\text{W}_{\text{N2}}$& $\text{width of nmos}$ & $\text{25:1.25:30 (\(\upmu\)m)}$\\ \midrule
        \multirow{4}{*}{\makecell[l]{Mixer \\ \textbf{specs}: \\ dc power \textbar\ voltage swing \\ \textbar\ conversion gain \textbar\ noise figure}} & $\text{C}$& $\text{coupling capacitor}$ & $\text{0.5:0.1:1.5 (pF)}$\\
           & $\text{R}$& $\text{load resistor}$ & $\text{200:25:500 (\(\Omega\))}$\\
           & $\text{W}_{\text{N1}}$& $\text{width of nmos}$ & $\text{15:1:25 (\(\upmu\)m)}$\\ 
           & $\text{W}_{\text{N2}}$& $\text{width of nmos}$ & $\text{5:1:15 (\(\upmu\)m)}$\\ \midrule
       \multirow{7}{*}{\makecell[l]{VCO \\ \textbf{specs}: \\ dc power \textbar\ frequency \\ \textbar\ phase noise \textbar\ tuning range \\ \textbar\ output power}} & $\text{C}$& $\text{capacitor in resonant tank}$& $\text{100:25:200 (fF)}$\\
           & $\text{L}$& $\text{inductor in resonant tank}$& $\text{2:1:6 (nH)}$\\
           & $\text{R}_{\text{p}}$& $\text{parallel resistor}$& $\text{1:1:4 (k\(\Omega\))}$\\
           & $\text{W}_{\text{N1}}$& $\text{width of nmos}$& $\text{24:8:56 (\(\upmu\)m)}$\\ 
           & $\text{W}_{\text{N2}}$& $\text{width of nmos}$& $\text{11:1:12 (\(\upmu\)m)}$\\ 
           & $\text{W}_{\text{N3}}$& $\text{width of nmos}$& $\text{96:32:160 (\(\upmu\)m)}$\\ 
           & $\text{W}_{\text{var}}$& $\text{width of nmos capacitor}$& $\text{75:12.5:125 (\(\upmu\)m)}$\\ \midrule
        \multirow{7}{*}{\makecell[l]{PA \\ \textbf{specs}: \\ dc power \textbar\ $\text{S}_{\text{11}}$ \textbar\ $\text{S}_{\text{22}}$  \\ \textbar\ power gain \textbar\ PAE \\ \textbar\ drain efficiency \textbar\ $\text{P}_{\text{sat}}$}} & $\text{L}_{\text{ip}}$& $\text{input primary inductor}$& $\text{175:175:525 (pH)}$\\
           & $\text{L}_{\text{is}}$& $\text{input secondary inductor}$& $\text{40:40:120 (pH)}$\\
           & $\text{L}_{\text{m}}$& $\text{inter-stage matching inductor}$& $\text{87.5:87.5:263 (pH)}$\\
           & $\text{L}_{\text{op}}$& $\text{output primary inductor}$& $\text{238:238:714 (pH)}$\\ 
           & $\text{L}_{\text{os}}$& $\text{output secondary inductor}$& $\text{30:30:90 (pH)}$\\
           & $\text{W}_{\text{N1}}$& $\text{width of nmos}$& $\text{16:3:28 (\(\upmu\)m)}$\\ 
           & $\text{W}_{\text{N2}}$& $\text{width of nmos}$& $\text{24:4:40 (\(\upmu\)m)}$\\ 
         \bottomrule
    \end{tabular}
    }
\end{table}

For RF applications, the dataset also includes LNAs, Mixers, VCOs, and PAs, all of which are essential building blocks in innovative and advancing wireless systems \cite{wireless_systems}. The LNA amplifies weak input signals while maintaining low noise across a wide bandwidth, making it indispensable in RF receiver front-ends. Mixers are responsible for frequency upconversion and downconversion in RF transmitters and receivers, respectively. VCOs generate periodic signals with tunable frequency, utilizing various topologies to provide negative resistances, e.g., the cross-coupled configuration.
PAs deliver substantial power to transmitting antennas while ensuring high efficiency. 
Table~\ref{tab:homogeneous_params} summarizes the design parameters and their sweeping ranges for homogeneous circuits.

\subsection{Datasets for Heterogeneous Circuits}

In addition to homogeneous circuits, we also investigate heterogeneous systems, which comprise multiple circuit blocks with distinct functionalities, such as a transmitter incorporating VCO and PA. These circuits are more complex, as they integrate different components to achieve multifunctional objectives. This provides a richer dataset for analyzing interactions between diverse circuits and their combined performance.

\begin{figure*}[!ht]
    \centering
    \subfloat[Transmitter]{\includegraphics[width=0.7\textwidth]{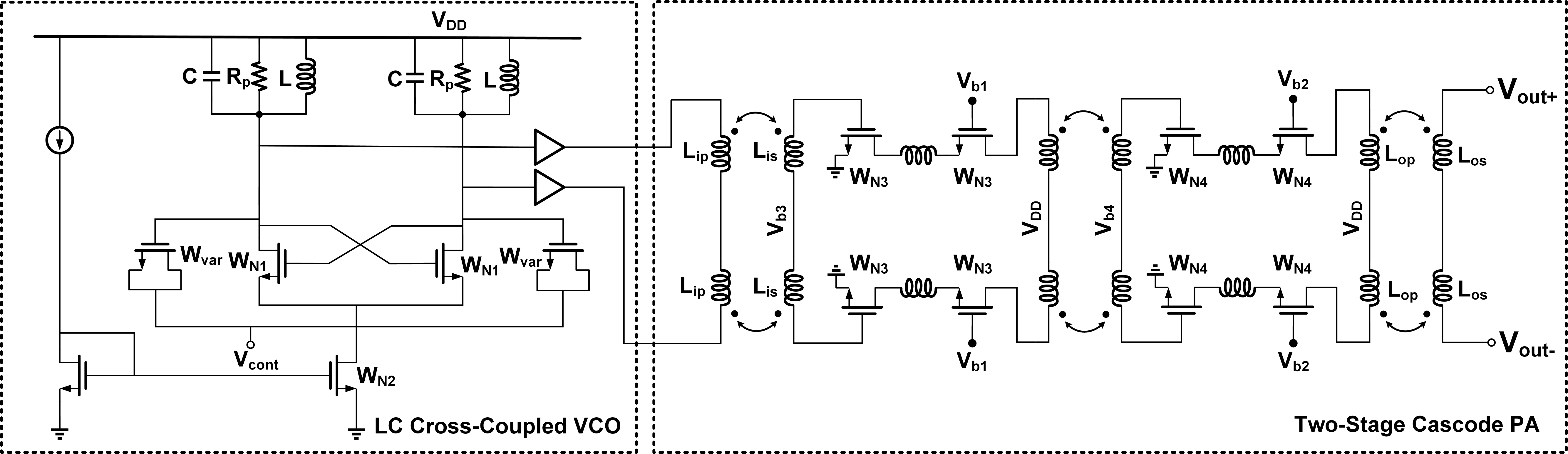} 
    \label{fig:transmitter}} \\
    \subfloat[Receiver]{\includegraphics[width=0.7\textwidth]{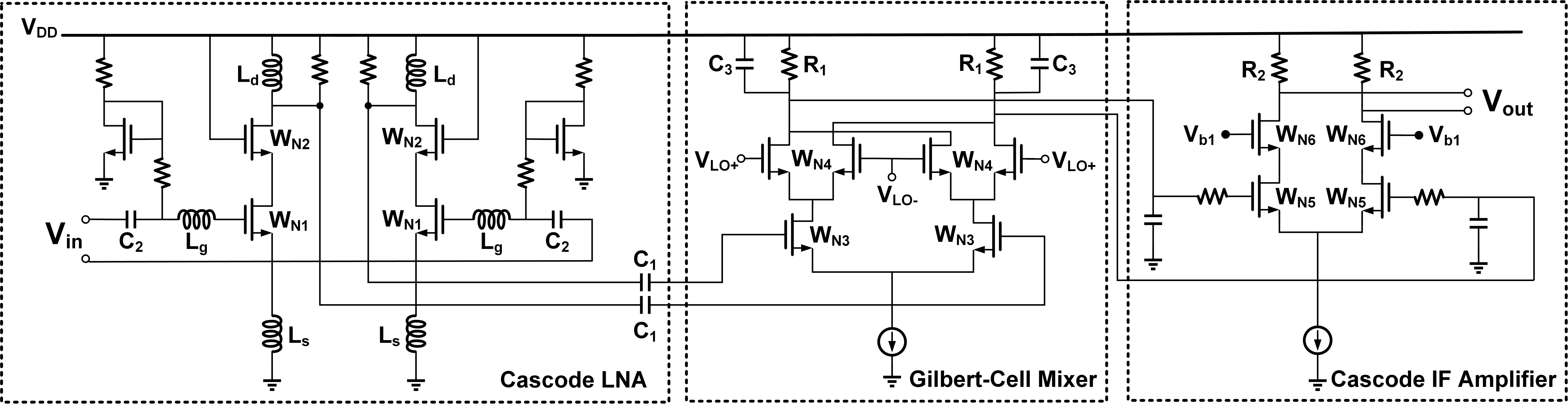}
    \label{fig:receiver}}
    \caption{\centering Schematics of the heterogeneous transceiver system: (a) Transmitter, (b) Receiver. These circuits represent key components of the transceiver chain, designed to meet diverse performance specifications across varying tasks.}
    \label{fig:transceiver_schem}
\end{figure*}

\begin{figure}[!ht]
	\centering
            \subfloat{\includegraphics[width=0.48\textwidth]{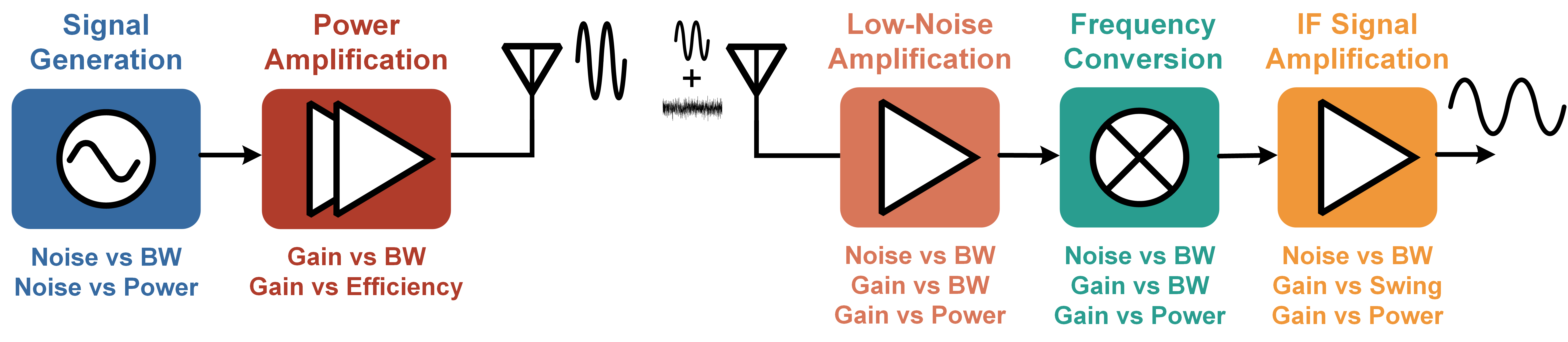}}
            \hfil
        \centering \caption{\centering Heterogeneous architecture of a 28\,GHz wireless transceiver highlighting functionality and design trade-offs for each block: The transmitter (left) includes a voltage-controlled oscillator (VCO) and power amplifier (PA). The receiver (right) consists of a low-noise amplifier (LNA), mixer, and cascode voltage amplifier (CVA).}
        \label{fig:transceiver}
\end{figure}

Specifically, we study a transmitter and a receiver system operating at 28\,GHz, commonly used in high-speed communication systems~\cite{Millimiter_wave_circuit}. These systems combine blocks such as VCOs, PAs, LNAs, Mixers, and CVAs.
In the transmitter system (Figure \ref{fig:transmitter}), the VCO generates a periodic signal with a tunable frequency governed by a control voltage, which is then amplified by the PA to deliver sufficient power for signal transmission. In the receiver system (Figure \ref{fig:receiver}), an LNA amplifies the weak input signal while minimizing noise. A Mixer converts the signal from radio frequency to intermediate frequency (IF), and a CVA provides further amplification for signal processing~\cite{razavi_RFIC, li2002multi}. Figure~\ref{fig:transceiver} illustrates the transceiver system. Furthermore, the sweeping ranges for the design parameters of heterogeneous systems are detailed in Table~\ref{tab:heterogeneous_params}.

\vspace{-4mm}

\begin{table}[!htb]
    \centering
    \caption{Heterogeneous circuits and the chosen parameters for each block. The sweeping range of selected design parameters is written in the form of $[\text{beg}, \text{increment}, \text{end}]$.}
    \label{tab:heterogeneous_params}
    \noindent\resizebox{0.48\textwidth}{!}{
       \begin{tabular}{l|l|cc}
        \toprule
         \textbf{Heterogeneous Circuits} & \textbf{Individual Block} & \textbf{Parameter} & \textbf{Sweeping Range} \\ \midrule
         \multirow{12}{*}{\makecell[l]{Transmitter System \\ \textbf{specs}: \\ dc power \textbar\ bandwidth \\ \textbar\ output power
         \textbar\ voltage swing}} & \multirow{6}{*}{\makecell[l]{Voltage-Controlled Oscillator (VCO) \\ \textbf{specs}: \\ phase noise \textbar\ tuning range}}& $\text{C}$ &  $\text{50:50:150 (fF)}$ \\
         & & $\text{L}$ & $\text{60:60:180 (pH)}$ \\ 
         & & $\text{R}_{\text{p}}$ & $\text{300:100:500 (\(\Omega\))}$ \\ 
         & & $\text{W}_{\text{N1}}$ & $\text{7.5:2.5:12.5 (\(\upmu\)m)}$ \\ 
         & & $\text{W}_{\text{N2}}$ & $\text{187.5:12.5:212.5 (\(\upmu\)m)}$ \\ 
         & & $\text{W}_{\text{var}}$ & $\text{70:10:90 (\(\upmu\)m)}$ \\ 
         \cmidrule{2-4}
         & \multirow{6}{*}{\makecell[l]{Power Amplifier (PA)  \\ \textbf{specs}: \\ power gain \textbar\ drain efficiency \textbar\ PAE}} & $\text{L}_{\text{ip}}$ & $\text{175:175:350 (pH)}$ \\
         & & $\text{L}_{\text{is}}$ & $\text{60:60:120 (pH)}$ \\ 
         & & $\text{L}_{\text{op}}$ & $\text{360:353:713 (pH)}$ \\ 
         & & $\text{L}_{\text{os}}$ & $\text{45:45:90 (pH)}$ \\
         & & $\text{W}_{\text{N3}}$ & $\text{22:5:32 (\(\upmu\)m)}$ \\ 
         & & $\text{W}_{\text{N4}}$ & $\text{16:5:26 (\(\upmu\)m)}$ \\ \midrule

         \multirow{14}{*}{\makecell[l]{Receiver System  \\ \textbf{specs}: \\ dc power \textbar\ gain \textbar\ noise figure}}& \multirow{7}{*}{\makecell[l]{Low-Noise Amplifier (LNA)  \\ \textbf{specs}: \\ power gain \textbar\ $\text{S}_{\text{11}}$ \textbar\ noise figure}}& $\text{C}_{\text{1}}$ & $\text{130:50:180 (fF)}$ \\
         & & $\text{C}_{\text{2}}$ & $\text{170:50:220 (fF)}$ \\ 
         & & $\text{L}_{\text{d}}$ & $\text{180:50:230 (pH)}$ \\ 
         & & $\text{L}_{\text{g}}$ & $\text{850:100:950 (pH)}$ \\ 
         & & $\text{L}_{\text{s}}$ & $\text{80:10:90 (pH)}$ \\ 
         & & $\text{W}_{\text{N1}}$ & $\text{20:3:26 (\(\upmu\)m)}$ \\ 
         & & $\text{W}_{\text{N2}}$ & $\text{37.5:2.5:42.5 (\(\upmu\)m)}$ \\ 
         \cmidrule{2-4}
         & \multirow{4}{*}{\makecell[l]{Mixer \\ \textbf{specs}: \\ voltage swing \textbar\ conversion gain}} & $\text{C}_\text{3}$ & $\text{1:0.1:1.1 (pF)}$ \\
         & & $\text{R}_\text{1}$ & $\text{400:100:500 (\(\Omega\))}$ \\ 
         & & $\text{W}_{\text{N3}}$ & $\text{14:2:18 (\(\upmu\)m)}$ \\ 
         & & $\text{W}_{\text{N4}}$ & $\text{6:2:10 (\(\upmu\)m)}$ \\
         \cmidrule{2-4}
         & \multirow{3}{*}{\makecell[l]{Cascode Voltage Amplifier (CVA) \\ \textbf{specs}: \\ gain}} & $\text{R}_{\text{2}}$ & $\text{300:100:400 (\(\Omega\))}$ \\
         & & $\text{W}_{\text{N5}}$ & $\text{26:2:30 (\(\upmu\)m)}$ \\ 
         & & $\text{W}_{\text{N6}}$ & $\text{14:2:18 (\(\upmu\)m)}$ \\ 
         \bottomrule
      \end{tabular}
    }
\end{table}

\subsection{Dataset Collection Procedure}
\begin{figure*}[!ht]
    \centering
    \subfloat{\includegraphics[width=\textwidth]{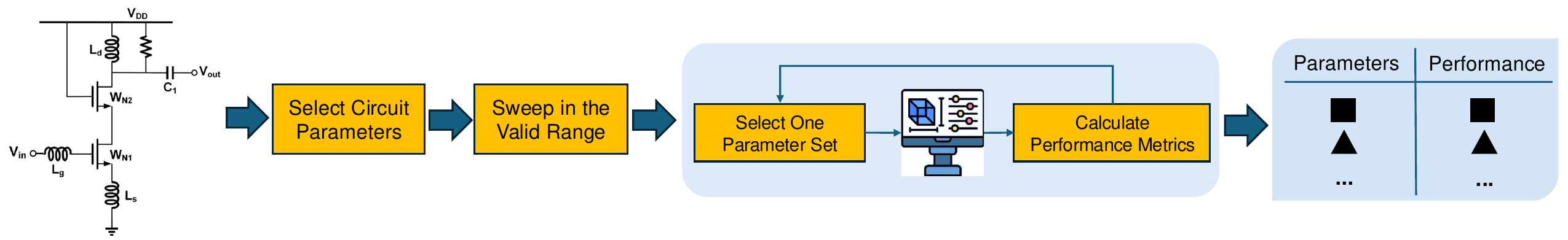}}
    \caption{\centering Procedure for creating dataset including homogeneous and heterogeneous circuits. This pipeline illustrates parameter selection, range sweeping, simulation, and performance metric computation to generate the final dataset.}
    \label{fig::dataset_proc}
\end{figure*}

To generate the dataset, we followed the procedure illustrated in Figure \ref{fig::dataset_proc}. First, a schematic for each circuit was designed using a 45 nm CMOS technology in Cadence Virtuoso~\cite{Cadence02}. Then, key design parameters were identified for each circuit, and sweeping ranges for these parameters were defined in the format $[\text{beg}, \text{increment}, \text{end}]$. For every parameter combination, simulations were performed using Cadence Virtuoso to calculate the corresponding performance metrics. Finally, the results were stored as rows in the dataset table, where each row consists of circuit parameters and their performance metrics. 

\begin{table*}[!htb]
    \caption{Dataset statistics for each circuit.}
    \label{tab:dataset_size}
    \centering
    \resizebox{0.8\textwidth}{!}{
    \begin{tabular}{l|ccccccccc}
        \toprule
        \textbf{Circuit} & \textbf{CSVA} & \textbf{CVA} & \textbf{TSVA} & \textbf{LNA} & \textbf{Mixer} & \textbf{VCO} & \textbf{PA} & \textbf{Transmitter} & \textbf{Receiver} \\ \midrule
        \multirow{1}{*}{\makecell[c]{\textbf{Dataset Size}}} & $\text{7.8k}$ & $\text{15.1k}$ & $\text{19.3k}$ & $\text{32k}$ & $\text{17.1k}$ & $\text{13.5k}$ & $\text{5.6k}$ & $\text{95.3k}$ & $\text{155.4k}$\\
        \midrule
        \textbf{Number of Circuit Parameters} & 4 & 4 & 6 & 7 & 4 & 6 & 7 & 12 & 14\\
        \midrule
        \textbf{Number of Performance Metrics} & 3 & 3 & 3 & 5 & 4 & 5 & 7 & 9 & 9\\
        \bottomrule
    \end{tabular}
    }
\end{table*} 


Table~\ref{tab:dataset_size} provides the dataset statistics for each circuit. Compared to homogeneous circuits, heterogeneous systems feature larger parameter spaces, highly non-linear interactions between individual blocks, and intricate trade-offs within the blocks and across the entire system. These complexities significantly increase the challenge for ML models to learn accurate mappings and necessitate a larger number of data points to ensure robust and reliable performance.

\section{Methodology}\label{sec:method}

\begin{figure}[!ht]
    \centering
    \subfloat{\includegraphics[width=0.48\textwidth]{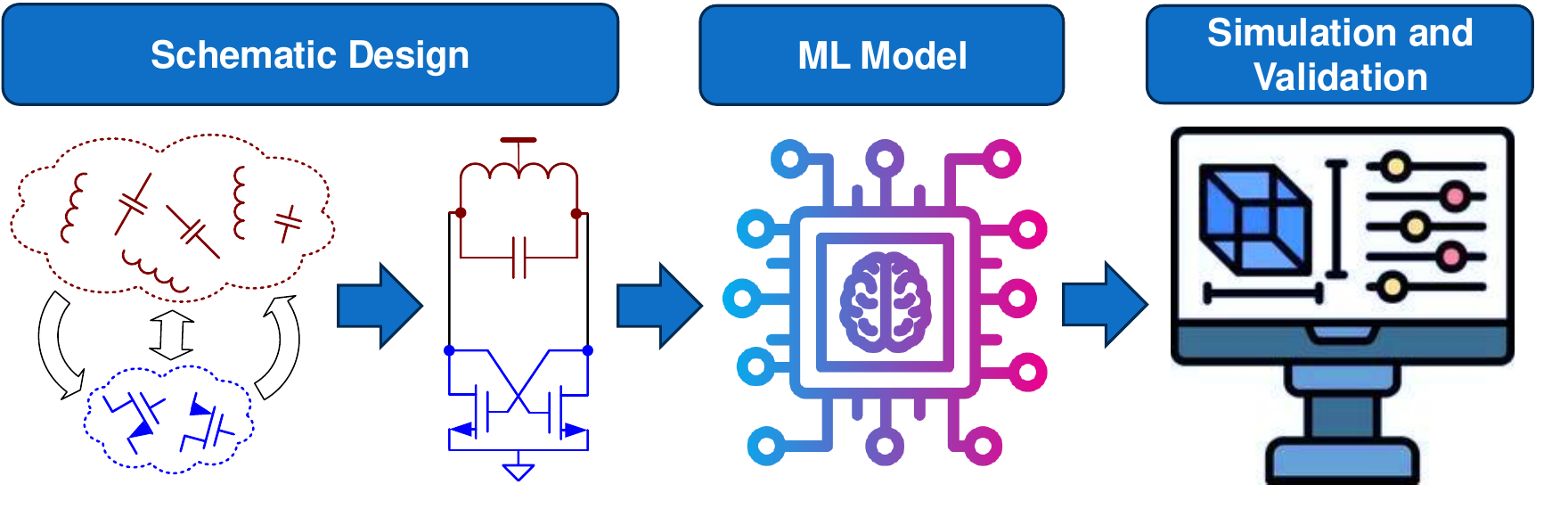}}
    \caption{\centering Pipeline of the proposed design methodology including three stages: (1) \textit{Schematic Design}, where circuit structures and key design parameters are defined; (2) \textit{ML Model}, where ML models are trained to predict circuit parameters from performance specifications; and (3) \textit{Simulation and Validation}, where the predicted parameters are validated through simulation to ensure compliance with performance targets.}
    \label{fig:pipeline}
\end{figure}

The proposed design methodology, illustrated in Figure~\ref{fig:pipeline}, consists of three primary stages. The first stage is \textbf{schematic design}, where circuit schematics are designed and key parameters are identified. In the second stage, \textbf{ML model}, supervised machine learning models are trained to map performance specifications to circuit parameters using the dataset generated from extensive simulations (Section \ref{sec:data}). Finally, in the \textbf{simulation and validation} stage, the predicted parameters are simulated to validate the model's accuracy and ensure that the performance metrics meet the desired specifications. This end-to-end pipeline significantly reduces the manual effort traditionally required in analog and RF circuit design by leveraging the predictive power of machine learning models.

In this section, we present the methodology for predicting circuit parameters from performance specifications using ML techniques. Our approach focuses on three main aspects: the choice of models, the evaluation metrics, and the end-to-end training and evaluation process. First, we describe the ML models used to learn the mapping from performance to circuit parameters, including neural networks and conventional regression methods. Next, we outline the evaluation metrics employed to assess the accuracy, robustness, and reliability of the models, ensuring a comprehensive analysis of their performance across various circuit types. Finally, we introduce the end-to-end training and simulation-based evaluation pipeline, which integrates ML predictions with circuit simulation tools to validate the performance of predicted parameters.

\subsection{Models}

In this section, we describe the ML models employed to predict circuit parameters $\bm{y}$ from input performance specifications $\bm{x}$. These models include advanced deep learning techniques and conventional regression methods chosen for their ability to capture the complex, non-linear relationships inherent in analog and RF circuit design. Specifically, we utilize the multi-layer perceptron (MLP), transformer, random forest (RF), k-nearest neighbors (kNN), and support vector regression (SVR) models. Each model approximates the mapping $\bm{y} = \mathcal{M}(\bm{x})$ (Equation \ref{eq:eq1}) with unique mathematical formulations and architectures suited to specific challenges in circuit design. The key parameters for these models are summarized in Table~\ref{tab:models_parameters}, and their detailed descriptions are provided below.

\textbf{1) Multi-Layer Perceptron (MLP):}
MLP is a feedforward neural network that approximates the mapping $\bm{y} = \mathcal{M}(\bm{x})$  by stacking multiple layers of linear transformations and non-linear activation functions:
\[
\bm{y} = \mathcal{M}(\bm{x}) = W^{(L)} z^{(L-1)} + b^{(L)},
\]
\[
z^{(l)} = \sigma(W^{(l)} z^{(l-1)} + b^{(l)}), \quad l = 1, \dots, L-1,
\]
\[
z^{(0)} = \bm{x}.
\]
Here, \(W^{(l)}\) and \(b^{(l)}\) are the weight matrices and bias vectors for layer \(l\), \(z^{(l)}\) is the activation at layer \(l\), and \(\sigma\) represents the non-linear activation function (e.g., ReLU). The depth and flexibility of MLP allow it to approximate highly non-linear relationships between $\bm{x}$ and $\bm{y}$.
MLP's strength lies in its ability to handle complex mappings, though it may be prone to overfitting with small datasets.

\textbf{2) Transformer:}
The Transformer \cite{NIPS23_Attention} leverages a self-attention mechanism to model complex dependencies between components of the performance vector \(\bm{y}\). It implements $\mathcal{M}$ as:
\[
\bm{y} = \mathcal{M}(\bm{x}) = W_{\text{out}} \text{FFN}(\text{MultiHead}(xQ, xK, xV)),
\]
where \(xQ\), \(xK\), and \(xV\) are the query, key, and value matrices derived from $\bm{x}$, and the multi-head attention mechanism is defined as:
\[
\text{MultiHead}(Q, K, V) = \text{Concat}(\text{head}_1, \dots, \text{head}_h)W^O,
\]
with each attention head computed as:
\[
\text{head}_i = \text{softmax}\left(\frac{Q_i K_i^T}{\sqrt{d_k}}\right)V_i.
\]
Here, \(\text{FFN}\) represents the feedforward layers applied after the attention mechanism \cite{arXiv19_Bert}.
The transformer's ability to capture both local and global dependencies in $\bm{x}$ makes it well-suited for complex mappings.

\textbf{3) Random Forest Regressor (RF):}
Random Forest (RF) approximates the mapping \(\bm{y} = \mathcal{M}(\bm{x})\) using an ensemble of decision trees \cite{ML01_RandomForest}:
\[
\bm{y} = \mathcal{M}(\bm{x}) = \frac{1}{N} \sum_{i=1}^N T_i(x),
\]
where \(T_i\) represents the output of the \(i\)-th decision tree trained on random subsets of data, and \(N\) is the number of trees. RF excels in modeling non-linear relationships with robustness to noise and overfitting, making it practical for circuit design.

\textbf{4) k-Nearest Neighbors Regressor (kNN):}
kNN implements \(\mathcal{M}\) by averaging the parameters of the \(k\) closest samples in the training set:
\[
\bm{y} = \mathcal{M}(\bm{x}) = \frac{1}{k} \sum_{i \in \mathcal{N}_k(x)} y_i,
\]
where \(\mathcal{N}_k(x)\) denotes the indices of the \(k\) nearest neighbors of \(\bm{x}\). This method captures localized mappings, though its computational cost scales with the dataset size \cite{kramer2013k}.

\textbf{5) Support Vector Regressor (SVR):} SVR solves a constrained optimization problem for each circuit parameter \(y^j\) to find a hyperplane that fits \(\bm{x}\) to \(y^j\) within a margin $\epsilon$, approximating \(y^j\) as $\mathbf{w}^T_j x + b_j$. Here, $\mathbf{w}_j$ and $b_j$ are computed by solving:
\[
\arg\min_{\mathbf{w}, b} \frac{1}{2} \| \mathbf{w} \|^2 + C \sum_{i=1}^N \max(0, |y^j_i - (\mathbf{w}^T x_i + b)| - \epsilon), 
\]
\noindent
where $N$ is the number of training data points, $\mathbf{w}$ and $b$ define the hyperplane, and $C$ controls the trade-off between margin width and error tolerance~\cite{awad2015support}. Using kernel functions, SVR maps \(\bm{x}\) into a higher-dimensional space, enabling it to handle non-linear mappings effectively \cite{MIT04_Kernel}. These models collectively address the challenges of mapping \(\bm{x}\) to \(\bm{y}\), leveraging their unique strengths to tackle diverse circuit complexities. However, a key limitation of SVR is that, unlike other models that jointly learn multiple outputs (multiple circuit parameters in our case), SVR does not consider dependencies between outputs, which can negatively impact accuracy. 

\begin{table*}[!htb]
    \caption{Models and the chosen parameters for each model.}
    \label{tab:models_parameters}
    \centering
    \resizebox{0.99\textwidth}{!}{
    \begin{tabular}{l|ccc}
        \toprule
        \textbf{Model} & \textbf{Parameter} & \textbf{Description} & \textbf{Value} \\ \midrule
        \multirow{6}{*}{\makecell[l]{Transformer}} & $\texttt{dim\_model}$ & $\text{first fully connected layer dim}$ & $\text{200}$ \\
                             & $\texttt{num\_heads}$ & $\text{heads in the multi-head attention models}$ & $\text{2}$\\
                             & $\texttt{dim\_hidden}$& $\text{load resistor}$ & $\text{200}$\\
                             & $\texttt{dropout\_p}$& $\text{the dropout probability}$ & $\text{0.1}$\\
                             & $\texttt{num\_encoder\_layers}$& $\text{number of layers in transformer encoder}$ & $\text{6}$\\
                             & $\texttt{activation}$& $\text{activation function of transformer encoder}$ & $\text{relu}$\\ \midrule
        \multirow{2}{*}{\makecell[l]{Multi Layer Perception (MLP)}} & $\texttt{num\_layers}$& $\text{number of fully connected layers}$ & $\text{7}$\\
           & $\texttt{dim\_layers}$& $\text{dimension of layers}$ & $\text{[200, 300, 500, 500, 300, 200]}$\\ \midrule
        \multirow{2}{*}{\makecell[l]{Support Vector Regressor (SVR)}} & $\texttt{kernel}$& $\text{kernel type of the algorithm}$ & $\text{rbf}$\\
           & $\texttt{multi\_target\_regression\_type}$& $\text{type of combining multiple SVRs}$ & $\text{MultiOutputRegression}$\\
          \midrule
        \multirow{2}{*}{\makecell[l]{k-Nearest Neighbors Regressor (KNN)}} & $\texttt{n\_neighbors}$& $\text{number of neighbors}$ & $\text{5}$\\
           & $\texttt{weights}$& $\text{weight function used in prediction}$ & $\text{uniform}$\\
            \midrule
        \multirow{2}{*}{\makecell[l]{Random Forest Regressor (RF)}} & $\texttt{n\_estimators}$& $\text{number of trees in the forest}$ & $\text{100}$\\
           & $\texttt{criterion}$& $\text{function to measure the quality of a split}$ & $\text{squared\_error ($l_2$ Loss)}$\\
         \bottomrule
    \end{tabular}
    }
\end{table*}
\subsection{Metrics}

To evaluate the performance of our machine learning models, we adopt a two-step evaluation approach. First, we use the predicted circuit parameters as inputs to the Cadence simulation tool to obtain the corresponding performance metrics. These simulated performance values, denoted as $\hat{\bm{x}}$, are compared against the desired performance metrics $\bm{x}$ specified in the dataset.

The primary evaluation metric is the \textit{relative error}, which quantifies the deviation between the predicted performance $\bm{\hat{x}_i}$ and the target performance $\bm{x_i}$:
\begin{equation}
    i\text{th Performance Error} = \frac{\text{\textbar}\bm{x}_i - \hat{\bm{x}}_i\text{\textbar}}{\bm{x}_i}.
\end{equation}

To assess the overall accuracy of the predicted parameters for a given circuit, we compute the \textit{mean relative error} across all performance metrics as: 
\begin{equation}
    \text{Mean Relative Error} = \frac{1}{N} \sum_{i=1}^N \frac{\text{\textbar}\bm{x}_i - \hat{\bm{x}}_i\text{\textbar}}{\bm{x}_i},
\end{equation}
where \( N \) is the total number of performance metrics.

This aggregated performance metric provides a single value that summarizes the accuracy of each predicted parameter set.

\paragraph{Error Distribution Analysis}
In addition to computing the mean relative error, we analyze the distribution of errors by plotting histograms of the relative errors for each model. These histograms provide insights into the variability of prediction accuracy and highlight the models' tendencies to produce outliers or consistently accurate predictions.

\paragraph{Comparison Metrics}
To compare the performance of models with each other and across different circuits, we report the following key statistical metrics derived from the error distributions:
\begin{itemize}
    \item \textbf{Mean and Standard Deviation:} The mean relative error and its standard deviation summarize the central tendency and spread of errors, respectively.
    \item \textbf{75th Percentile (P75):} This metric indicates the relative error threshold below which 75\% of all errors fall. A lower P75 value reflects better overall performance.
    \item \textbf{90th Percentile (P90):} Similar to P75, P90 reflects the threshold for 90\% of errors, providing insights into the upper error bounds.
    \item \textbf{\% of Errors Below Thresholds:} We report the percentage of errors smaller than 2\% and 5\% to measure how frequently models achieve high accuracy.
    \item \textbf{Outlier Percentage (\% $>$ 20\%):} This metric highlights the proportion of predictions with large errors (greater than 20\%), which is critical for identifying unreliable predictions.
\end{itemize}

Each of these metrics captures different aspects of the error distribution, allowing for a comprehensive comparison of model performance. For example, the mean and P75 provide insights into general accuracy, while the outlier percentage highlights model robustness in avoiding large prediction errors.

\subsection{Model Training and Simulation-Based Evaluation}

Our codebase provides a seamless end-to-end model training and evaluation pipeline, as illustrated in Algorithm~\ref{alg:training} and \ref{alg:evaluation}. This pipeline ensures smooth interaction between the ML workflow and the simulation process, enabling accurate and efficient evaluation of predicted circuit parameters.

During the \textbf{training stage}, we follow the standard ML workflow. The input to the ML model consists of performance specifications, and the target output is the corresponding circuit parameters. The model is trained to minimize the $\ell_1$ loss, which measures the distance between predicted and actual circuit parameters.

\begin{algorithm}[H]
\caption{End-to-End Model Training Pipeline}
\label{alg:training}
\resizebox{0.48\textwidth}{!}{
\begin{minipage}{\linewidth}
\begin{algorithmic}[1]
\REQUIRE Training dataset $\mathcal{D}_{\text{train}} = \{ (\bm{x}, \bm{y}) \}$, machine learning model $\mathcal{M}$, maximum iterations \texttt{maxIter}, loss function $\mathcal{L}$.
\ENSURE Trained model $\mathcal{M}$.

\STATE \textbf{Initialize} model parameters of $\mathcal{M}$.

\FOR{$t = 1$ to \texttt{maxIter}}
    \STATE \textbf{Load data:} Sample a mini-batch of $b$ training pairs $\{ (\bm{x}_j, \bm{y}_j) \}_{j=1}^b$ from $\mathcal{D}_{\text{train}}$.
    \FOR{each training pair $(\bm{x}_j, \bm{y}_j)$ in the mini-batch}
        \STATE \textbf{Predict parameters:} Use the current model to predict the circuit parameters:
        \[
        \hat{\bm{y}}_j = \mathcal{M}(\bm{x}_j).
        \]
        \STATE \textbf{Compute loss:} Evaluate the loss between the predicted and ground truth circuit parameters:
        \[
        \text{Loss}_j = \mathcal{L}(\hat{\bm{y}}_j, \bm{y}_j).
        \]
    \ENDFOR
    \STATE \textbf{Update model:} Compute the gradient of the average loss over the mini-batch:
    \[
    \mathcal{L}_{\text{batch}} = \frac{1}{b} \sum_{j=1}^b \text{Loss}_j.
    \]
    Perform a gradient update on the model parameters to minimize $\mathcal{L}_{\text{batch}}$.
\ENDFOR

\STATE \textbf{Return} the trained model $\mathcal{M}$.

\end{algorithmic}
\end{minipage}
}
\end{algorithm}

In the \textbf{evaluation stage}, we assess the performance of the trained model using the evaluation dataset. For each performance specification vector $\bm{x}$, the trained model predicts the circuit parameters $\hat{\bm{y}}$. These parameters are then passed to the Cadence simulator to compute the corresponding simulated performance metrics, denoted as $\hat{\bm{x}}$. By comparing the simulated performance $\hat{\bm{x}}$ with the desired performance $\bm{x}$, we compute the relative error for each performance. The output of the evaluation stage includes relative error for each performance specification, aggregated summaries (e.g., mean relative error), and histograms of error distributions. 

\begin{algorithm}[H]
\caption{End-to-End Evaluation Pipeline}
\label{alg:evaluation}
\resizebox{0.48\textwidth}{!}{
\begin{minipage}{\linewidth}
\begin{algorithmic}[1]
\REQUIRE Trained model $\mathcal{M}$, evaluation dataset $\mathcal{D}_{\text{eval}} = \{ \bm{x} \}$, circuit simulator $\mathcal{S}$, maximum iterations \texttt{maxIter}.
\ENSURE Relative error list $\mathcal{E}$.

\STATE \textbf{Initialize} relative error list $\mathcal{E} \leftarrow \{\}$.

\FOR{$t = 1$ to \texttt{maxIter}}
    \STATE \textbf{Load data:} Select input performance vector $\bm{x} = \{x_1, x_2, \dots, x_N\} \in \mathcal{D}_{\text{eval}}$.
    \STATE \textbf{Predict circuit parameters:} Use the trained model $\mathcal{M}$ to predict:
    \[
    \hat{\bm{y}} = \mathcal{M}(\bm{x}).
    \]
    \STATE \textbf{Simulate performance:} Run the circuit simulator $\mathcal{S}$ to generate simulated performance:
    \[
    \hat{\bm{x}} = \mathcal{S}(\hat{\bm{y}}) = \{\hat{x}_1, \hat{x}_2, \dots, \hat{x}_N\}.
    \]
    \STATE \textbf{Compute errors:} For each performance metric \( i = 1, \dots, N \), compute the relative error:
    \[
    e_i = \frac{|x_i - \hat{x}_i|}{x_i}.
    \]
    \STATE Append $\{e_1, e_2, \dots, e_N\}$ to the error list $\mathcal{E}$.
\ENDFOR

\STATE \textbf{Return} relative error list $\mathcal{E}$ and compute aggregated metrics (e.g., mean relative error, histograms).
\end{algorithmic}
\end{minipage}
}
\end{algorithm}

A key feature of this pipeline is the inclusion of a Cadence simulator in the evaluation phase. This allows us to accurately obtain performance metrics under realistic circuit conditions, including effects with inherent randomness, such as noise figure in low-noise amplifiers (LNA).
This advanced analysis enables a more comprehensive evaluation compared to prior works, which often overlook such nuances.

\section{Experiments}\label{sec:experiment}

In this section, we evaluate ML models for predicting circuit parameters from performance metrics, categorizing the circuits into two groups: \textbf{homogeneous circuits} and \textbf{heterogeneous circuits}. For each circuit, we analyze the accuracy and robustness of the ML models by comparing their predictions against simulation results and reporting evaluation metrics, including relative error distributions, mean error, error percentiles, and outlier percentages. This circuit-by-circuit analysis highlights the strengths and limitations of ML models for different circuit complexities and performance requirements.


To account for the different numerical ranges of performance metrics and circuit parameters, we normalize all data to the range \([-1, 1]\). For each circuit, we randomly split the dataset into 90\% for training and 10\% for testing. We train the neural network models, including transformers and MLPs, for 100 epochs using the Adam optimizer~\cite{arXiv17_Adam} with a learning rate of 0.001. Each training process is repeated multiple times with different random seeds to ensure the robustness of results.


\subsection{Homogeneous Circuits}
\begin{figure}[ht]
    \centering
    \vspace{-3mm}
    \subfloat{\includegraphics[width=0.24\textwidth]{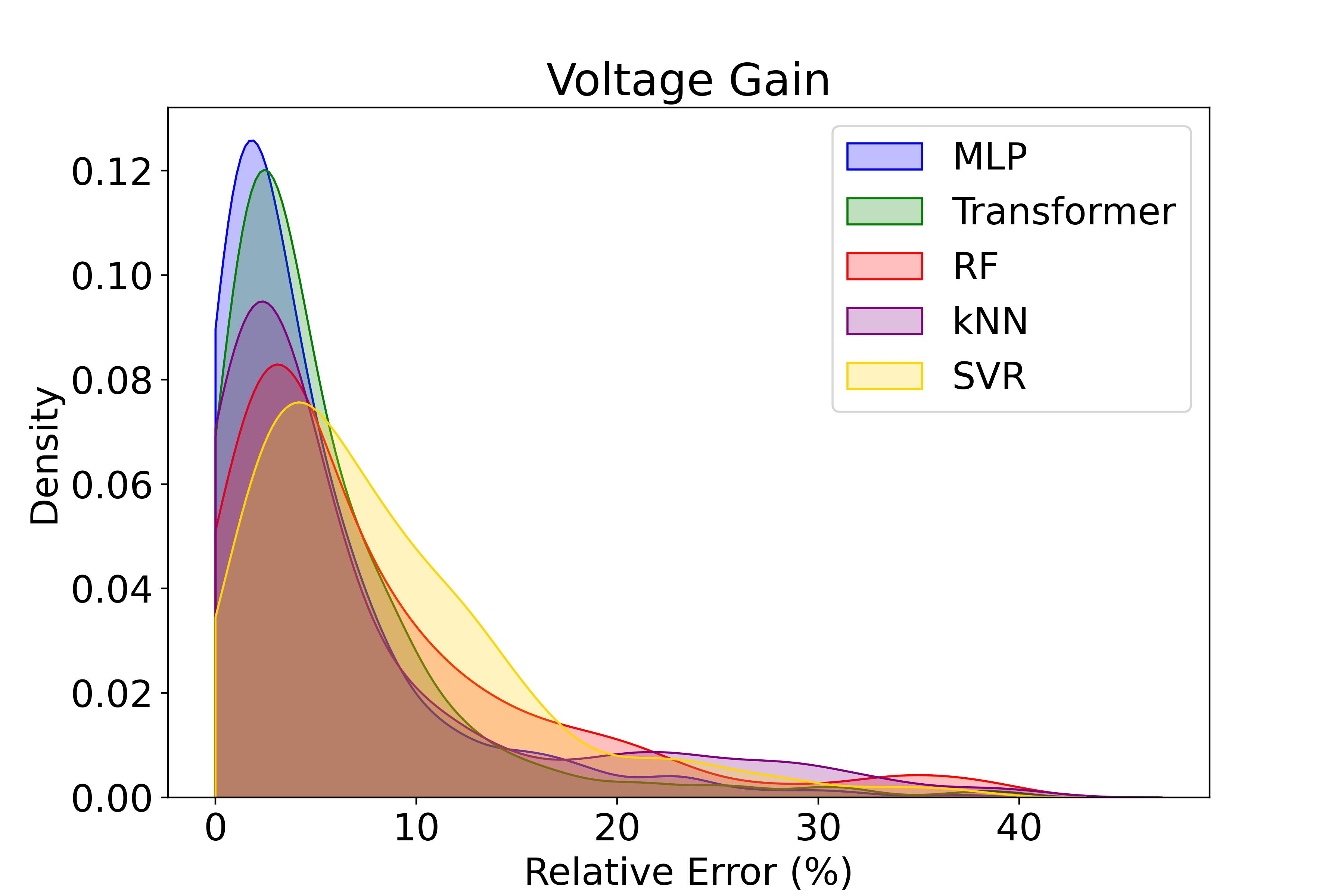}}
    \subfloat{\includegraphics[width=0.24\textwidth]{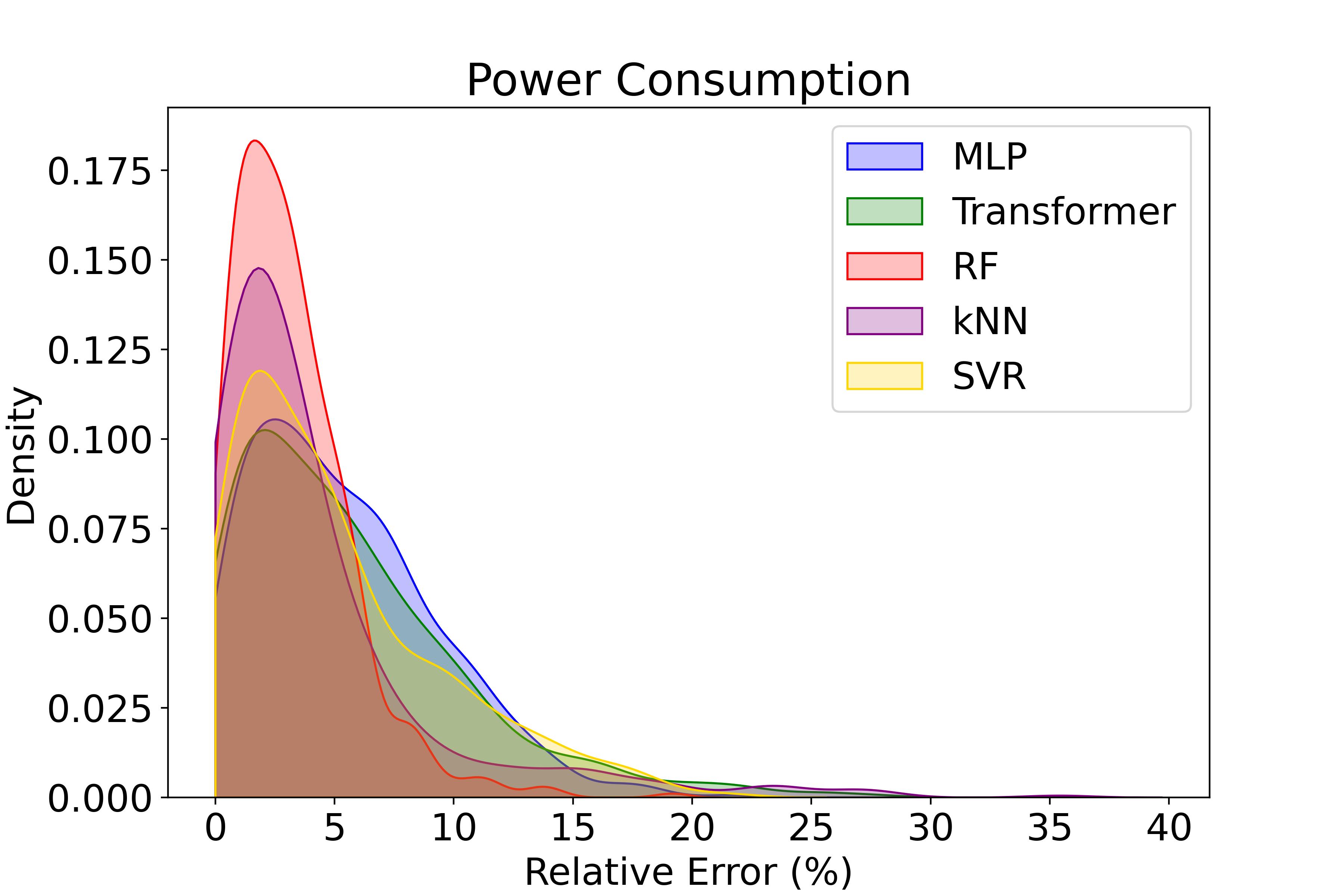}} \\ [-1.5ex]
    \subfloat{\includegraphics[width=0.24\textwidth]{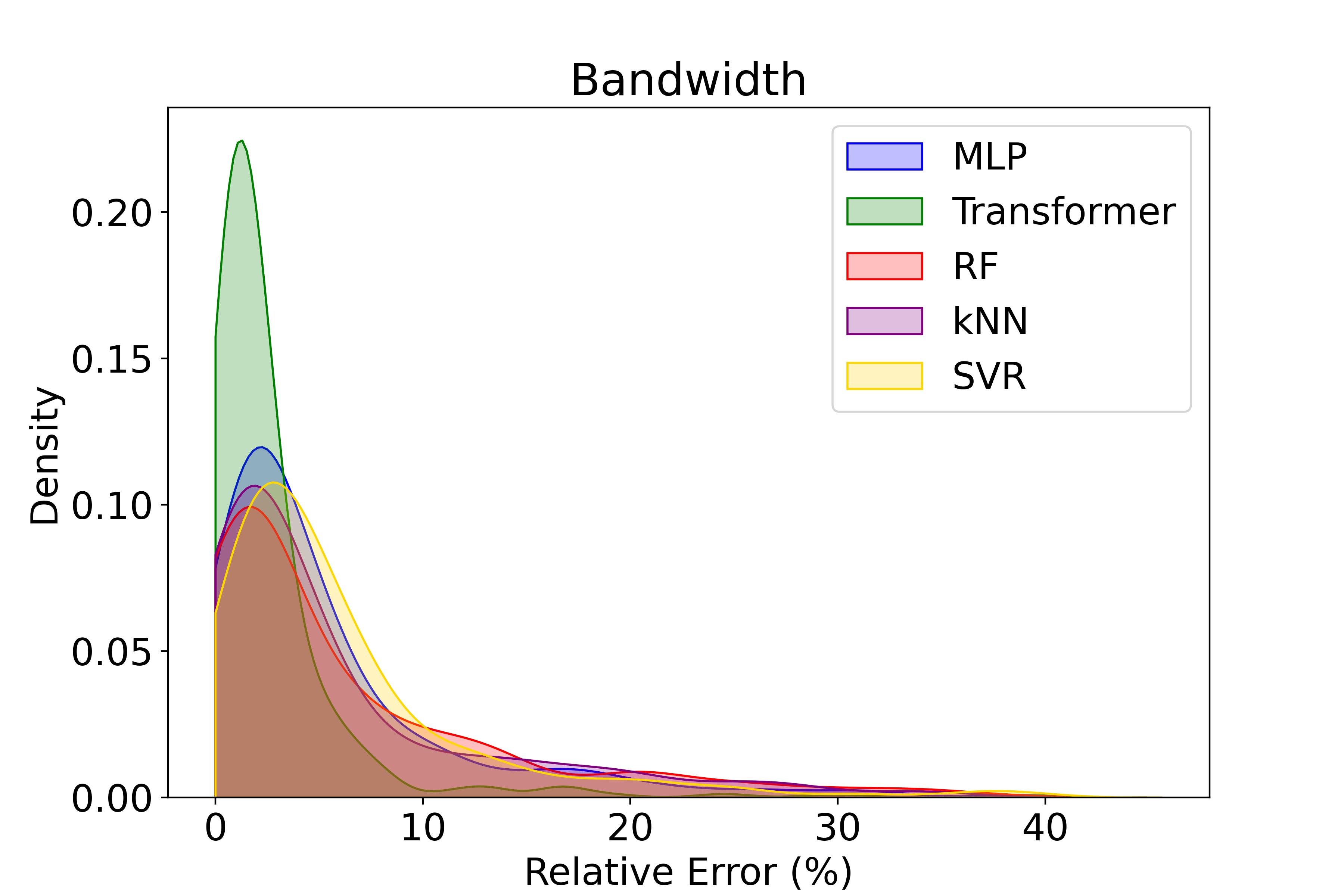}}
    \subfloat{\includegraphics[width=0.24\textwidth]{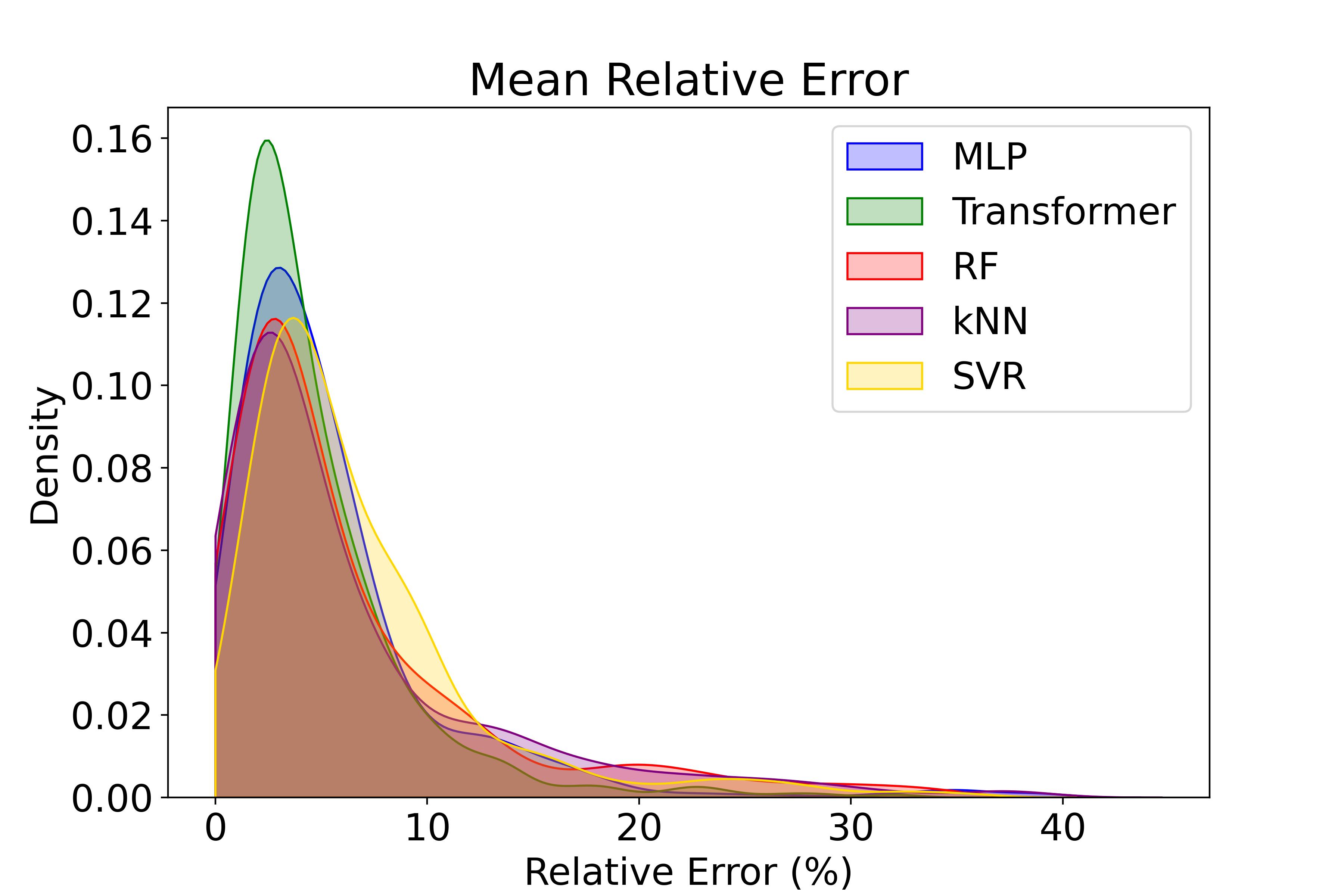}}
    \caption{\centering Relative error histogram of individual performance metrics and mean relative error histogram for CSVA.}
    \label{fig:csva_plots}
\end{figure}

\textbf{Common-Source Voltage Amplifier (CSVA).}  The results for CSVA, presented in Table~\ref{tab:csva_summary} and Figure~\ref{fig:csva_plots}, show that the \textbf{Transformer} model outperforms other models with the most accurate predictions and minimal error spread. The \textbf{MLP} follows closely, delivering competitive performance with low outlier rates. In contrast, traditional methods, \textbf{RF}, \textbf{kNN}, and \textbf{SVR}, exhibit higher error distributions and less robustness, as seen in their wider histograms and greater outlier percentages. These results highlight the effectiveness of deep learning models for predicting CSVA parameters.

\vspace{-3mm}

\begin{table}[ht!]
\caption{Statistical summary of mean relative error for CSVA}
\label{tab:csva_summary}
\centering
\renewcommand{\arraystretch}{0.8} 
\resizebox{0.48\textwidth}{!}{
\begin{tabular}{cccccccc}
\toprule
\raisebox{0.5em}{\textbf{Model}} & 
\raisebox{0.5em}{\textbf{Mean}} & 
\raisebox{0.5em}{\textbf{Std}} & 
\raisebox{0.5em}{\textbf{P75}} & 
\raisebox{0.5em}{\textbf{P90}} & 
\textbf{\shortstack{\% Errors \\ $<$ 2\% }} & 
\textbf{\shortstack{\% Errors \\ $<$ 5\% }} & 
\textbf{\shortstack{\% Outlier \\ ($>$ 20\%)}} \\

\midrule
\textbf{MLP} & 5.96 & 7.54 & 6.46 & 12.27 & 19.8 & 60.2 & 3.2 \\

\midrule
\textbf{Transformer} & \textbf{5.00} & \textbf{6.13} & \textbf{5.84} & \textbf{9.58} & \textbf{26.4} & \textbf{67.4} & \textbf{2.4} \\

\midrule
\textbf{RF} & 7.99 & 11.69 & 8.80 & 19.45 & 20.8 & 59.4 & 9.4 \\

\midrule
\textbf{kNN} & 7.80 & 11.76 & 7.95 & 17.81 & 25.6 & 61.4 & 8.6 \\
        
\midrule
\textbf{SVR} & 9.56 & 14.91 & 9.04 & 16.61 & 9.6 & 49.8 & 8.6 \\
\bottomrule
\end{tabular}
}
\end{table}

\begin{figure}[ht!]
    \centering
    \vspace{-5mm}
    \subfloat{\includegraphics[width=0.24\textwidth]{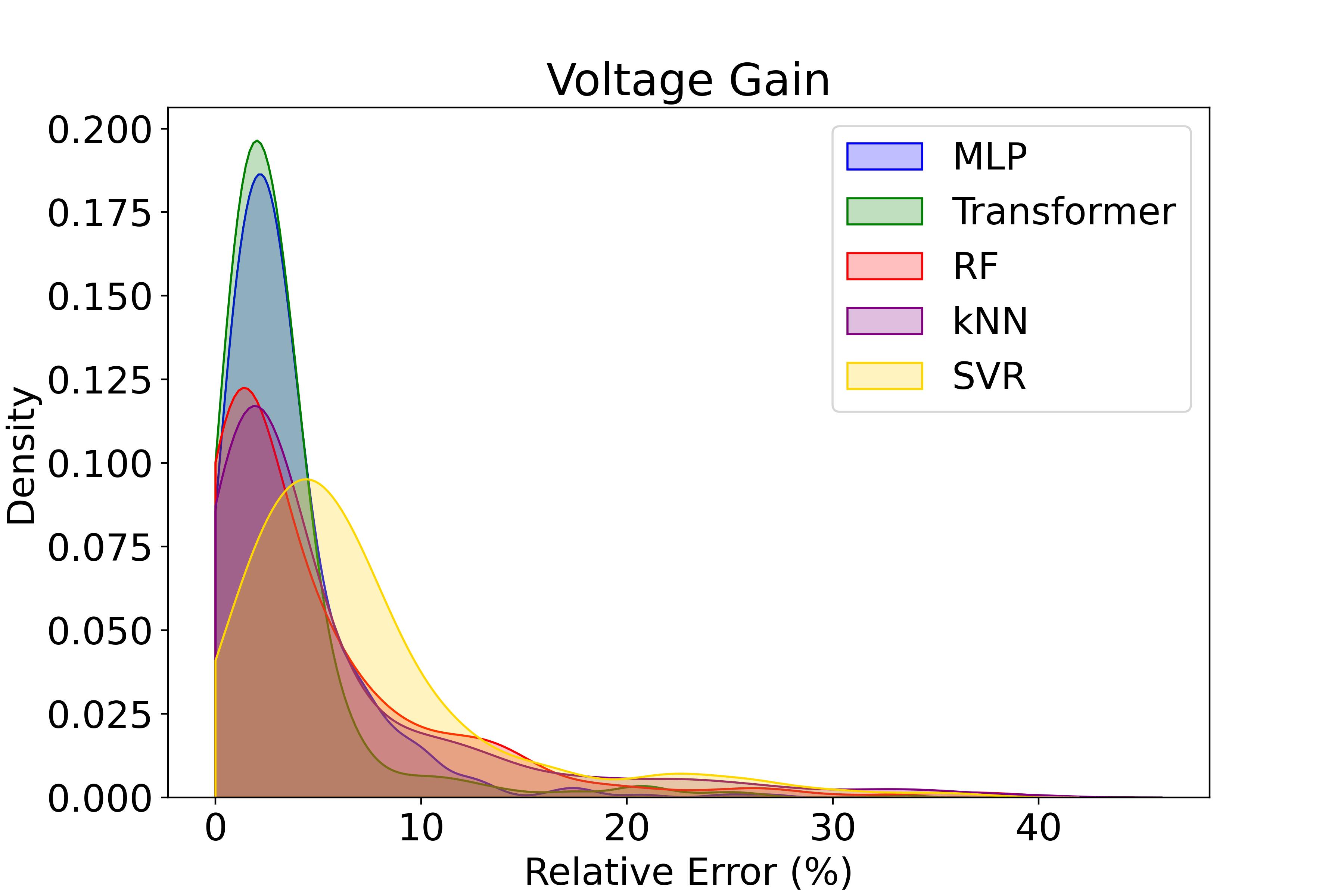}}
    \subfloat{\includegraphics[width=0.24\textwidth]{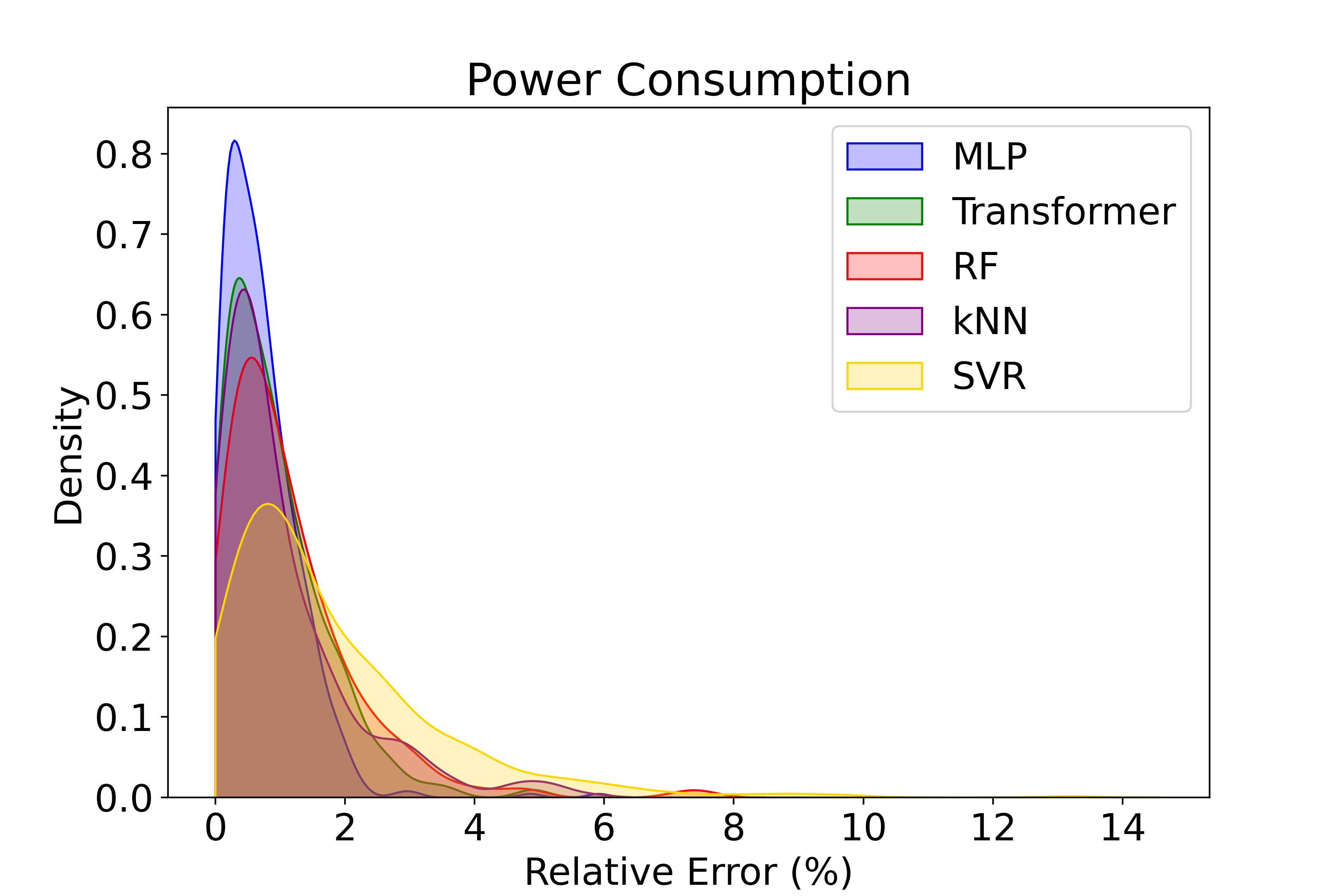}} \\ [-1.5ex]
    \subfloat{\includegraphics[width=0.24\textwidth]{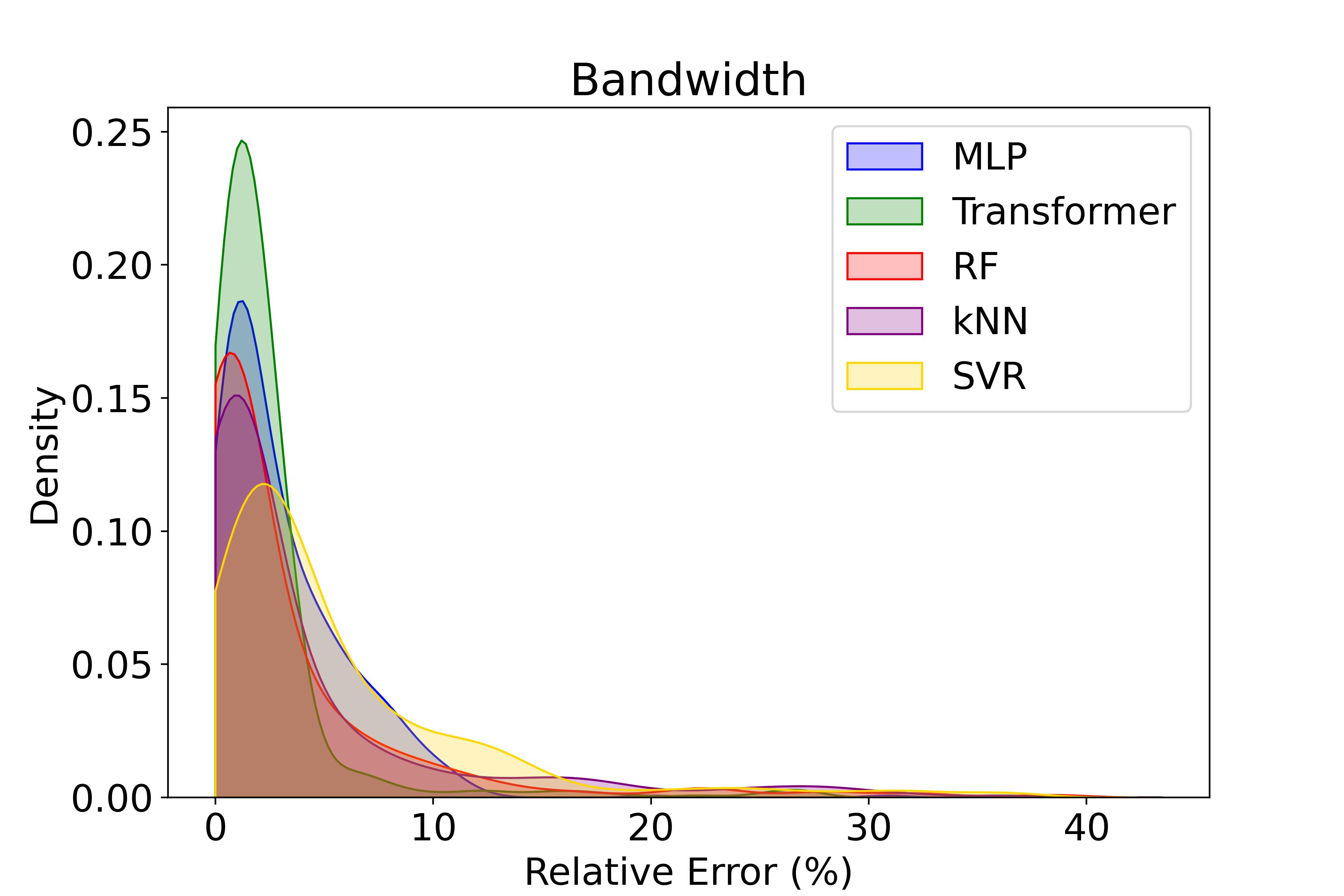}}
    \subfloat{\includegraphics[width=0.24\textwidth]{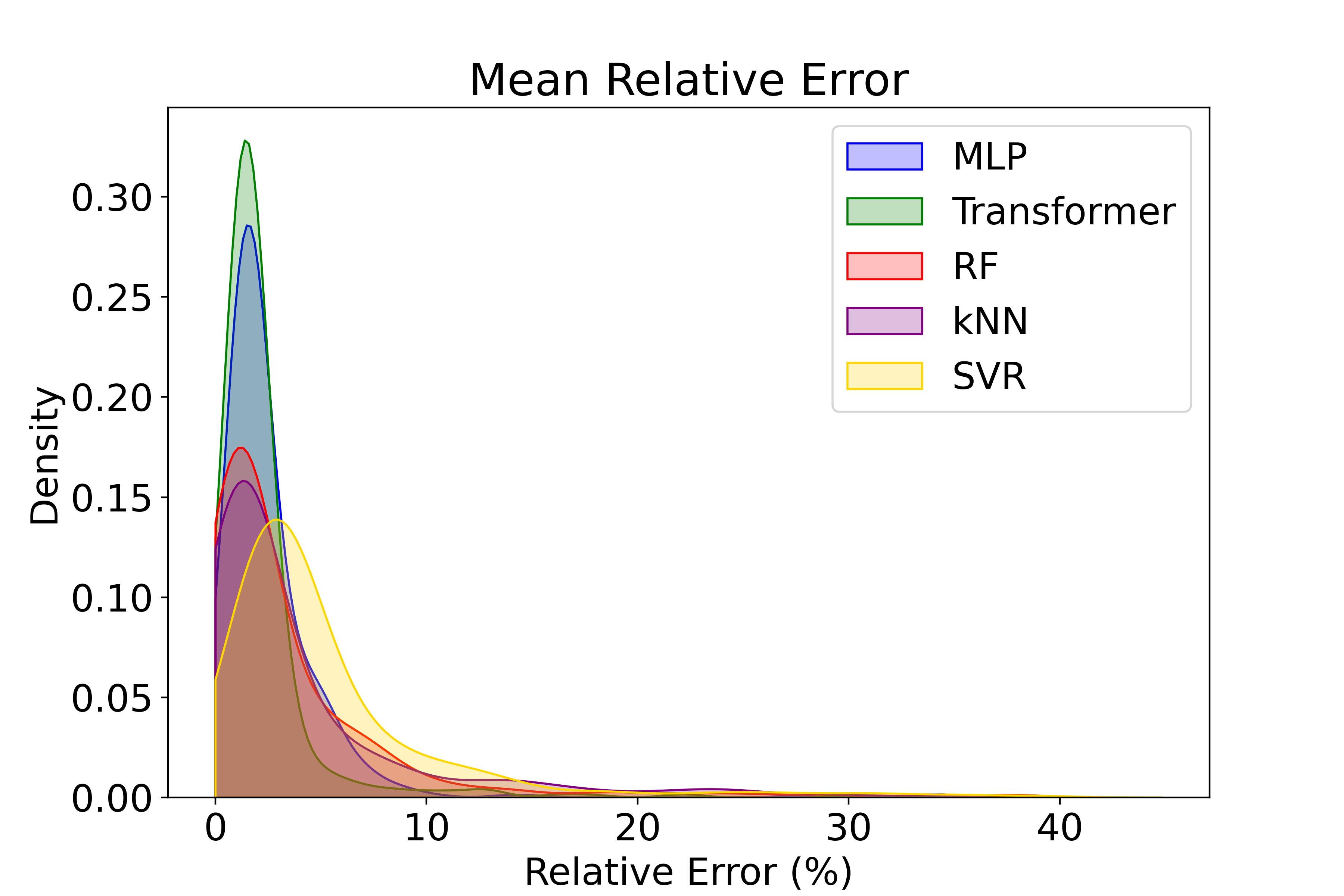}}
    \caption{\centering Relative error histogram of individual performance metrics and mean relative error histogram for CVA.}
    \label{fig:cva_plots}
\end{figure}

\textbf{Cascode Voltage Amplifier (CVA).} The results for CVA, presented in Table~\ref{tab:cva_summary} and Figure~\ref{fig:cva_plots}, show that the \textbf{Transformer} model outperforms all others with the lowest mean error, smallest spread, and the highest percentage of low-error predictions. The \textbf{MLP} follows closely with competitive performance and minimal outliers. In contrast, traditional methods (\textbf{RF}, \textbf{kNN}, and \textbf{SVR}) display larger error distributions, with SVR having the highest error spread and outlier rate. These results confirm the superior accuracy and robustness of deep learning models, for CVA parameter prediction.

\vspace{-3mm}

\begin{table}[ht!]
\caption{Statistical summary of mean relative error for CVA}
\label{tab:cva_summary}
\centering
\renewcommand{\arraystretch}{0.8} 
\resizebox{0.48\textwidth}{!}{
\begin{tabular}{cccccccc}
\toprule
\raisebox{0.5em}{\textbf{Model}} & 
\raisebox{0.5em}{\textbf{Mean}} & 
\raisebox{0.5em}{\textbf{Std}} & 
\raisebox{0.5em}{\textbf{P75}} & 
\raisebox{0.5em}{\textbf{P90}} & 
\textbf{\shortstack{\% Errors \\ $<$ 2\% }} & 
\textbf{\shortstack{\% Errors \\ $<$ 5\% }} & 
\textbf{\shortstack{\% Outlier \\ ($>$ 20\%)}} \\

\midrule
\textbf{MLP} & 2.81 & 4.77 & 2.95 & 5.04 & 56.4	& 89.6 & \textbf{1.0} \\

\midrule
\textbf{Transformer} & \textbf{2.34} & \textbf{3.74} & \textbf{2.08} & \textbf{3.29} & \textbf{72.6} & \textbf{93.4} & \textbf{1.0} \\

\midrule
\textbf{RF} & 4.24 & 9.03 & 3.92 & 8.10 & 59.4 & 78.8 & 4.0 \\

\midrule
\textbf{kNN} & 4.61 & 8.24 & 4.24 & 12.59 & 57.0 & 77.8 & 5.4 \\
        
\midrule
\textbf{SVR} & 8.44 & 15.30 & 6.82 & 15.22 & 19.6 & 67.0 & 8.6 \\
\bottomrule
\end{tabular}
}
\end{table}

\begin{figure}[ht!]
    \centering
    \vspace{-3mm}
    \subfloat{\includegraphics[width=0.24\textwidth]{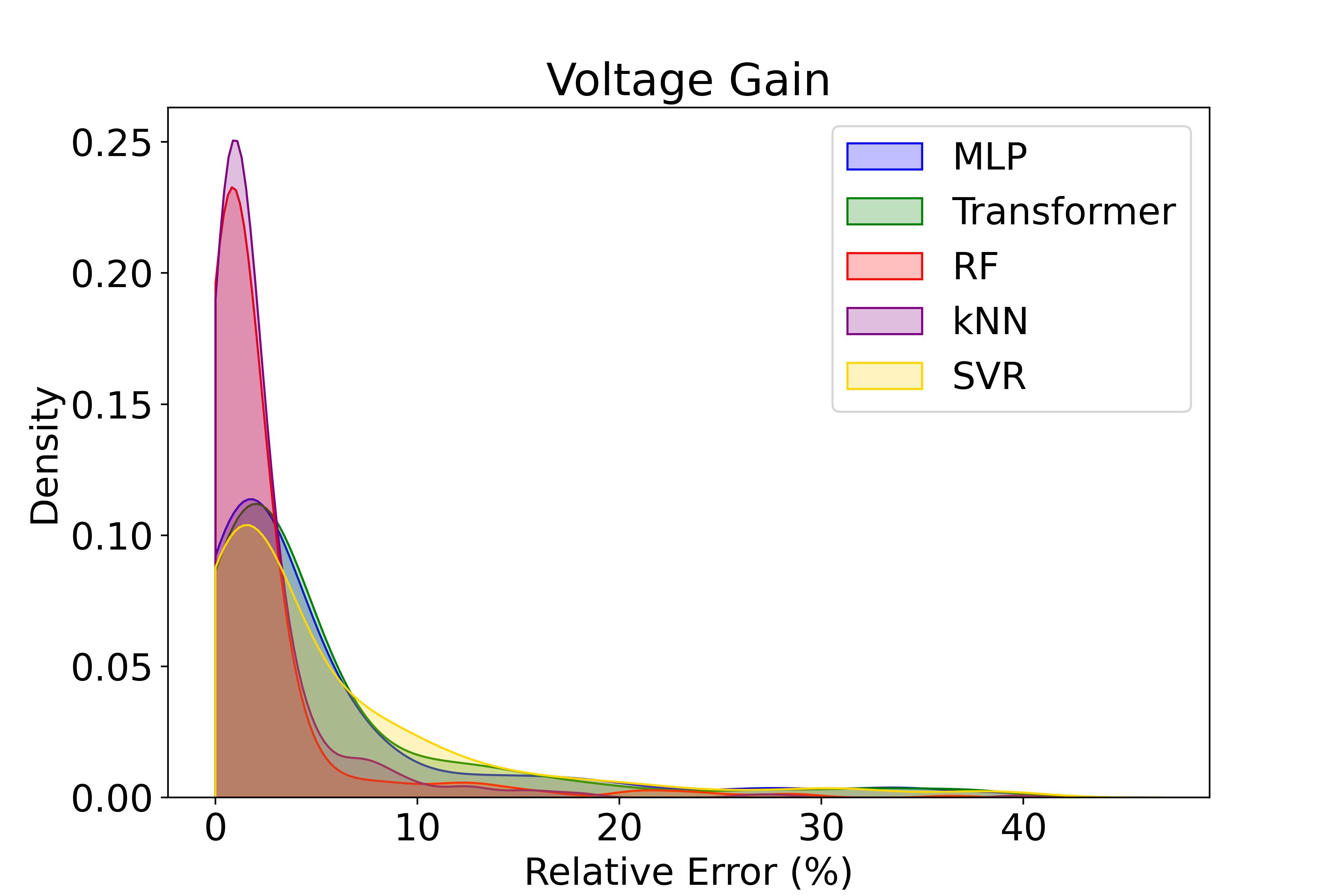}}
    \subfloat{\includegraphics[width=0.24\textwidth]{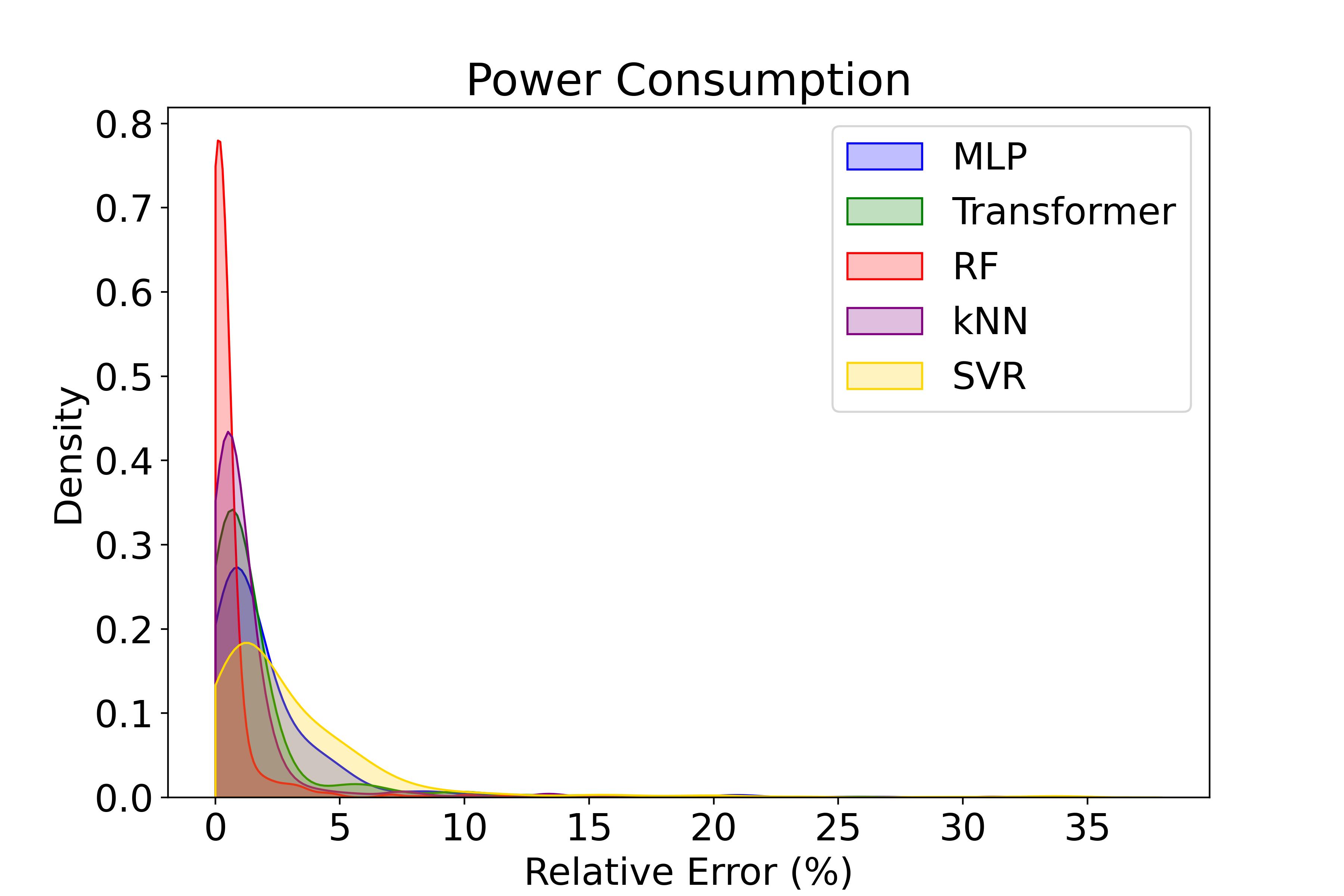}} \\ [-1.5ex]
    \subfloat{\includegraphics[width=0.24\textwidth]{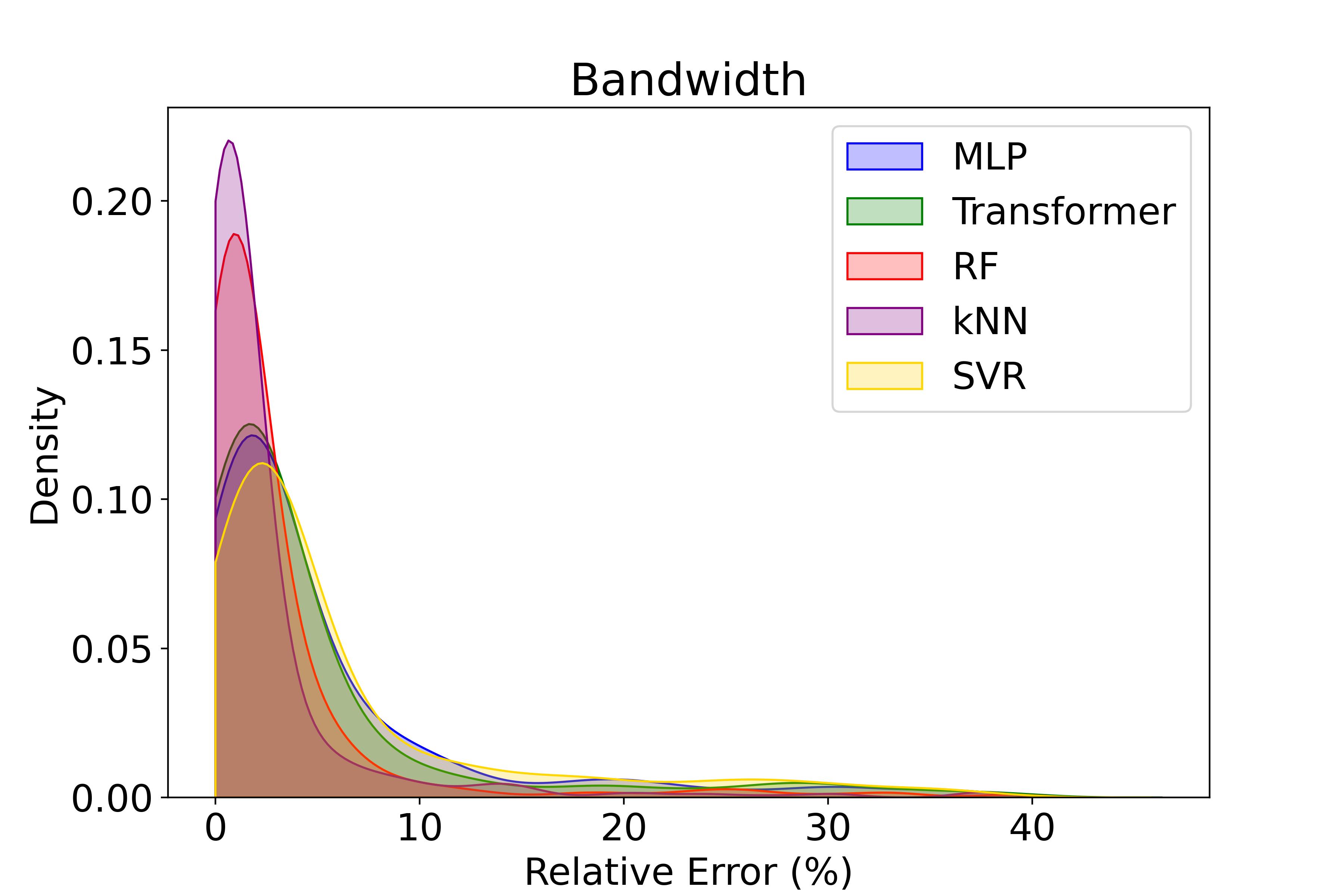}}
    \subfloat{\includegraphics[width=0.24\textwidth]{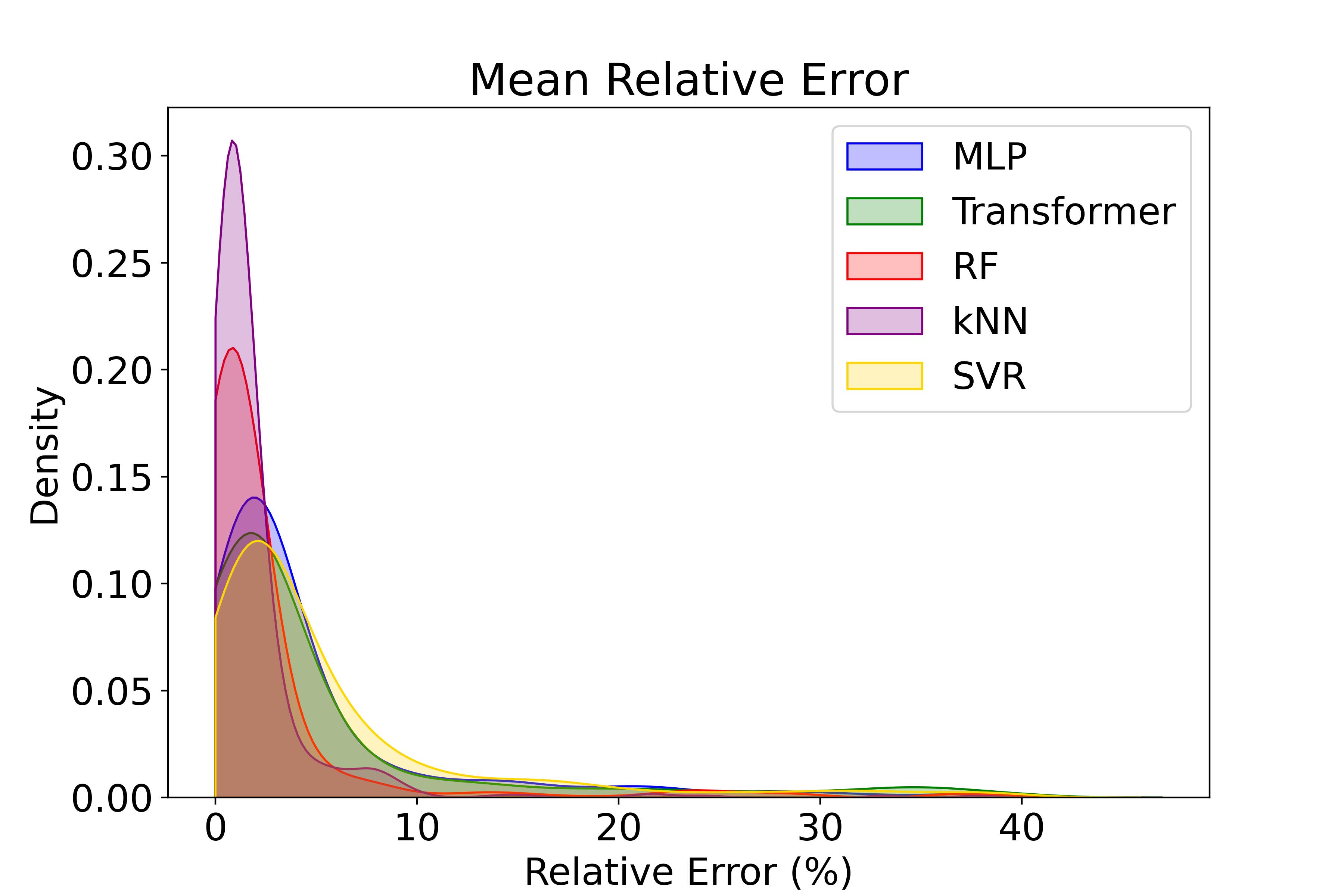}}
    \caption{\centering Relative error histogram of individual performance metrics and mean relative error histogram for TSVA.}
    \label{fig:tsva_plots}
\end{figure}

\textbf{Two-Stage Voltage Amplifier (TSVA).}
The results for TSVA, shown in Table~\ref{tab:tsva_summary} and Figure~\ref{fig:tsva_plots}, reveal that \textbf{RF} and \textbf{kNN} perform exceptionally well, achieving the highest percentage of low-error predictions and the smallest outlier rates. In particular, \textbf{RF} achieves the best overall accuracy with concentrated error distributions and minimal spread. In contrast, the \textbf{Transformer} and \textbf{MLP} exhibit higher mean errors and larger spreads, indicating challenges in capturing the parameter-performance mapping for TSVA. Among all models, \textbf{SVR} shows the weakest performance with the highest outlier rate and error spread. 
These results suggest that simpler regression-based methods can outperform deep learning approaches for circuits like TSVA.

\vspace{-3mm}

\begin{table}[ht!]
\caption{Statistical summary of mean relative error for TSVA}
\label{tab:tsva_summary}
\centering
\renewcommand{\arraystretch}{0.8} 
\resizebox{0.48\textwidth}{!}{
\begin{tabular}{cccccccc}
\toprule
\raisebox{0.5em}{\textbf{Model}} & 
\raisebox{0.5em}{\textbf{Mean}} & 
\raisebox{0.5em}{\textbf{Std}} & 
\raisebox{0.5em}{\textbf{P75}} & 
\raisebox{0.5em}{\textbf{P90}} & 
\textbf{\shortstack{\% Errors \\ $<$ 2\% }} & 
\textbf{\shortstack{\% Errors \\ $<$ 5\% }} & 
\textbf{\shortstack{\% Outlier \\ ($>$ 20\%)}} \\

\midrule
\textbf{MLP} & 8.66 & 19.26 & 5.47 & 21.46 & 43.6 & 74.2 & 11.0 \\

\midrule
\textbf{Transformer} & 11.53 & 27.36 & 6.18 & 34.83 & 49.8 & 73.0 & 15.0 \\

\midrule
\textbf{RF} & \textbf{3.30} & \textbf{10.74} & \textbf{1.60} & \textbf{4.66} & \textbf{80.0} & \textbf{90.6} & \textbf{4.6} \\

\midrule
\textbf{kNN} & 5.95 & 26.69 & 1.87 & 5.46 & 77.0 & 89.2 & \textbf{4.6} \\
        
\midrule
\textbf{SVR} & 12.43 & 24.88 & 8.08 & 37.53 & 37.8 & 64.4 & 14.6 \\
\bottomrule
\end{tabular}
}
\end{table}

\begin{figure}[ht!]
    \centering
    \vspace{-3mm}
    \subfloat{\includegraphics[width=0.24\textwidth]{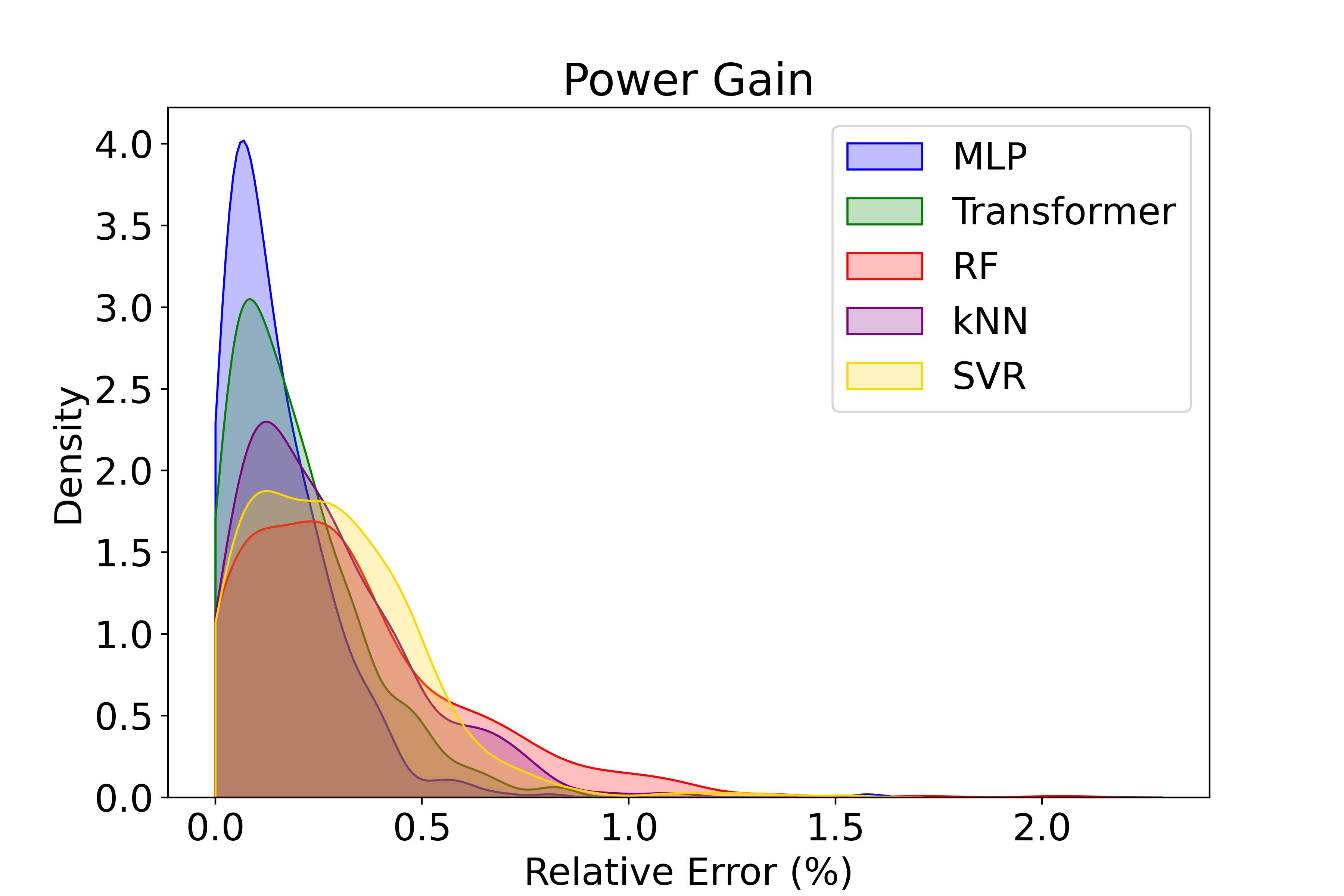}}
    \subfloat{\includegraphics[width=0.24\textwidth]{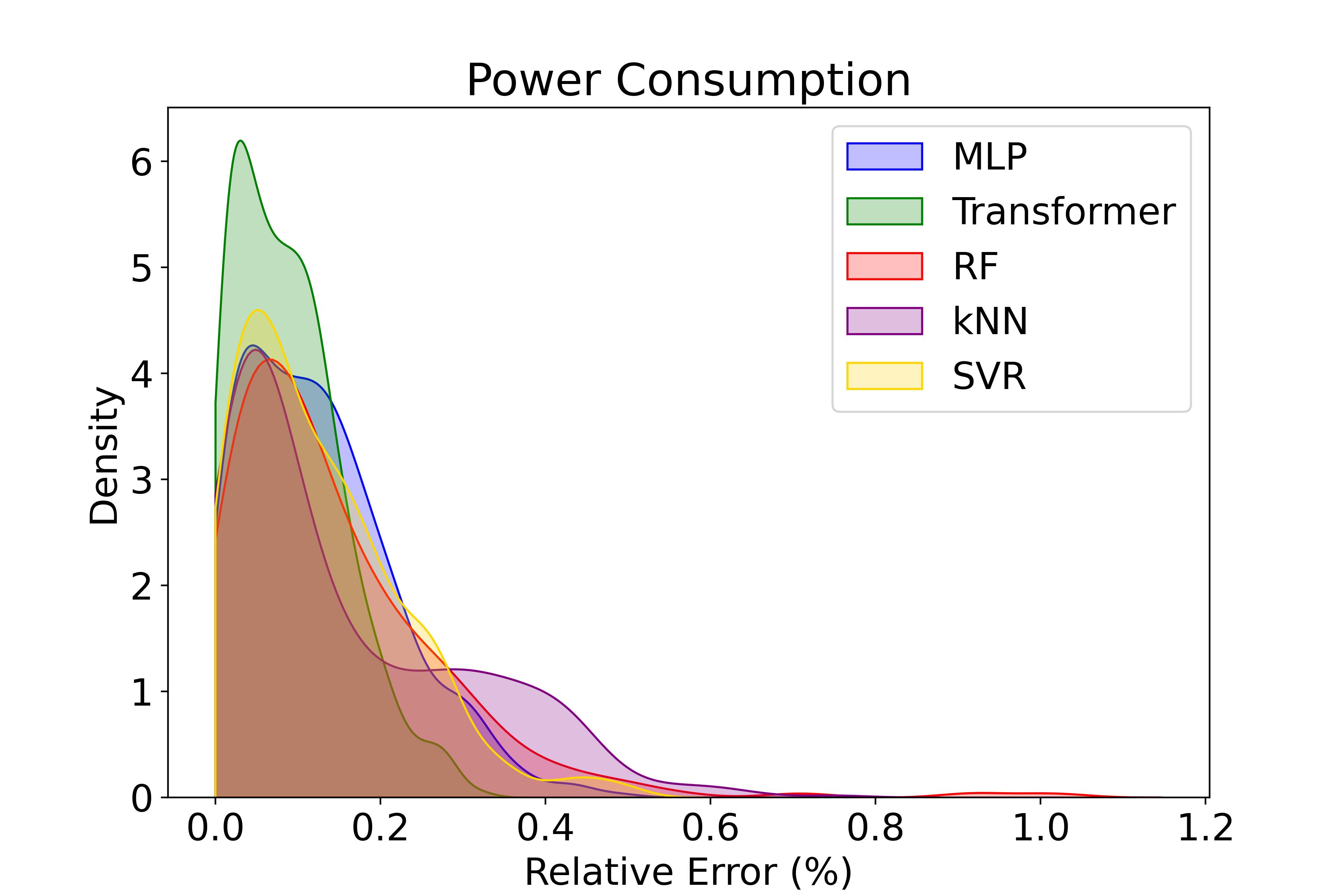}} \\ [-1.5ex]
    \subfloat{\includegraphics[width=0.24\textwidth]{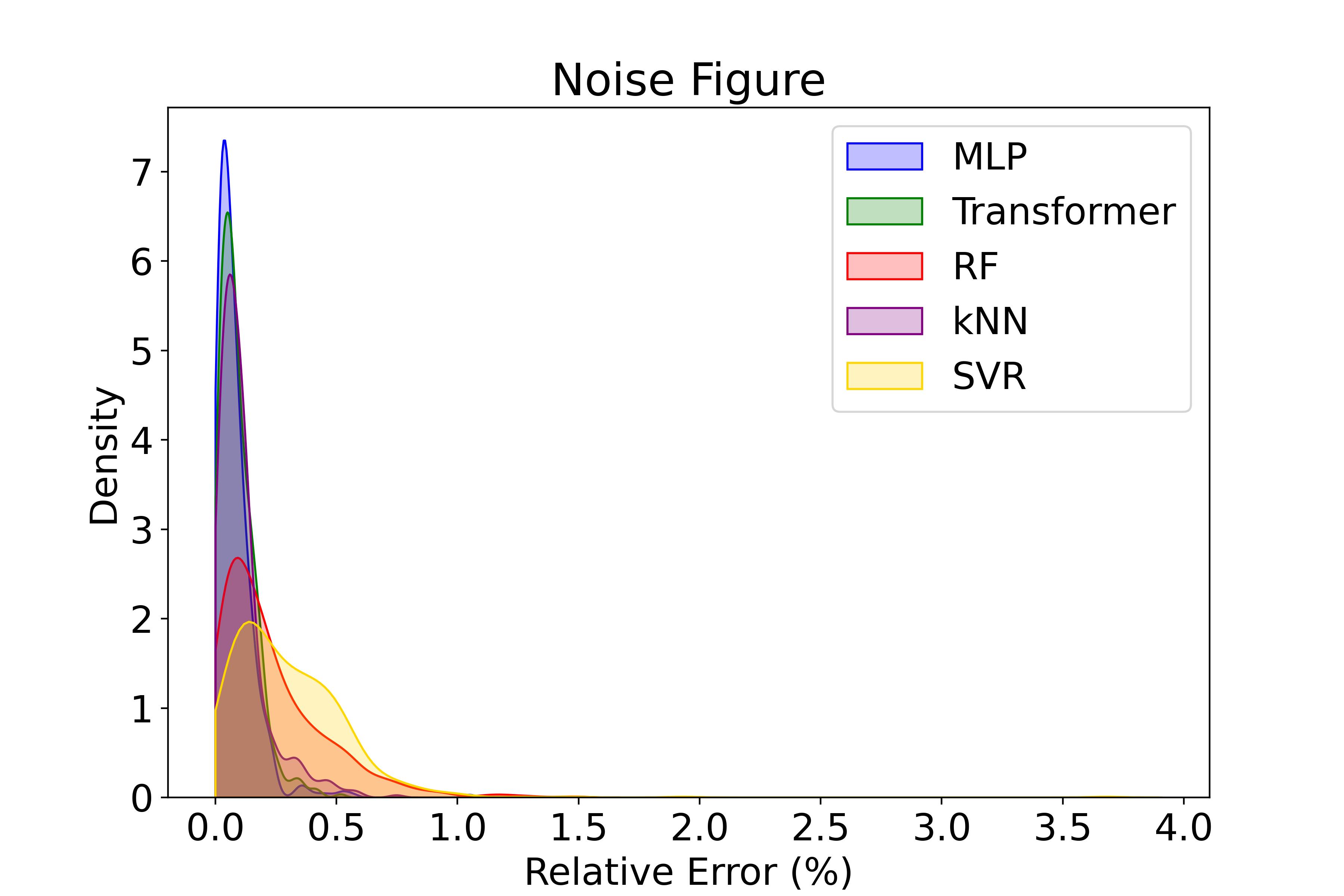}} 
    \subfloat{\includegraphics[width=0.24\textwidth]{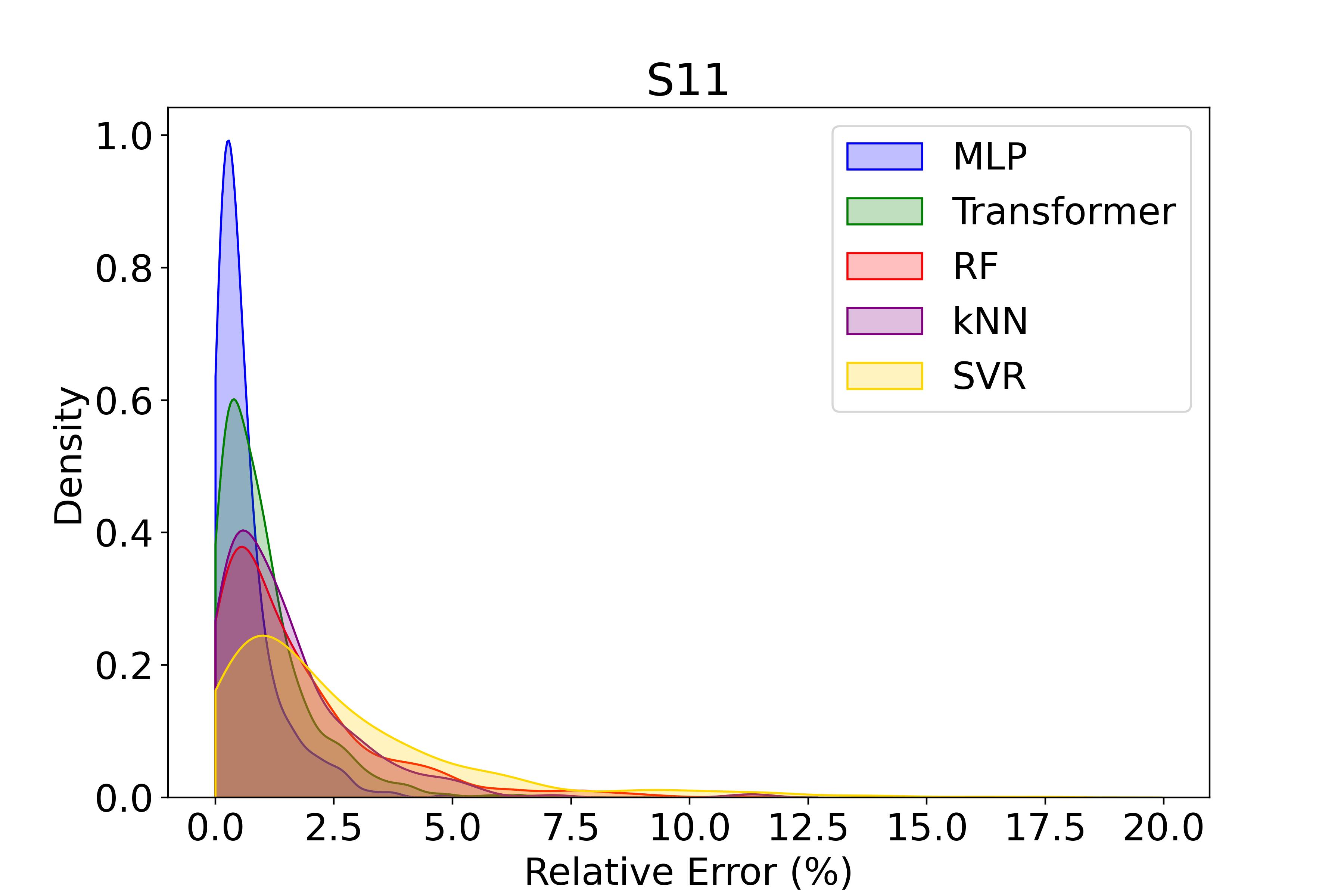}} \\ [-1.5ex]
    \subfloat{\includegraphics[width=0.24\textwidth]{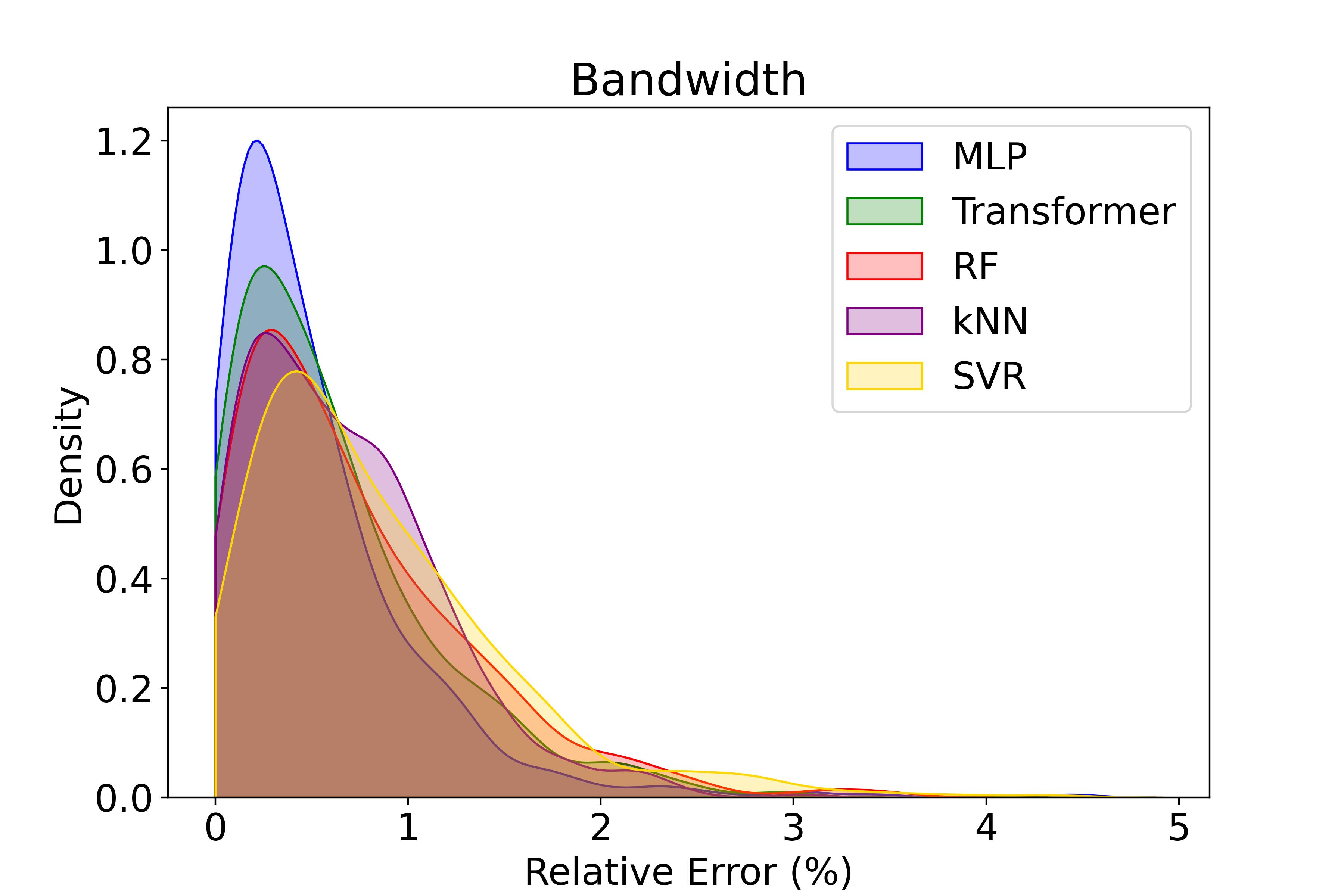}} 
    \subfloat{\includegraphics[width=0.24\textwidth]{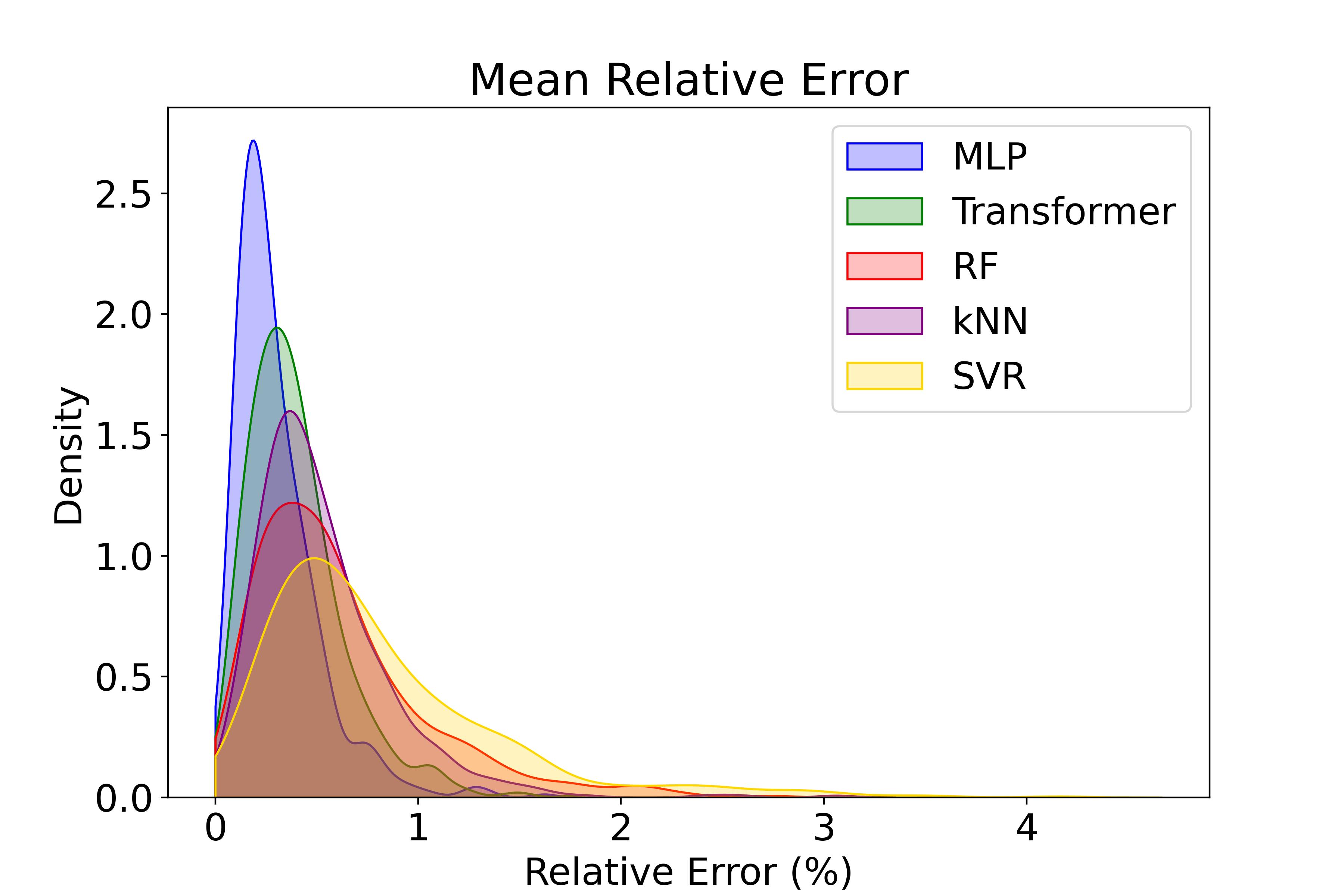}}
    \caption{\centering Relative error histogram of individual performance metrics and mean relative error histogram for LNA.}
    \label{fig:lna_plots}
\end{figure}

\textbf{Low-Noise Amplifier (LNA).} 
The results for LNA, shown in Table~\ref{tab:lna_summary} and Figure~\ref{fig:lna_plots}, demonstrate \textbf{outstanding performance} across all models. Both the \textbf{Transformer} and \textbf{MLP} achieve near-perfect accuracy, with all errors below 2\% and no outliers. Traditional methods (\textbf{RF}, \textbf{kNN}, and \textbf{SVR}) also perform remarkably well, with minimal error distributions and zero outliers. The mean relative errors and standard deviations for all models are remarkably low, as seen in the narrow and sharp error histograms. The results indicate that LNA's performance metrics are easier to predict accurately compared to other circuits, likely due to its well-behaved parameter-performance relationships. Overall, this highlights the robustness and reliability of all evaluated models for LNA.

\vspace{-3mm}

\begin{table}[ht!]
\caption{Statistical summary of mean relative error for LNA}
\label{tab:lna_summary}
\centering
\renewcommand{\arraystretch}{0.8} 
\resizebox{0.48\textwidth}{!}{
\begin{tabular}{cccccccc}
\toprule
\raisebox{0.5em}{\textbf{Model}} & 
\raisebox{0.5em}{\textbf{Mean}} & 
\raisebox{0.5em}{\textbf{Std}} & 
\raisebox{0.5em}{\textbf{P75}} & 
\raisebox{0.5em}{\textbf{P90}} & 
\textbf{\shortstack{\% Errors \\ $<$ 2\% }} & 
\textbf{\shortstack{\% Errors \\ $<$ 5\% }} & 
\textbf{\shortstack{\% Outlier \\ ($>$ 20\%)}} \\

\midrule
\textbf{MLP} & \textbf{0.30} & \textbf{0.21} & \textbf{0.38} & \textbf{0.54} & \textbf{100.0} & \textbf{100.0} & \textbf{0.0} \\

\midrule
\textbf{Transformer} & 0.40 & 0.24 & 0.51 & 0.70 & \textbf{100.0} & \textbf{100.0} & \textbf{0.0} \\

\midrule
\textbf{RF} & 0.61 & 0.43 & 0.77 & 1.19 & 98.2 & \textbf{100.0} & \textbf{0.0} \\

\midrule
\textbf{kNN} & 0.53 & 0.34 & 0.65 & 0.91 & 99.4 & \textbf{100.0} & \textbf{0.0} \\
        
\midrule
\textbf{SVR} & 0.81 & 0.59 & 1.03 & 1.49 & 94.8 & \textbf{100.0} & \textbf{0.0} \\
\bottomrule
\end{tabular}
}
\end{table}

\begin{figure}[ht!]
    \centering
    \vspace{-3mm}
    \subfloat{\includegraphics[width=0.24\textwidth]{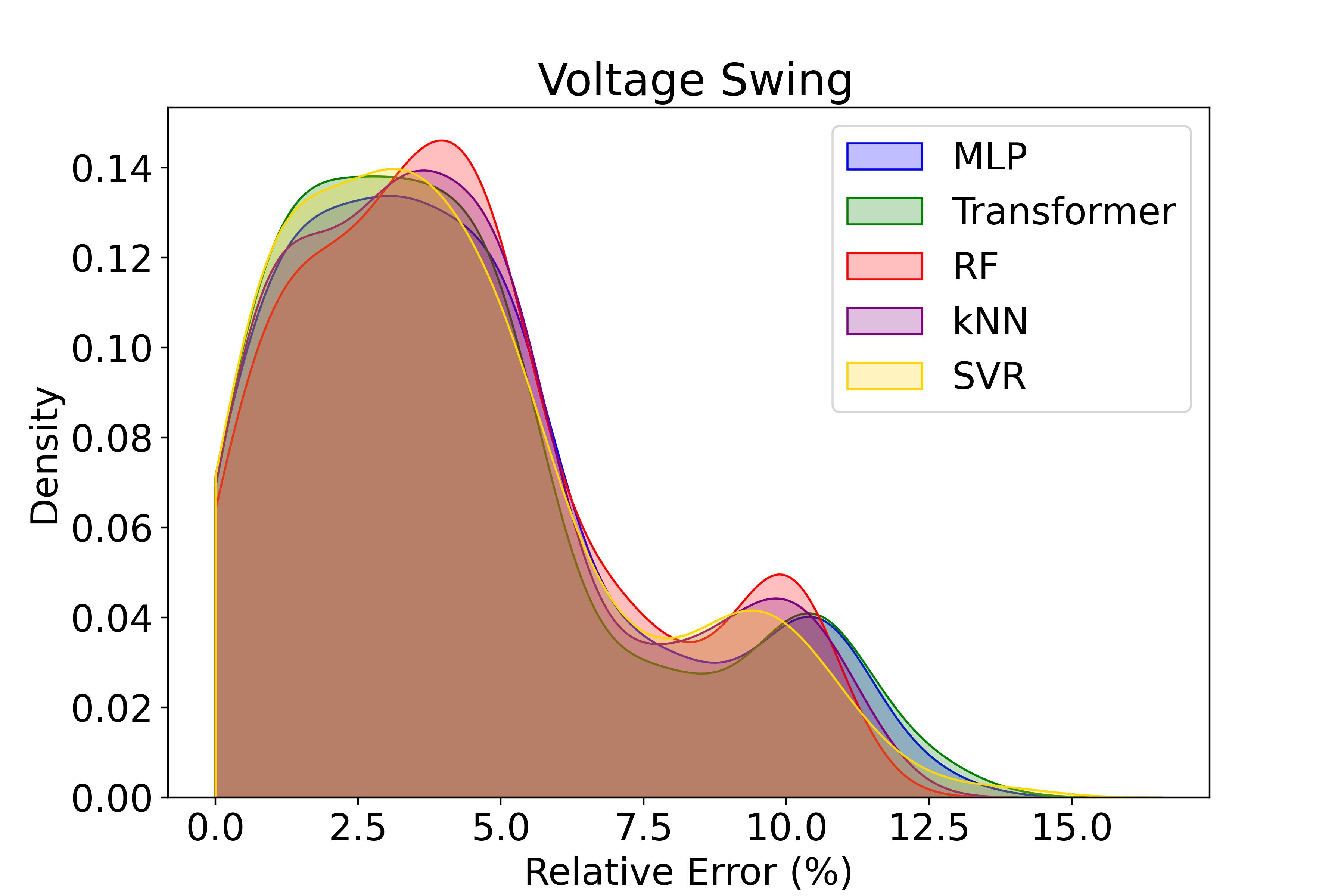}}
    \subfloat{\includegraphics[width=0.24\textwidth]{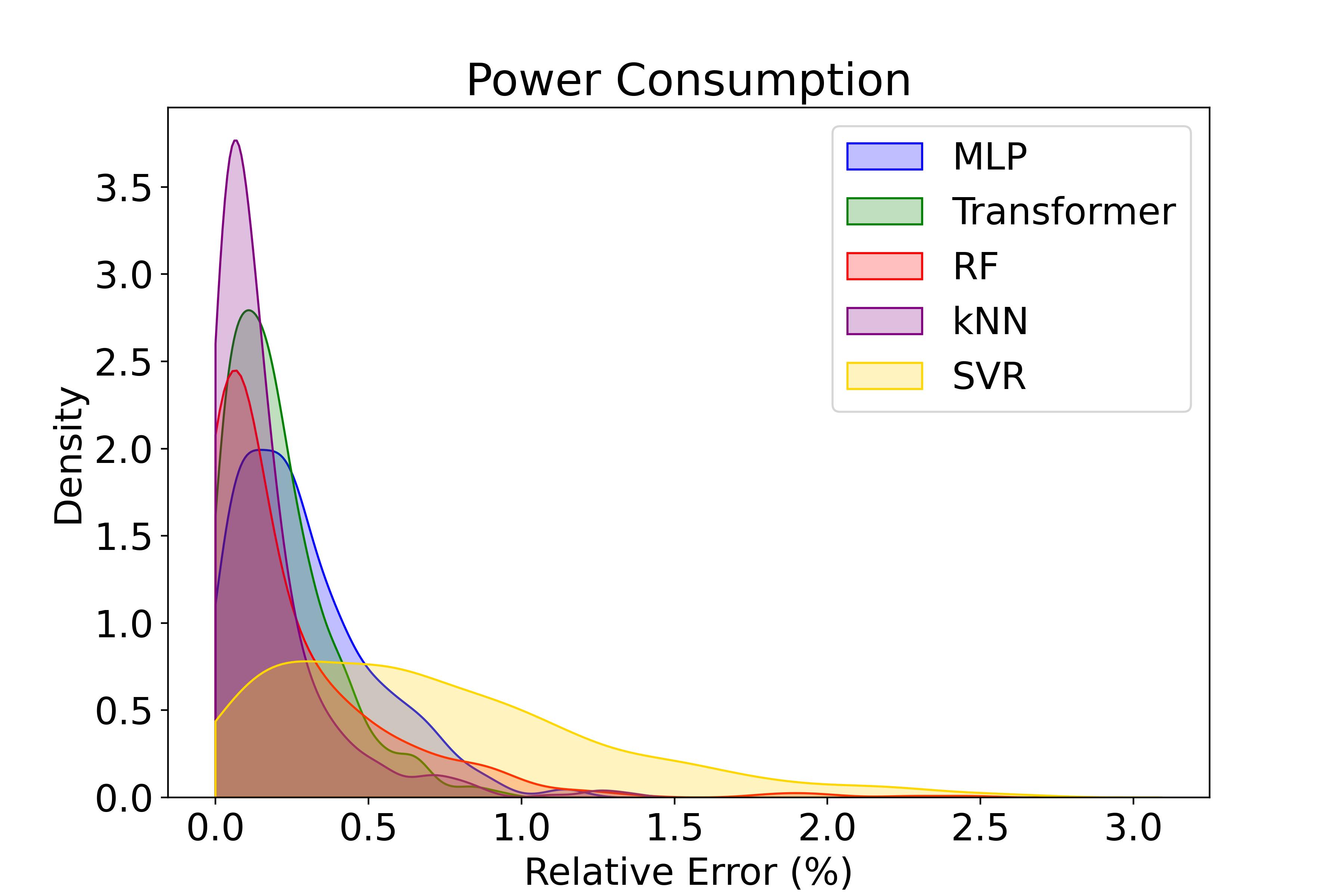}} \\ [-1.5ex]
    \subfloat{\includegraphics[width=0.24\textwidth]{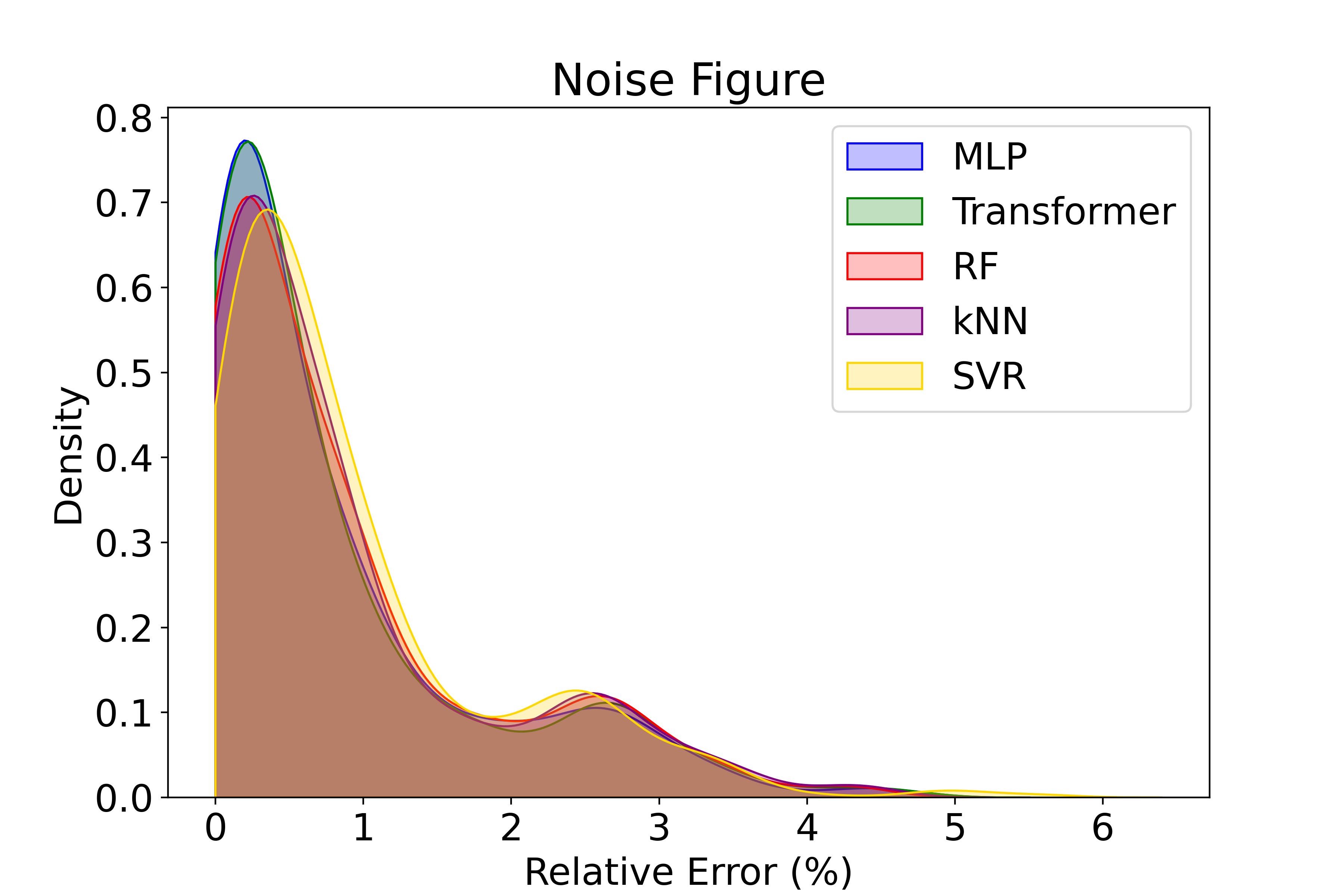}} 
    \subfloat{\includegraphics[width=0.24\textwidth]{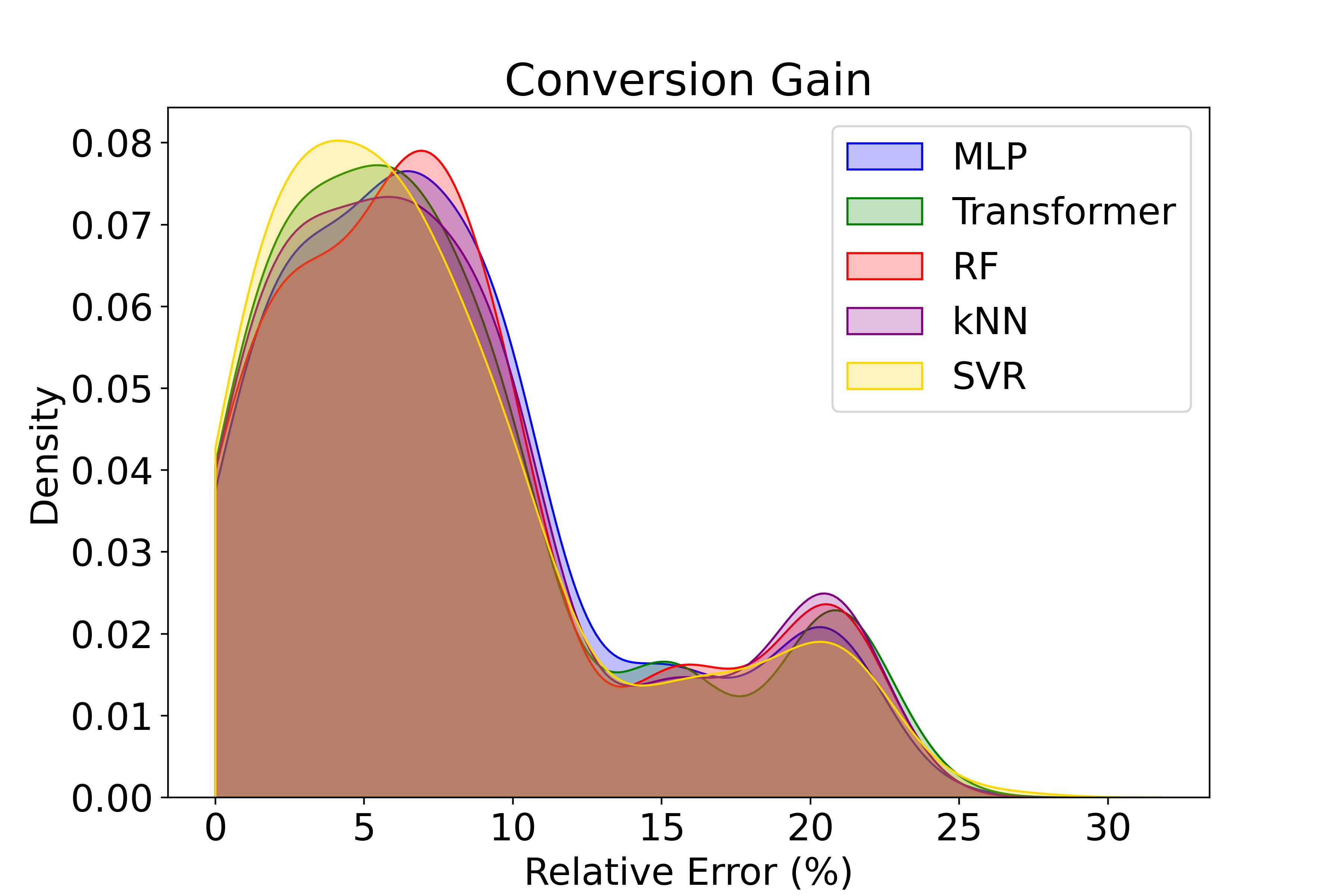}} \\ [-1.5ex]
    \subfloat{\includegraphics[width=0.24\textwidth]{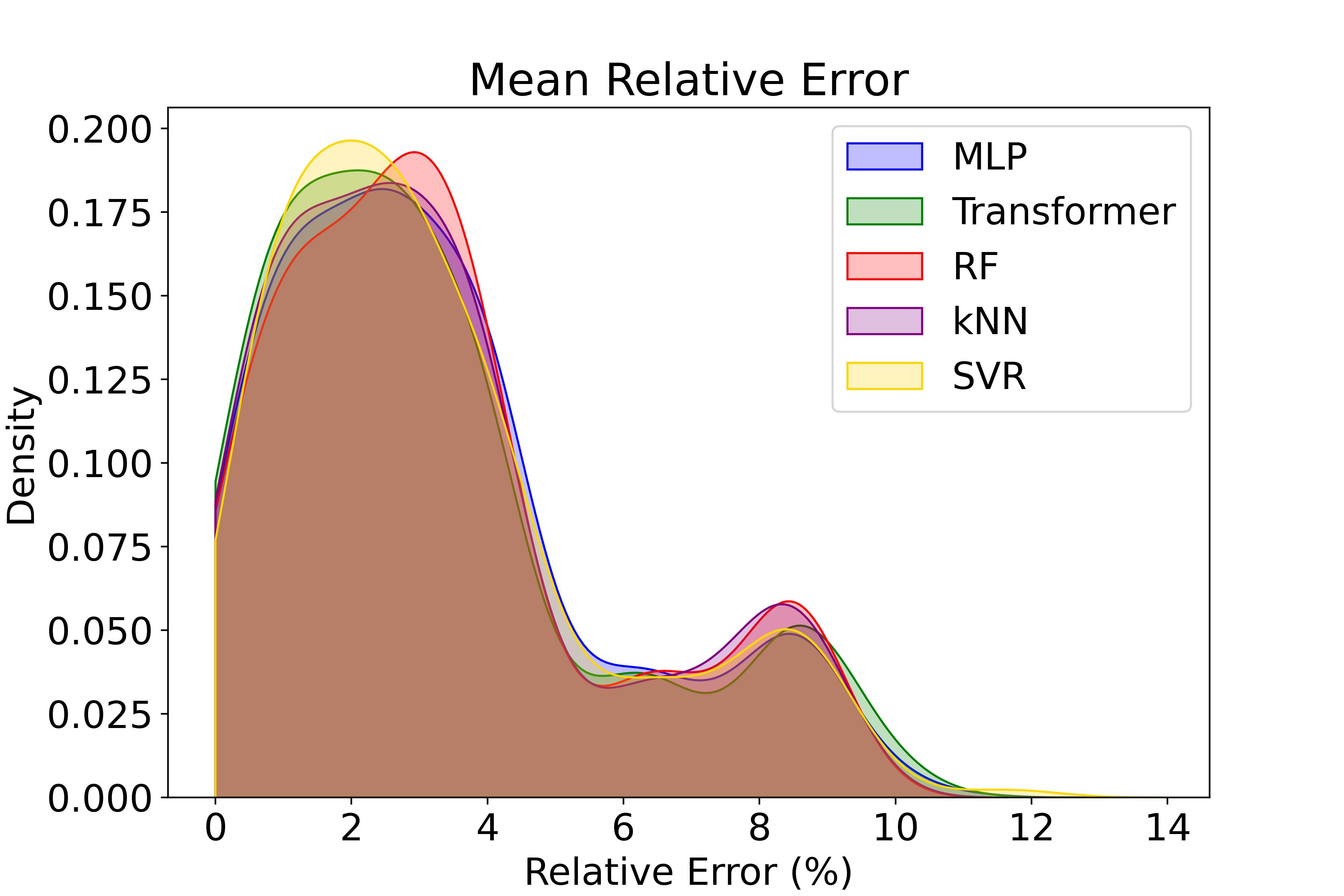}}
    \caption{\centering Relative error histogram of individual performance metrics and mean relative error histogram for Mixer.}
    \label{fig:mixer_plots}
\end{figure}

\textbf{Mixer.} The Mixer results, presented in Table~\ref{tab:mixer_summary} and Figure~\ref{fig:mixer_plots}, show that all models achieve comparable performance with minimal differences. The mean relative errors, as well as the P75 and P90 values, remain close across all methods, indicating that the Mixer system is less challenging for ML models to learn. While \textbf{MLP} and \textbf{Transformer} achieve slightly lower mean errors and standard deviations, traditional models such as \textbf{kNN}, \textbf{RF}, and \textbf{SVR} also perform consistently with no significant outliers. The relative error histograms further confirm that the predictions are tightly distributed across all metrics. Overall, the results highlight that the Mixer system is well-suited for ML-based parameter prediction, with all models delivering strong and reliable performance.

\vspace{-3mm}

\begin{table}[ht!]
\caption{Statistical summary of mean relative error for Mixer}
\label{tab:mixer_summary}
\centering
\renewcommand{\arraystretch}{0.8} 
\resizebox{0.48\textwidth}{!}{
\begin{tabular}{cccccccc}
\toprule
\raisebox{0.5em}{\textbf{Model}} & 
\raisebox{0.5em}{\textbf{Mean}} & 
\raisebox{0.5em}{\textbf{Std}} & 
\raisebox{0.5em}{\textbf{P75}} & 
\raisebox{0.5em}{\textbf{P90}} & 
\textbf{\shortstack{\% Errors \\ $<$ 2\% }} & 
\textbf{\shortstack{\% Errors \\ $<$ 5\% }} & 
\textbf{\shortstack{\% Outlier \\ ($>$ 20\%)}} \\

\midrule
\textbf{MLP} & 3.28 & \textbf{2.44} & 4.14 & \textbf{7.54} & 34.0 & \textbf{81.4} & \textbf{0.0} \\

\midrule
\textbf{Transformer} & \textbf{3.22} & 2.53 & 4.04 & 8.25 & \textbf{37.0} & 81.2 & \textbf{0.0} \\

\midrule
\textbf{RF} & 3.30 & 2.45 & \textbf{4.03} & 8.01 & 33.8 & 81.2 & \textbf{0.0} \\

\midrule
\textbf{kNN} & 3.27 & 2.46 & 4.09 & 7.91 & 34.2 & 81.2 & \textbf{0.0} \\
        
\midrule
\textbf{SVR} & 3.30 & 2.45 & 4.19 & 7.74 & 34.4 & 81.2 & \textbf{0.0} \\
\bottomrule
\end{tabular}
}
\end{table}

\begin{figure}[ht!]
    \centering
    \vspace{-3mm}
    \subfloat{\includegraphics[width=0.24\textwidth]{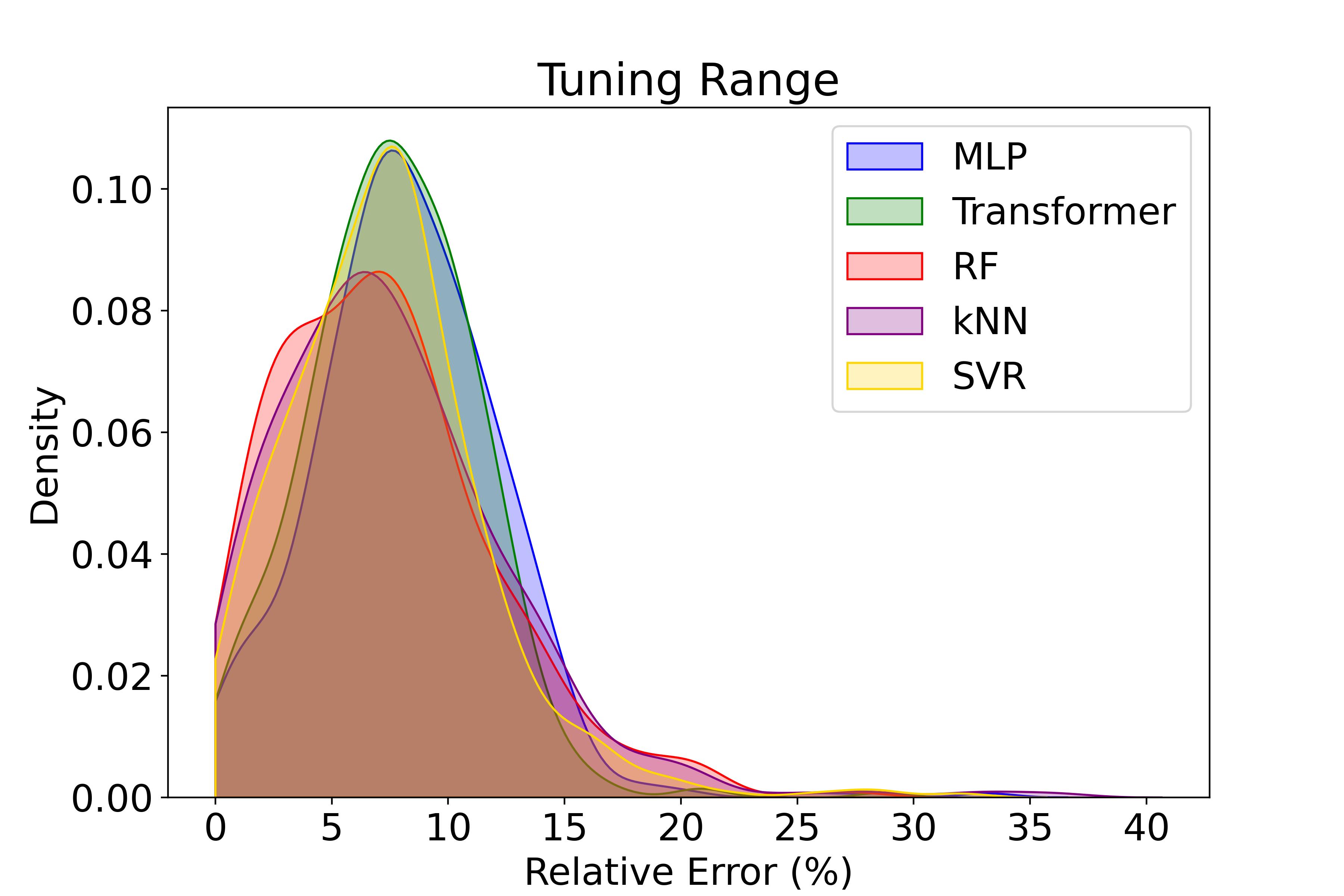}}
    \subfloat{\includegraphics[width=0.24\textwidth]{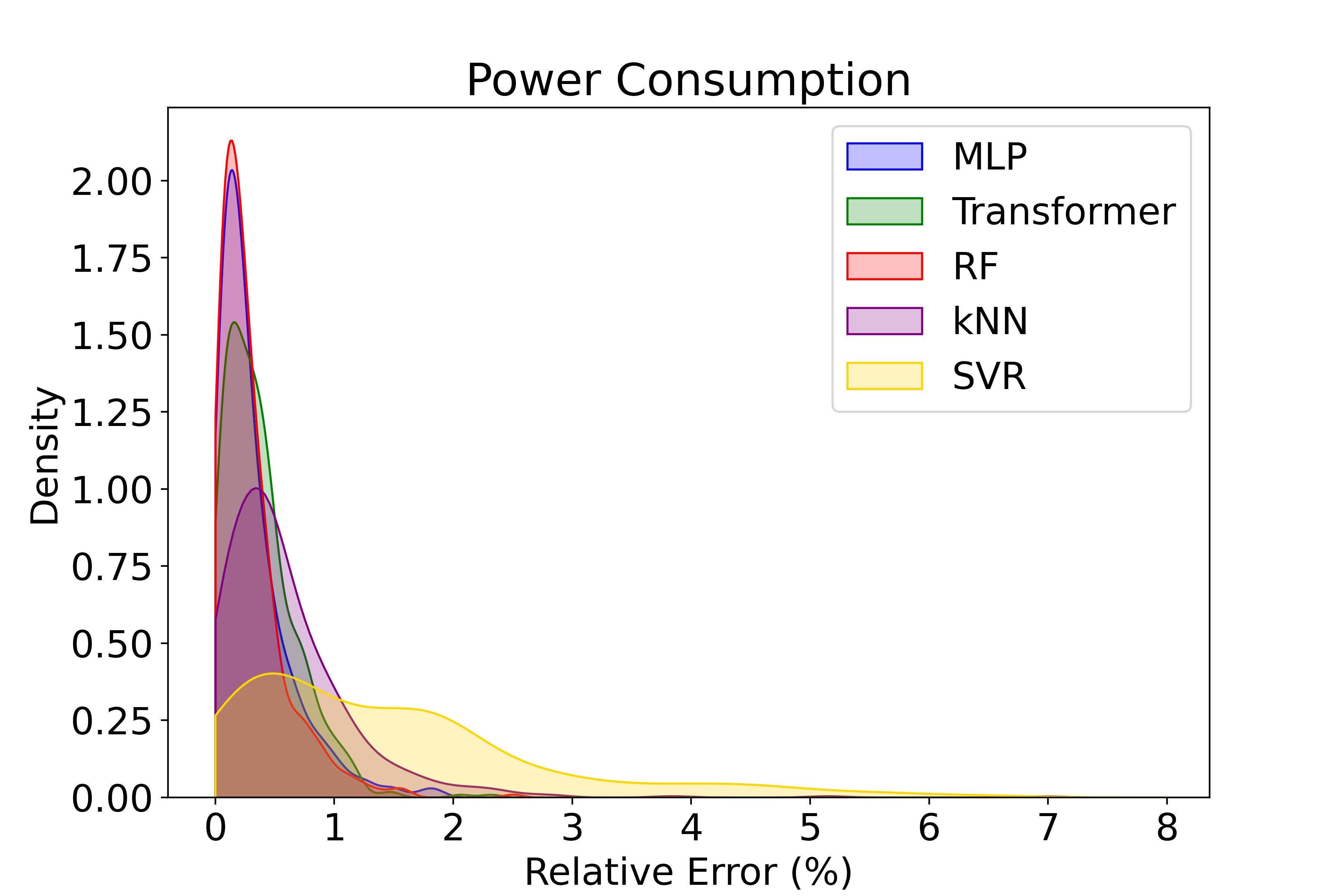}} \\ [-1.5ex]
    \subfloat{\includegraphics[width=0.24\textwidth]{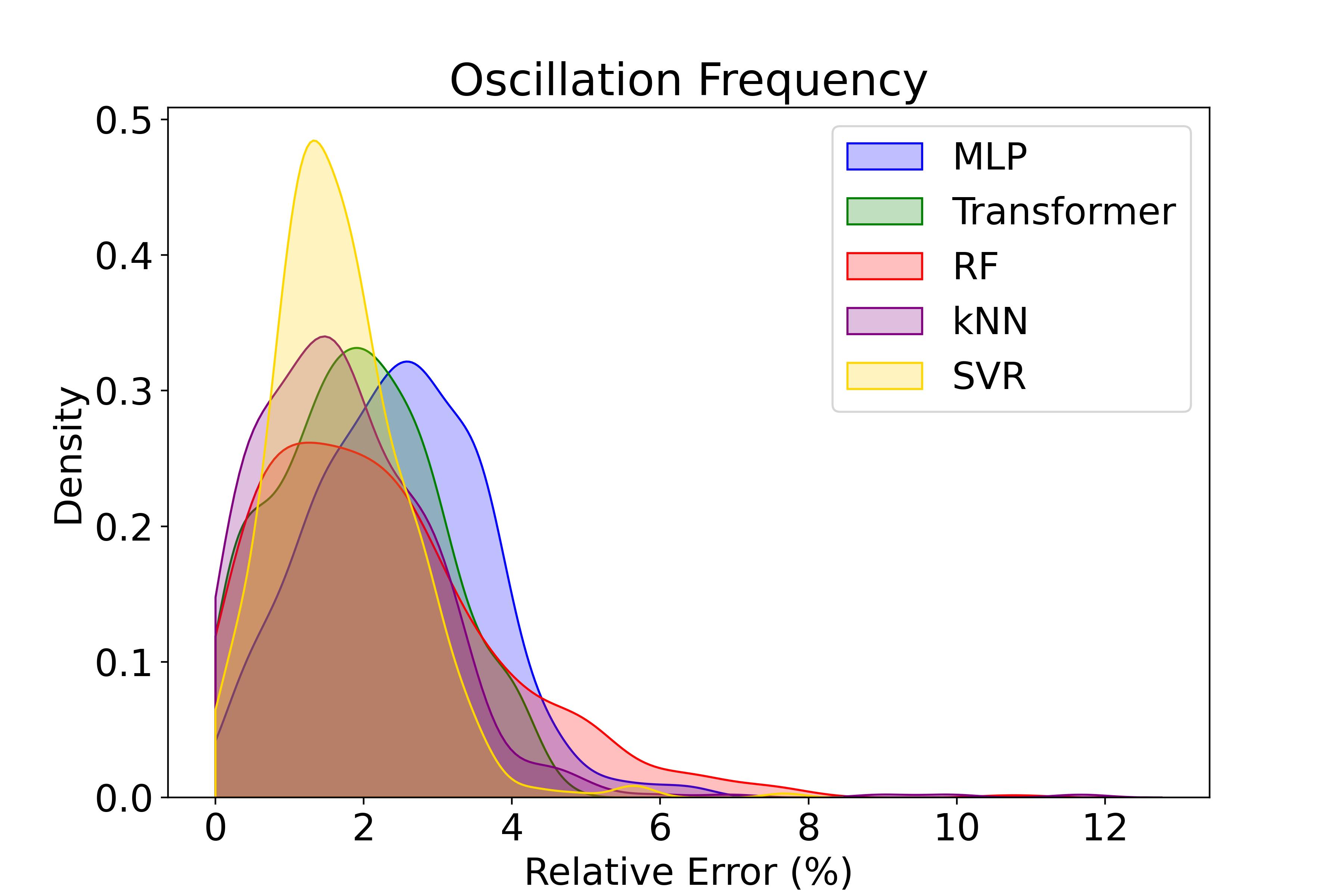}} 
    \subfloat{\includegraphics[width=0.24\textwidth]{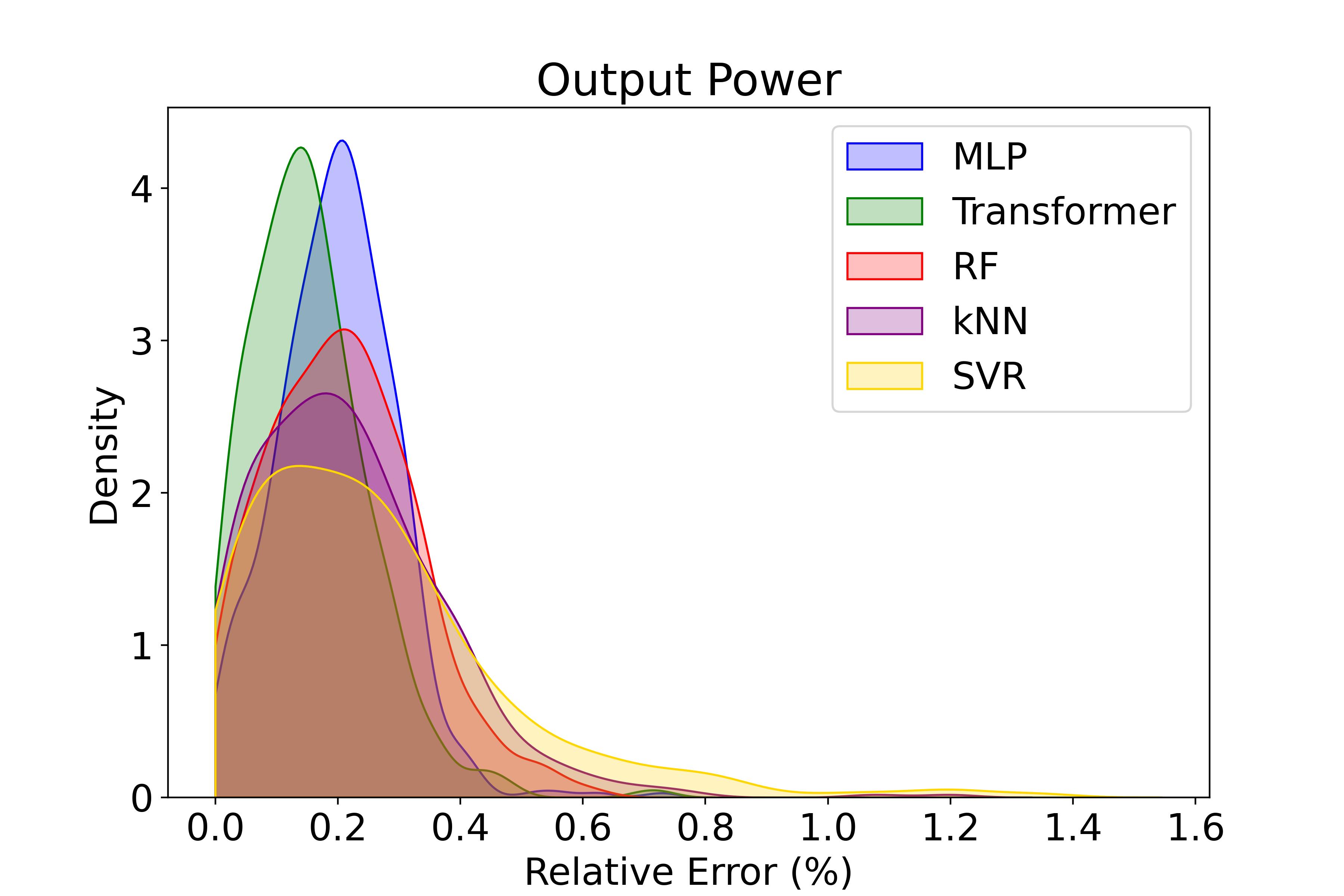}} \\ [-1.5ex]
    \subfloat{\includegraphics[width=0.24\textwidth]{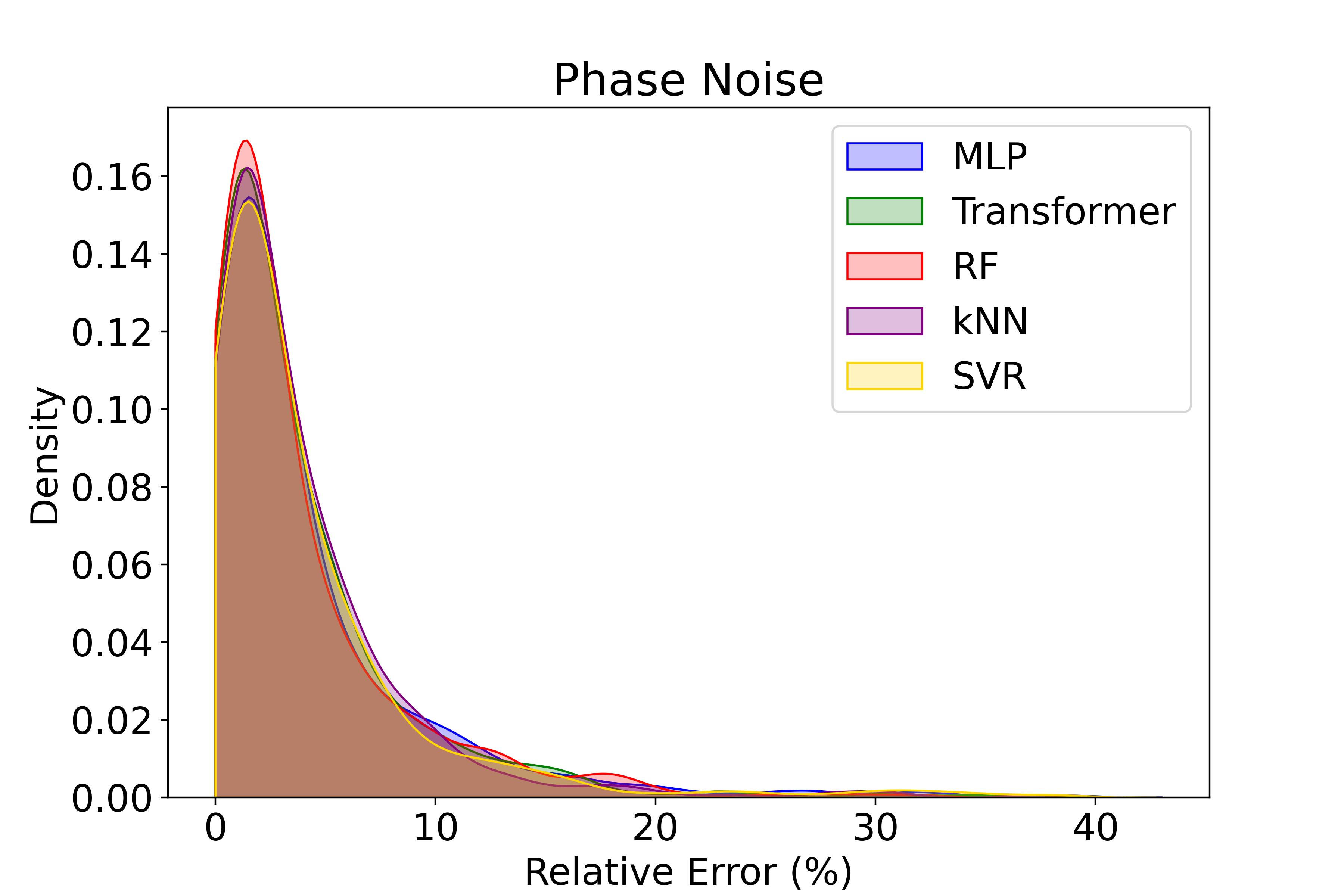}} 
    \subfloat{\includegraphics[width=0.24\textwidth]{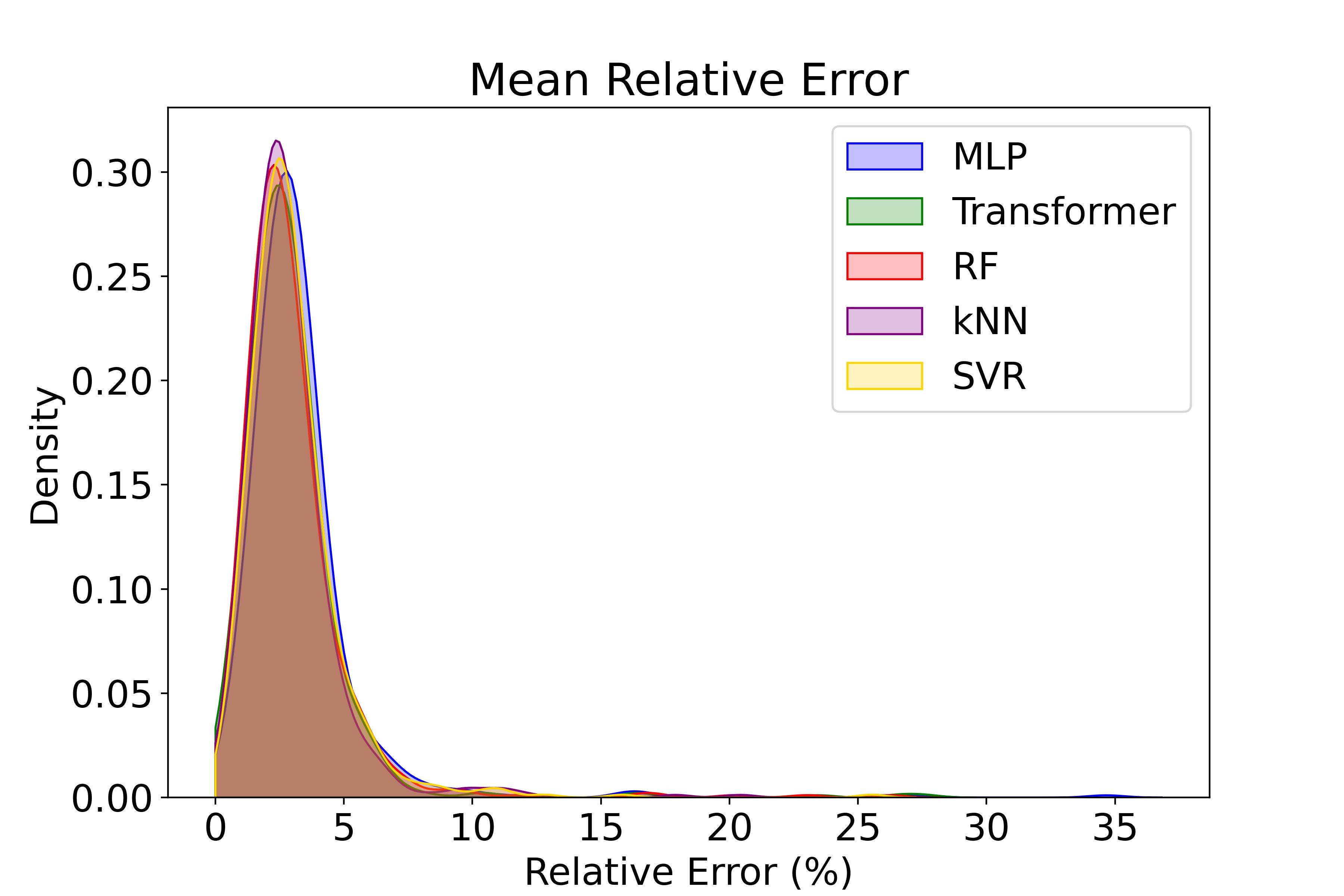}} 
    \caption{\centering Relative error histogram of individual performance metrics and mean relative error histogram for VCO.}
    \label{fig:vco_plots}
\end{figure}

\textbf{Voltage-Controlled Oscillator (VCO).}
The results for VCO, summarized in Table~\ref{tab:vco_summary} and Figure~\ref{fig:vco_plots}, reveal comparable performance across all models. Traditional methods such as \textbf{RF} and \textbf{kNN} deliver slightly better results, achieving lower standard deviations and competitive error percentiles. Notably, \textbf{SVR} achieves the lowest outlier rate, highlighting its consistency despite relatively higher mean errors.
Among neural networks, the \textbf{MLP} and \textbf{Transformer} models provide robust performance, with the \textbf{Transformer} achieving fewer errors below 5\% compared to MLP. However, the broader error spread across all models indicates the complexity of accurately predicting VCO performance metrics, such as phase noise and oscillation frequency. Overall, traditional methods exhibit a slight edge for VCO, emphasizing their ability to handle this circuit's parameter-performance trade-offs effectively.

\vspace{-3mm}

\begin{table}[ht!]
\caption{Statistical summary of mean relative error for VCO}
\label{tab:vco_summary}
\centering
\renewcommand{\arraystretch}{0.8} 
\resizebox{0.48\textwidth}{!}{
\begin{tabular}{cccccccc}
\toprule
\raisebox{0.5em}{\textbf{Model}} & 
\raisebox{0.5em}{\textbf{Mean}} & 
\raisebox{0.5em}{\textbf{Std}} & 
\raisebox{0.5em}{\textbf{P75}} & 
\raisebox{0.5em}{\textbf{P90}} & 
\textbf{\shortstack{\% Errors \\ $<$ 2\% }} & 
\textbf{\shortstack{\% Errors \\ $<$ 5\% }} & 
\textbf{\shortstack{\% Outlier \\ ($>$ 20\%)}} \\

\midrule
\textbf{MLP} & 7.70 & 57.13 & 3.71 & 5.02 & 18.8 & 89.8 & 1.8 \\

\midrule
\textbf{Transformer} & 7.62 & 60.10 & 3.52 & 4.85 & 30.2 & 90.6 & 2.0 \\

\midrule
\textbf{RF} & \textbf{6.95} & \textbf{52.06} & 3.43 & 5.24 & \textbf{30.6} & 89.4 & 2.0 \\

\midrule
\textbf{kNN} & 7.11 & 52.99 & \textbf{3.41} & \textbf{4.75} & 30.2 & \textbf{91.8} & 1.8 \\
        
\midrule
\textbf{SVR} & 7.42 & 58.73 & 3.62 & 5.09 & 26.8 & 89.4 & \textbf{1.6} \\
\bottomrule
\end{tabular}
}
\end{table}

\begin{figure}[ht!]
    \centering
    \vspace{-6mm}
    \subfloat{\includegraphics[width=0.24\textwidth]{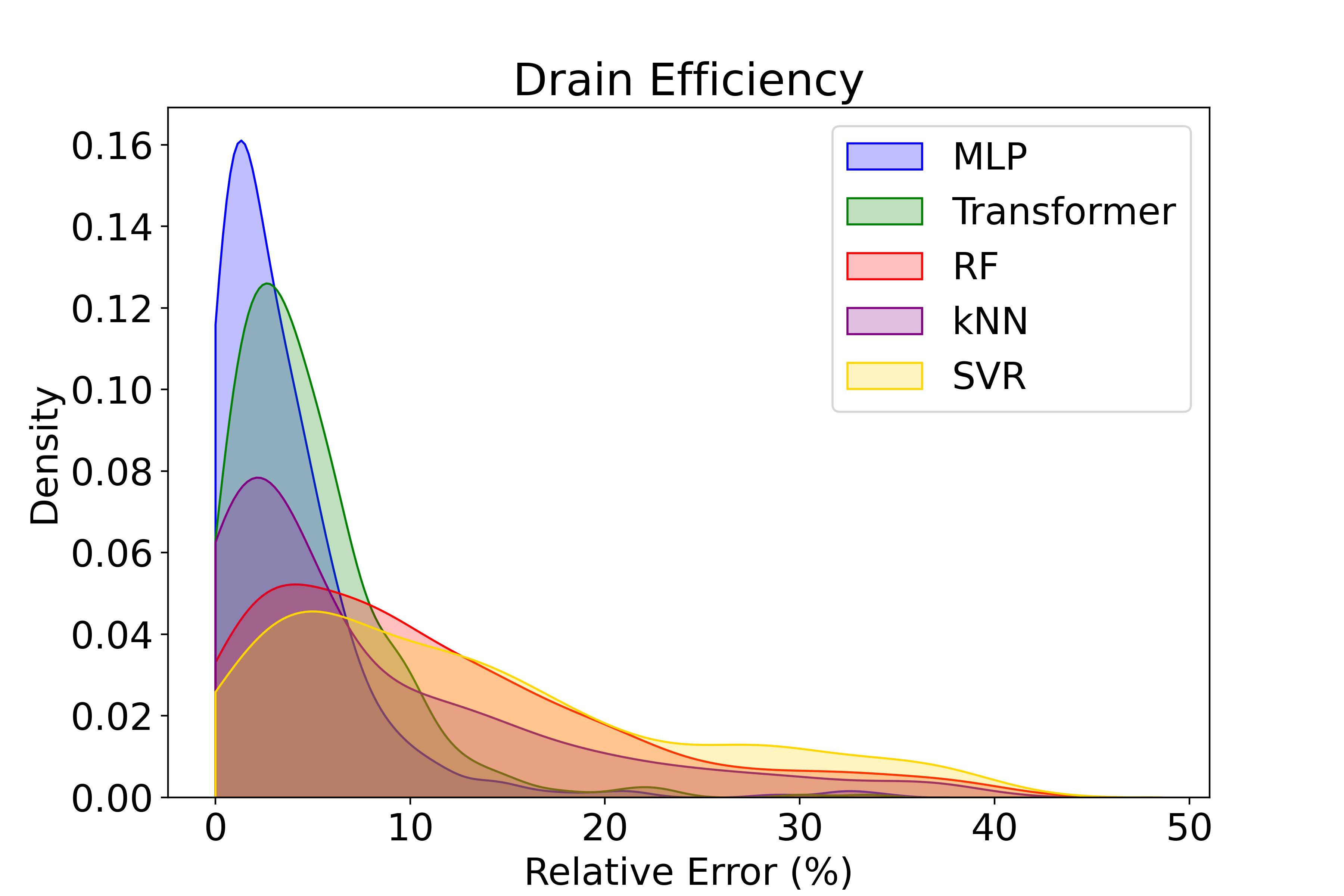}}
    \subfloat{\includegraphics[width=0.24\textwidth]{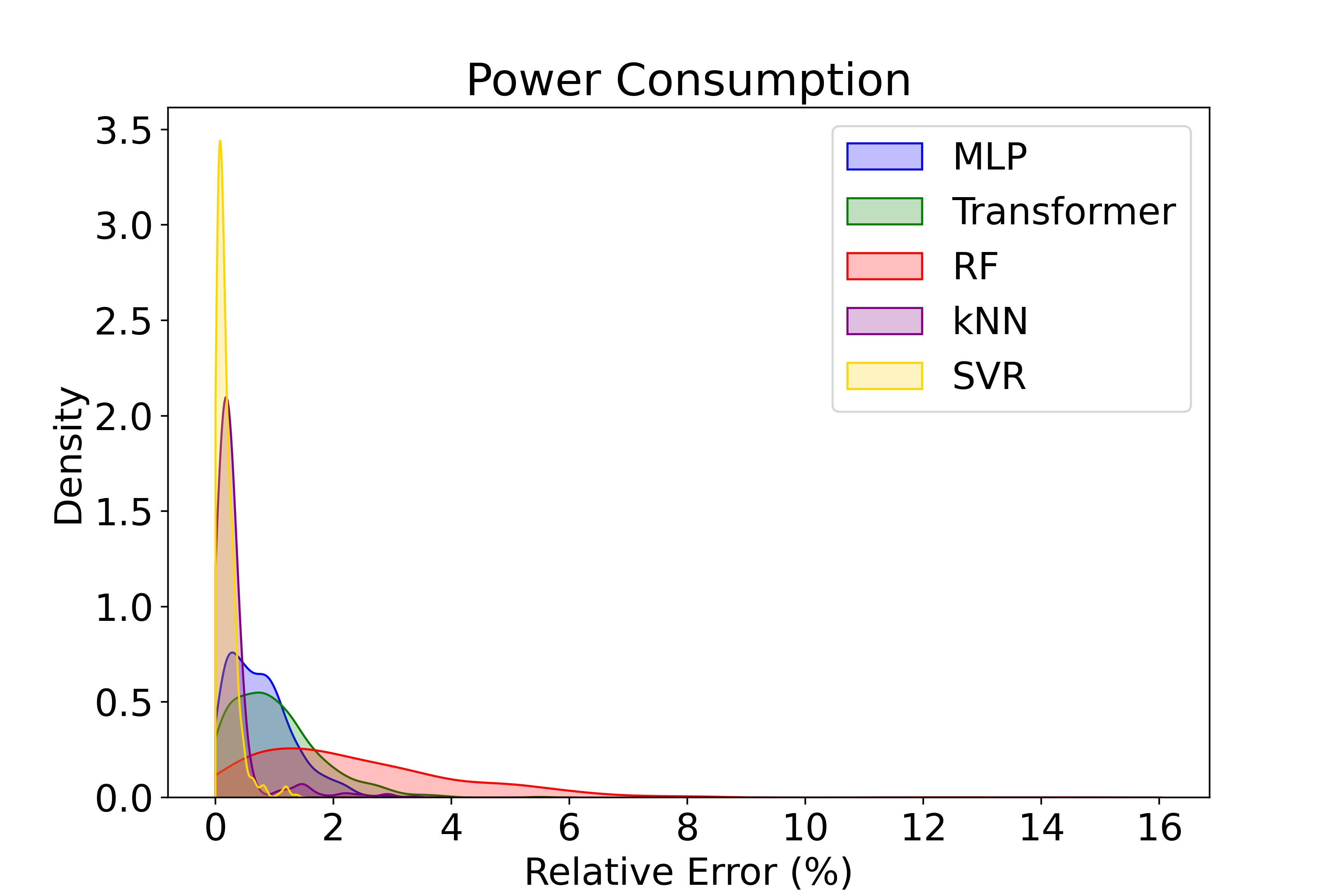}} \\ [-1.5ex]
    \subfloat{\includegraphics[width=0.24\textwidth]{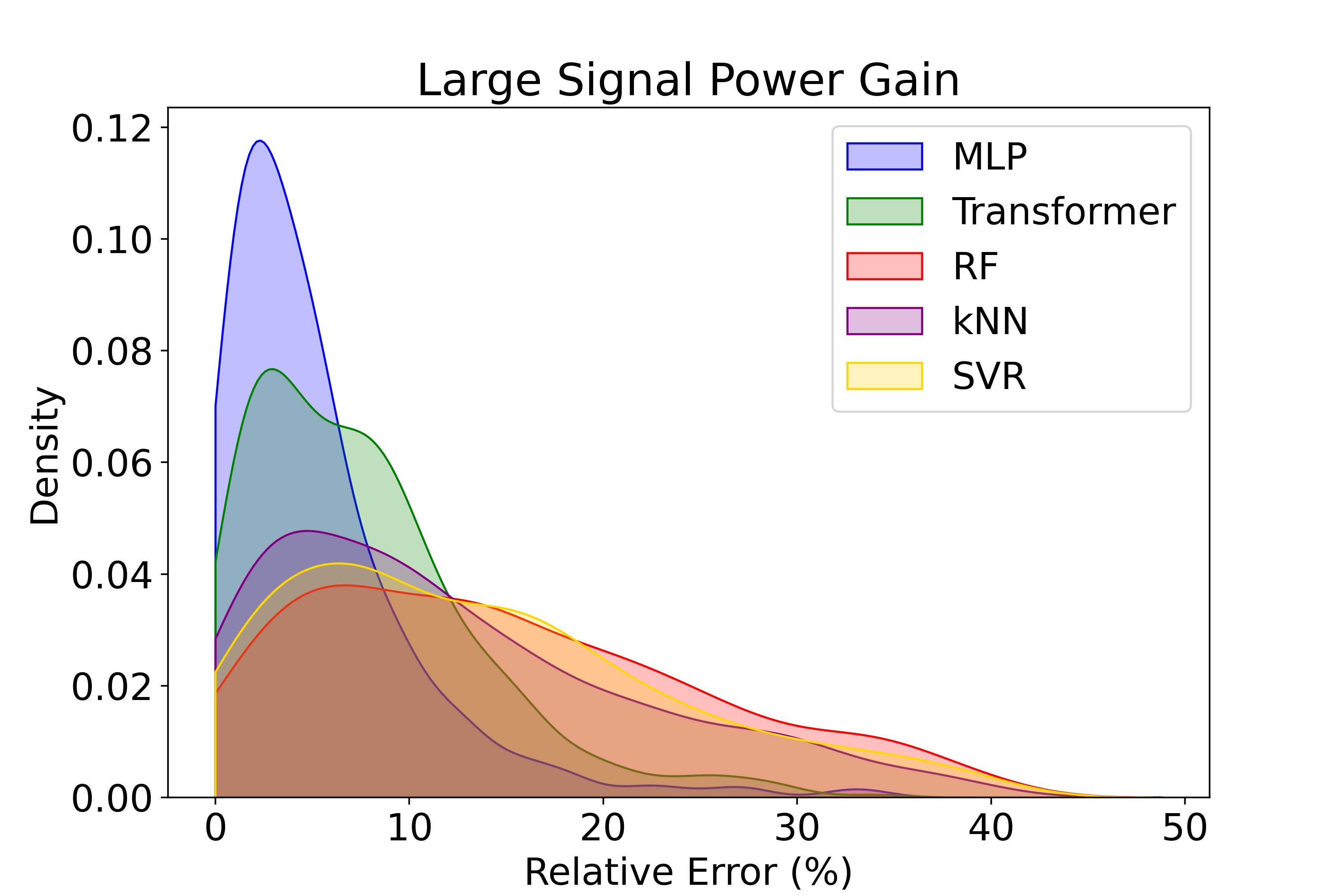}} 
    \subfloat{\includegraphics[width=0.24\textwidth]{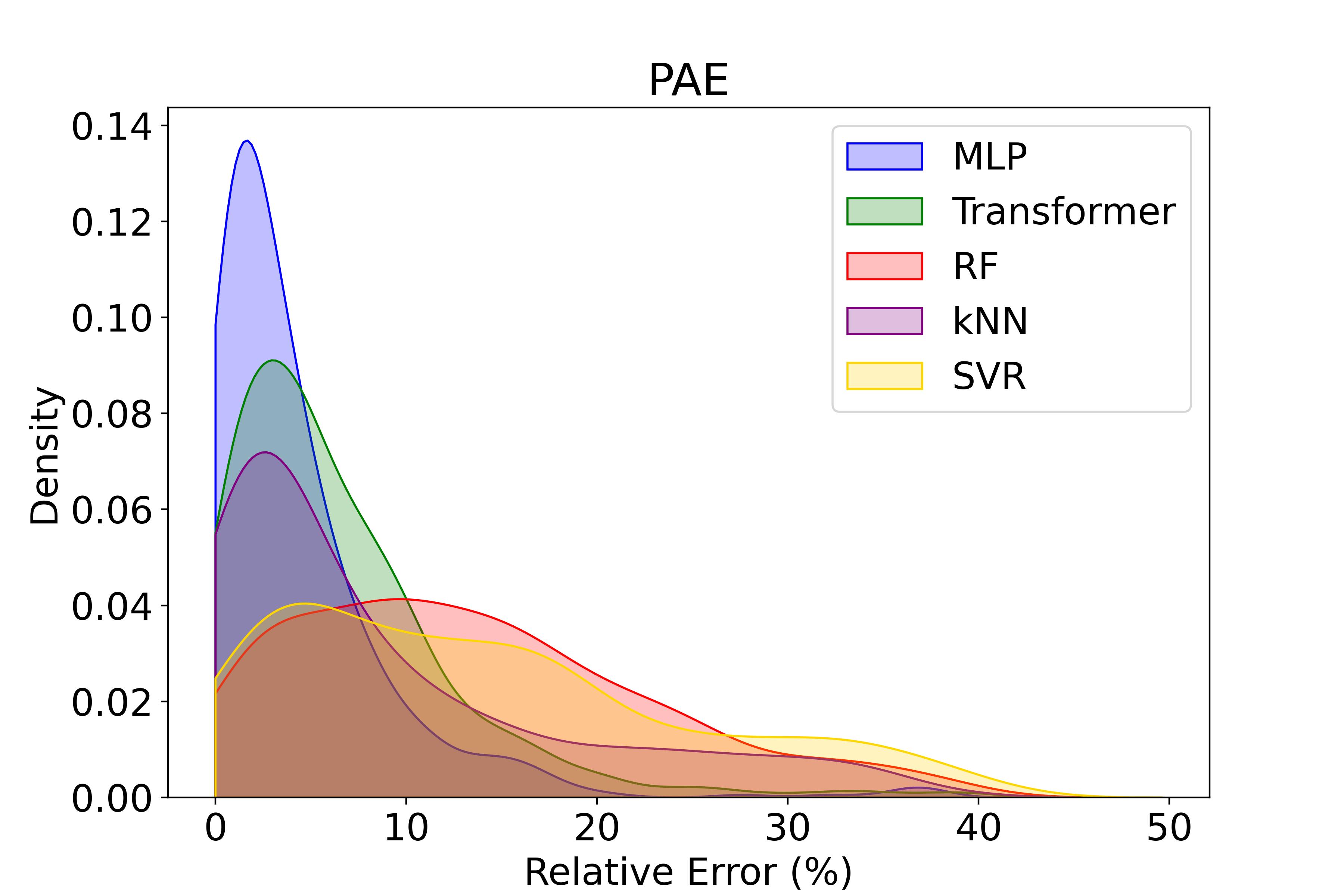}} \\ [-1.5ex]
    \subfloat{\includegraphics[width=0.24\textwidth]{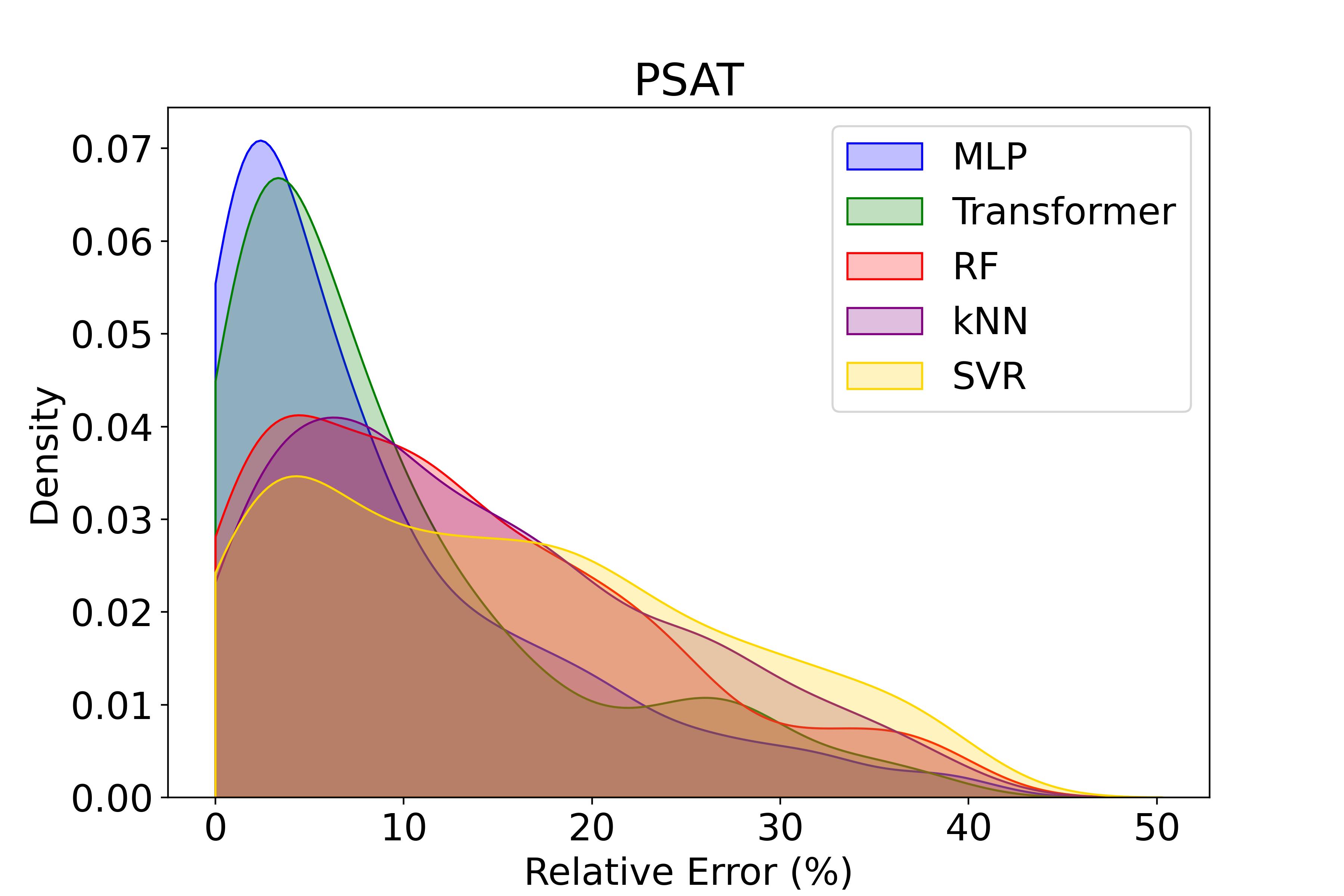}}
    \subfloat{\includegraphics[width=0.24\textwidth]{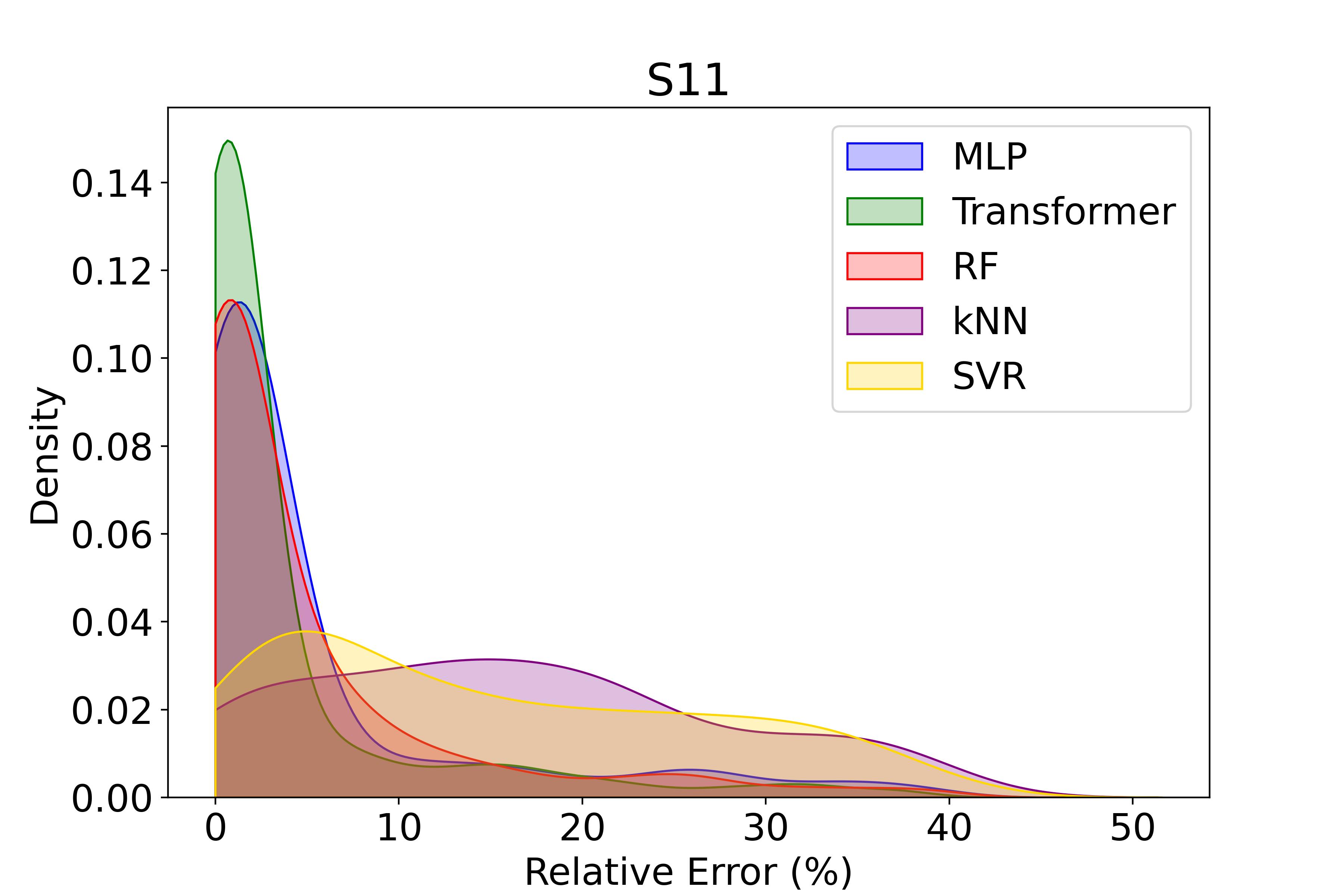}} \\ [-1.5ex]
    \subfloat{\includegraphics[width=0.24\textwidth]{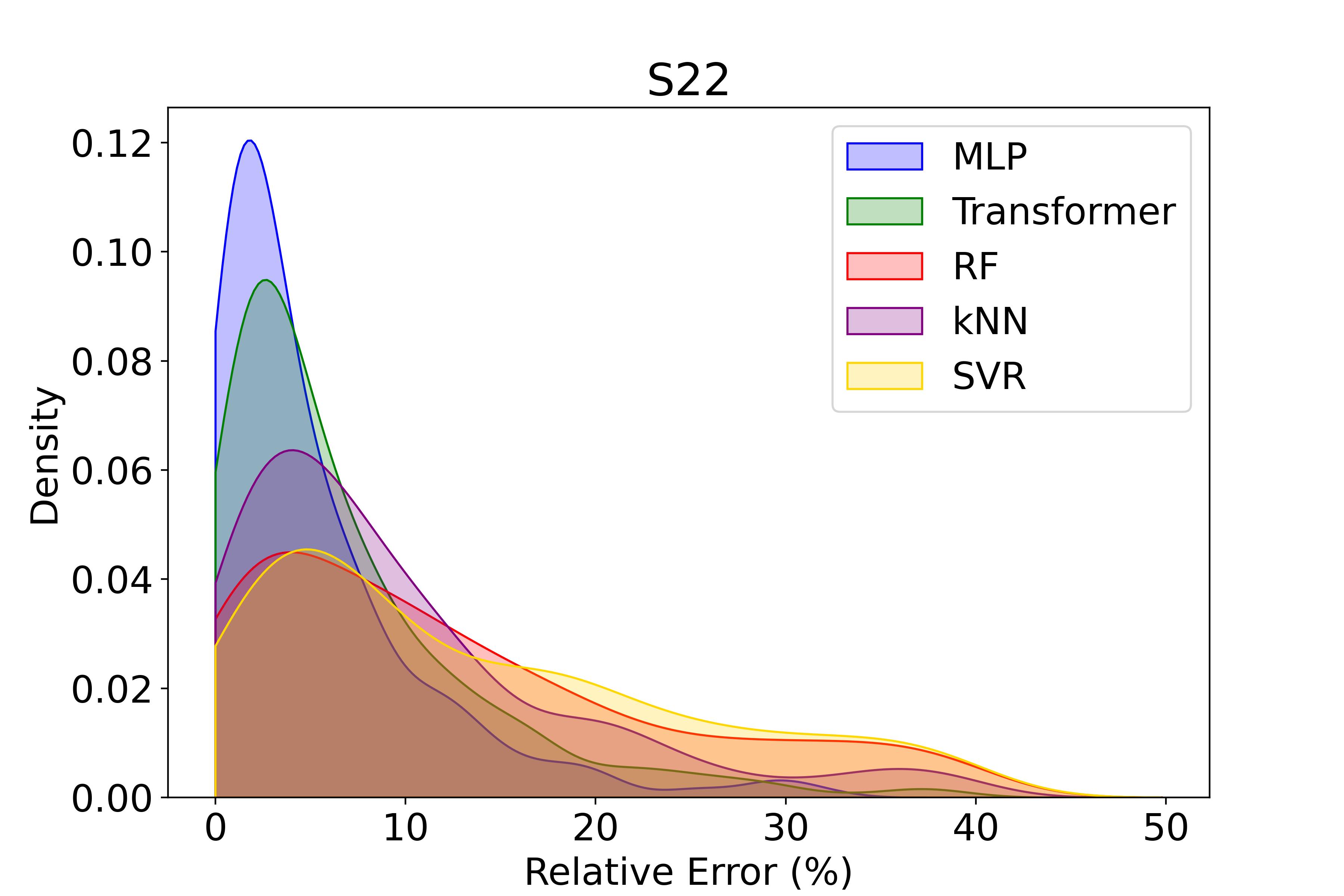}}
    \subfloat{\includegraphics[width=0.24\textwidth]{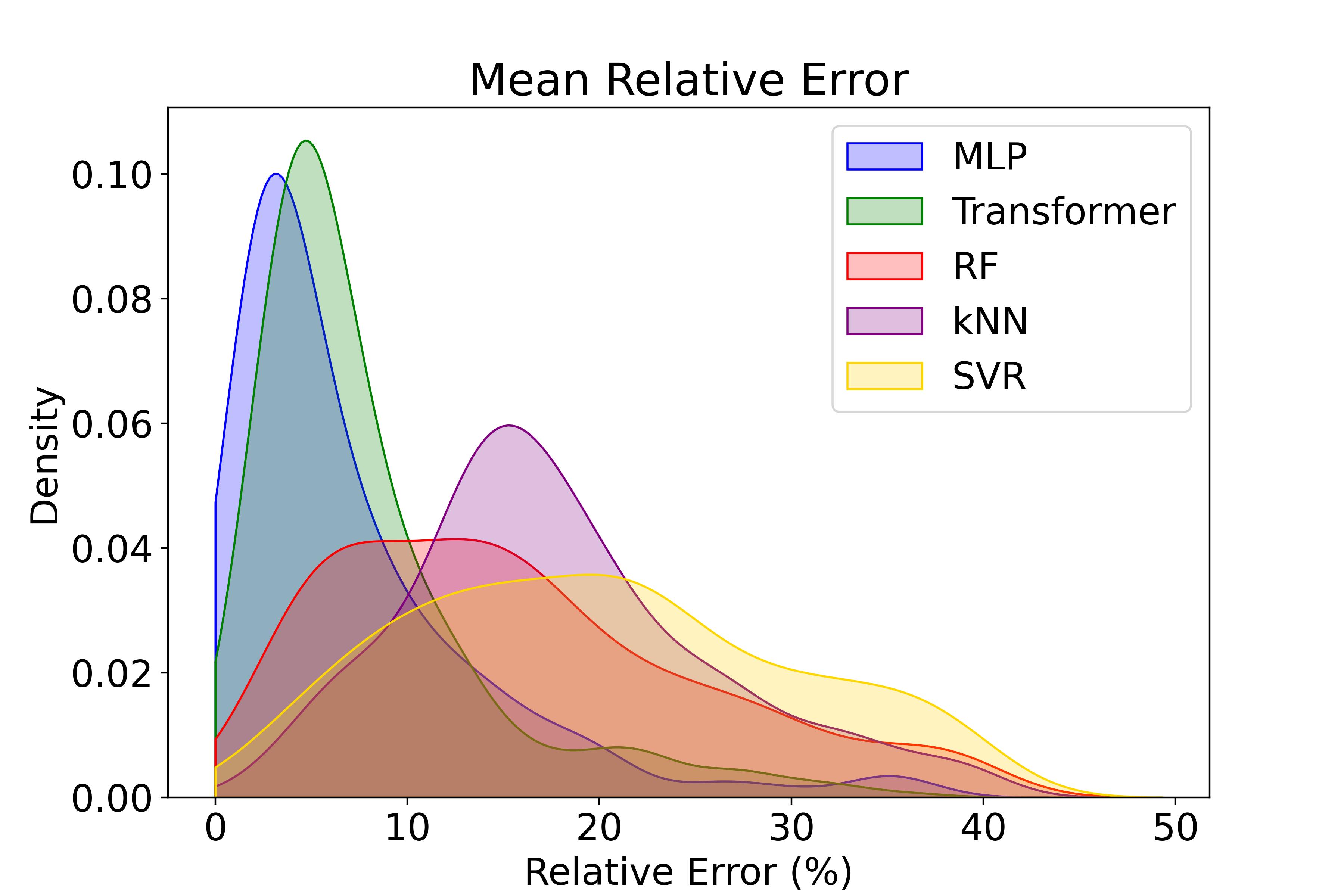}}
    \caption{\centering Relative error histogram of individual performance metrics and mean relative error histogram for PA.}
    \label{fig:pa_plots}
\end{figure}

\textbf{Power Amplifier (PA).}
The results for PA, summarized in Table~\ref{tab:pa_summary} and Figure~\ref{fig:pa_plots}, reveal significant challenges in accurately predicting the performance metrics. \textbf{MLP} achieves the lowest mean relative error and shows a relatively balanced error distribution, making it the most consistent model for PA among all tested approaches. The \textbf{Transformer}, while demonstrating some promise, struggles with higher standard deviations and a larger error spread. Traditional methods such as \textbf{RF} and \textbf{kNN} exhibit inconsistent performance, with kNN producing notable outliers and high P90 values. Notably, \textbf{SVR} performs poorly, yielding large mean errors and the highest outlier percentage, highlighting its inability to generalize well for PA's complex performance metrics.

The substantial error spread across all models indicates the nonlinearity and intricate trade-offs inherent in PA design, such as achieving high drain efficiency, PAE, and power gain simultaneously. This complexity underscores the difficulty of modeling PA behavior compared to other circuits.

\vspace{-4mm}

\begin{table}[ht!]
\caption{Statistical summary of mean relative error for PA}
\label{tab:pa_summary}
\centering
\renewcommand{\arraystretch}{0.8} 
\resizebox{0.48\textwidth}{!}{
\begin{tabular}{cccccccc}
\toprule
\raisebox{0.5em}{\textbf{Model}} & 
\raisebox{0.5em}{\textbf{Mean}} & 
\raisebox{0.5em}{\textbf{Std}} & 
\raisebox{0.5em}{\textbf{P75}} & 
\raisebox{0.5em}{\textbf{P90}} & 
\textbf{\shortstack{\% Errors \\ $<$ 2\% }} & 
\textbf{\shortstack{\% Errors \\ $<$ 5\% }} & 
\textbf{\shortstack{\% Outlier \\ ($>$ 20\%)}} \\

\midrule
\textbf{MLP} & \textbf{19.98} & 144.99 & \textbf{10.08} & \textbf{19.33} & \textbf{17.2} & \textbf{51.4} & \textbf{9.2} \\

\midrule
\textbf{Transformer} & 22.39 & 203.36 & 10.63 & 21.37 & 5.4 & 38.2 & 11.6 \\

\midrule
\textbf{RF} & 66.27 & 316.86 & 42.11 & 96.97 & 0.2 & 8.0 & 47.4 \\

\midrule
\textbf{kNN} & 37.70 & \textbf{98.20} & 29.23 & 63.31 & 0.0 & 2.0 & 43.8 \\
        
\midrule
\textbf{SVR} & 140.85 & 987.05 & 64.30 & 137.62 & 0.2 & 3.2 & 67.6 \\
\bottomrule
\end{tabular}
}
\end{table}


\subsection{Heterogeneous Circuits}

\begin{figure*}[ht]
    \centering
    \vspace{-6mm}
    \subfloat{\includegraphics[width=0.2\textwidth]{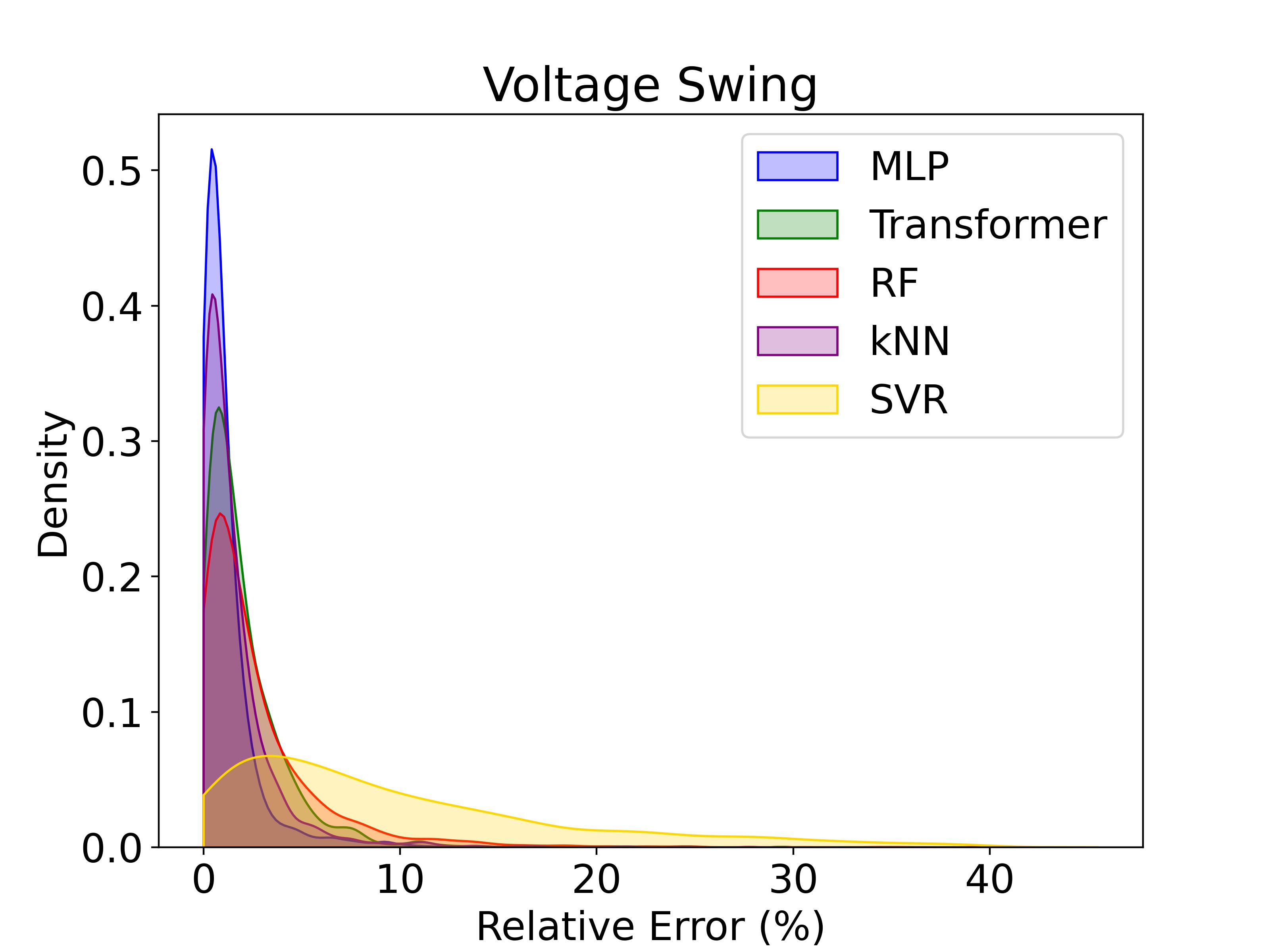}}
    \subfloat{\includegraphics[width=0.2\textwidth]{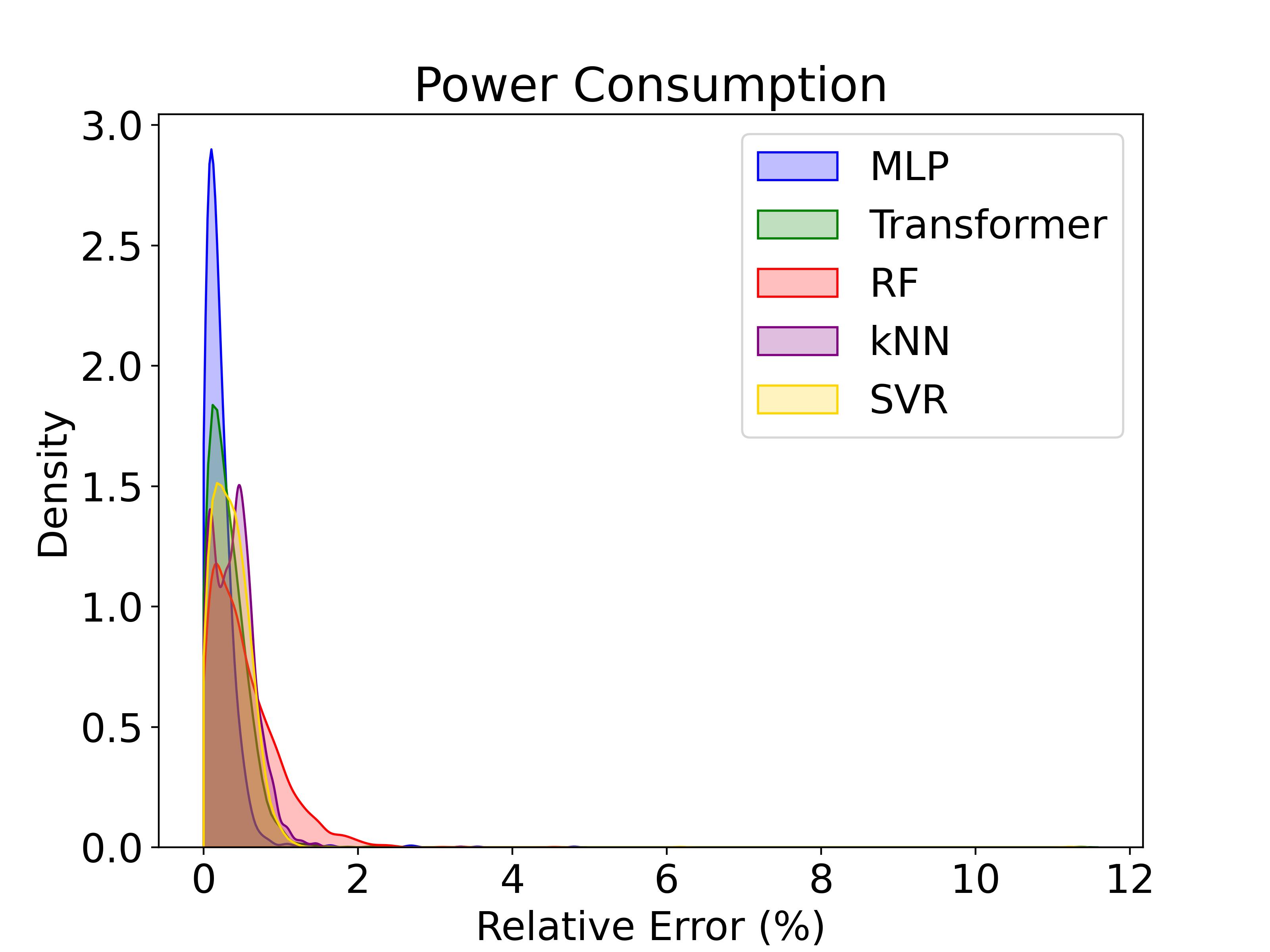}} 
    \subfloat{\includegraphics[width=0.2\textwidth]{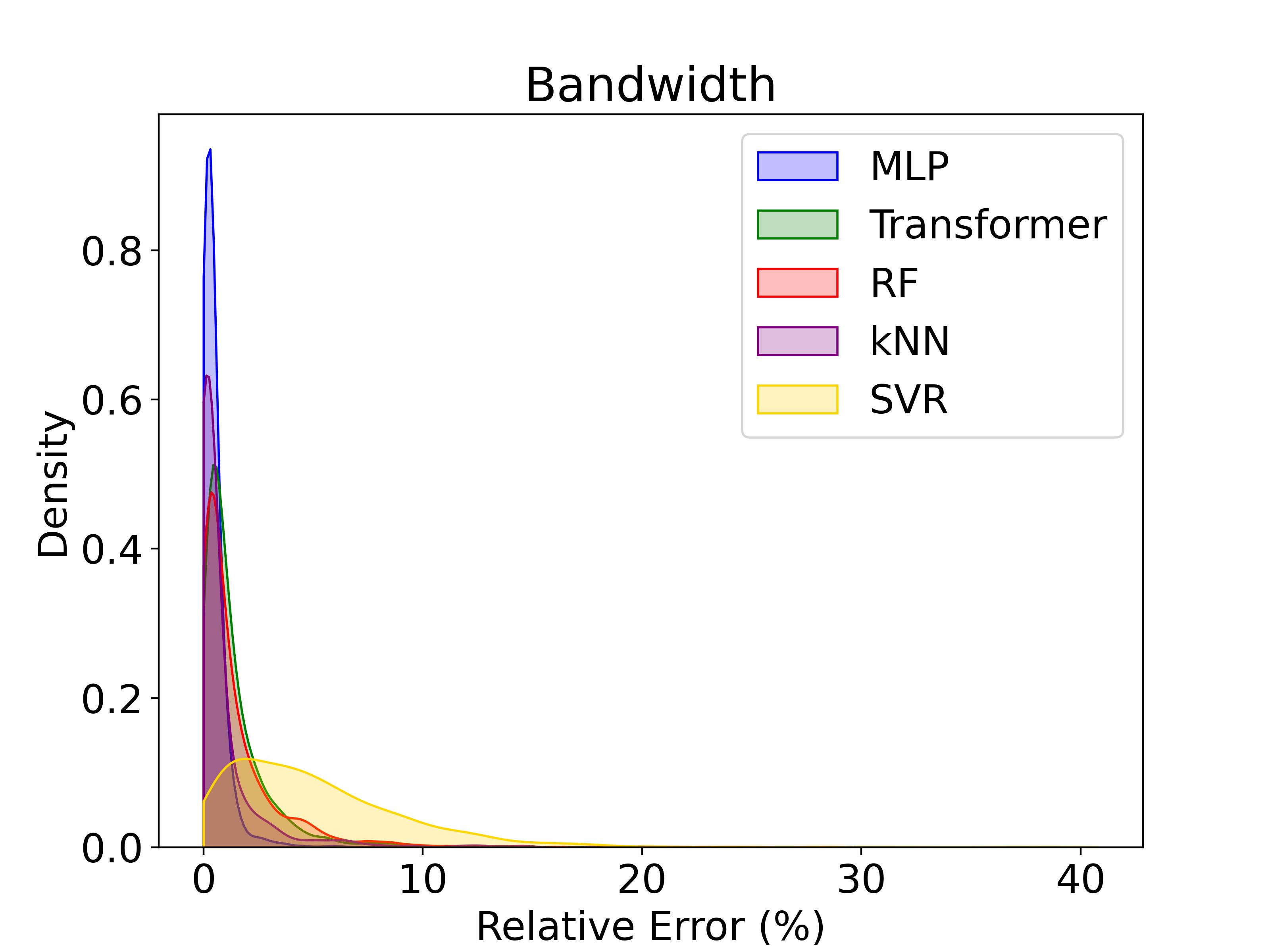}} 
    \subfloat{\includegraphics[width=0.2\textwidth]{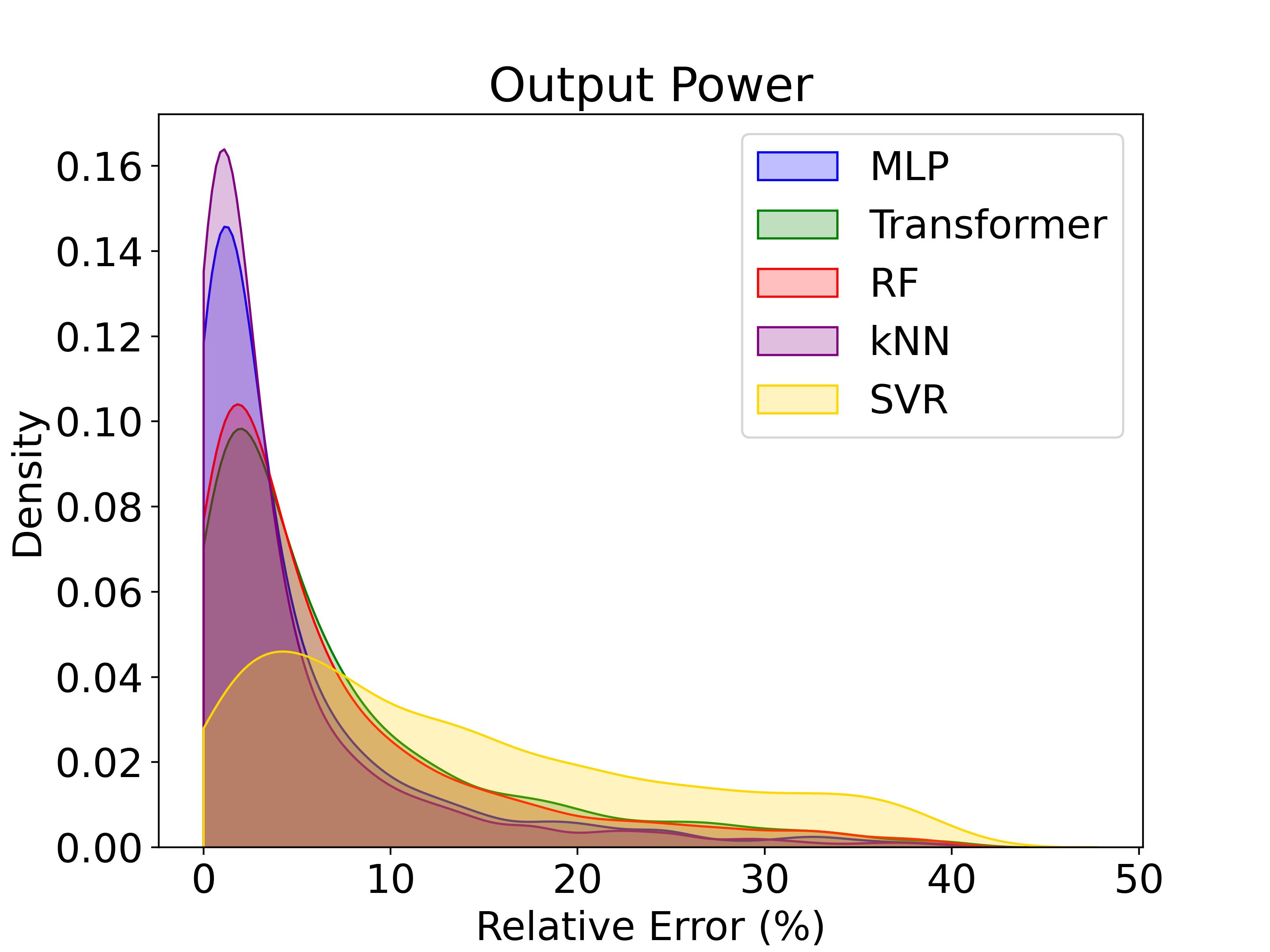}}
    \subfloat{\includegraphics[width=0.2\textwidth]{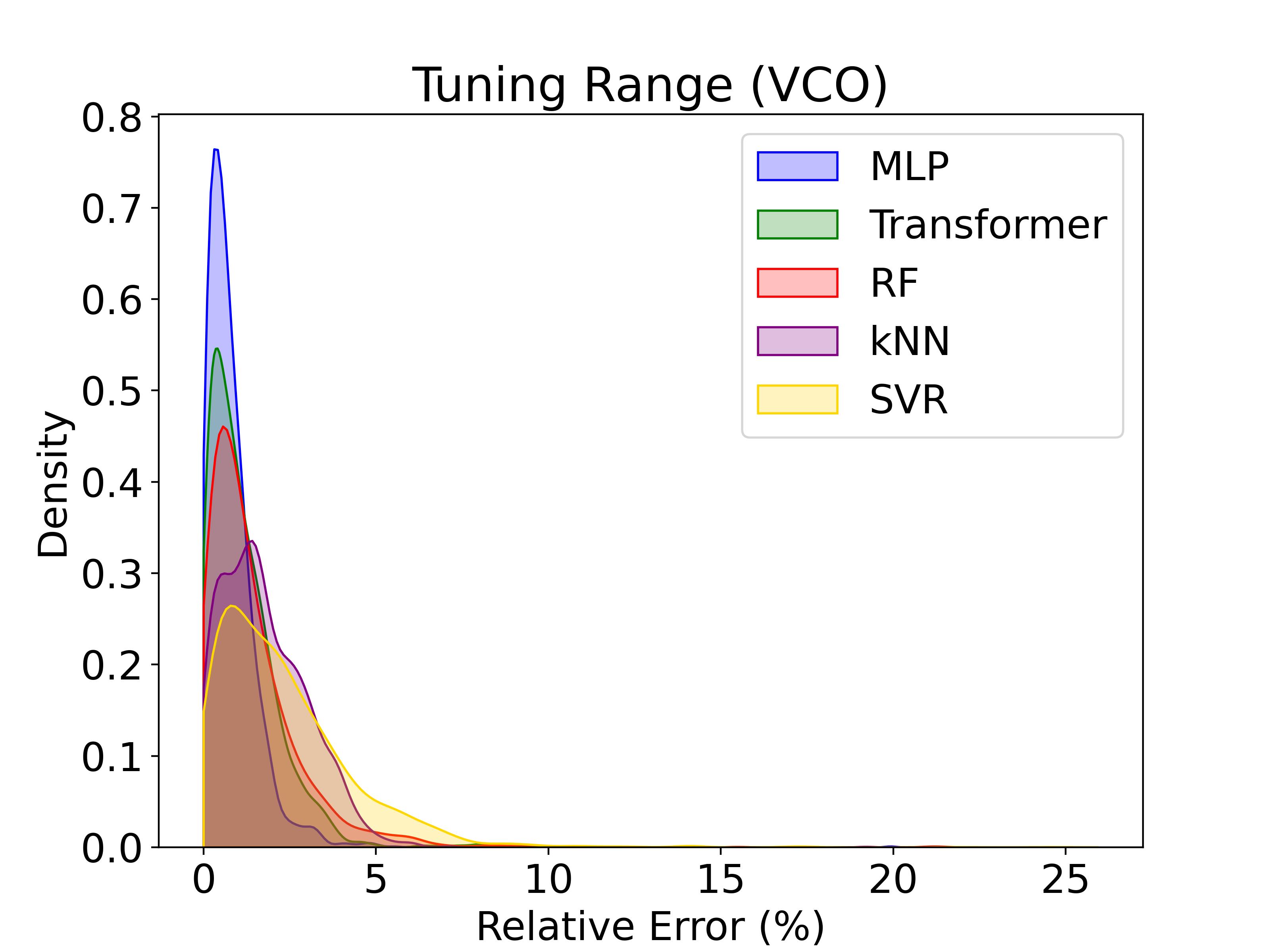}} \\ [-1.5ex]
    \subfloat{\includegraphics[width=0.2\textwidth]{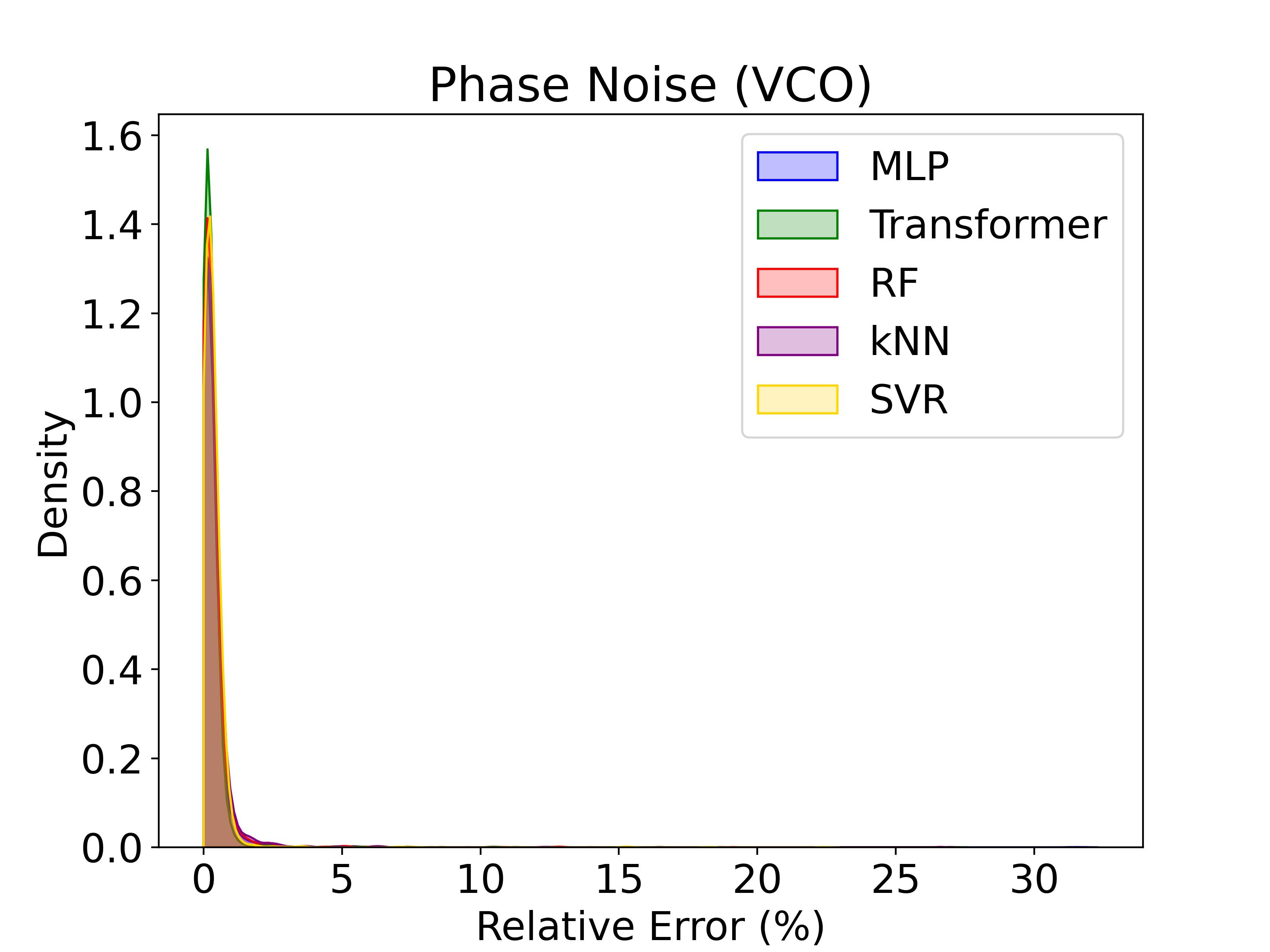}} 
    \subfloat{\includegraphics[width=0.2\textwidth]{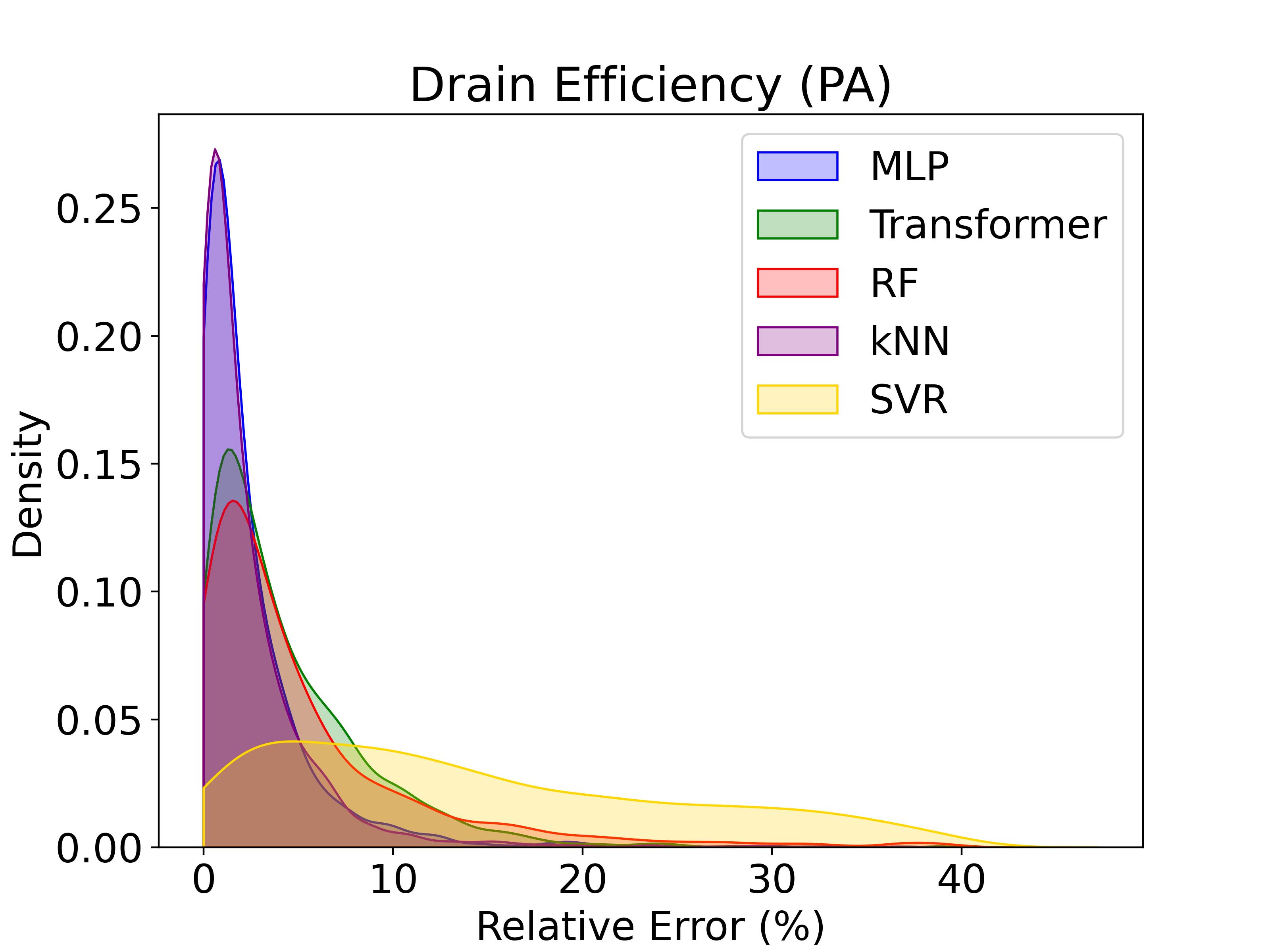}} 
    \subfloat{\includegraphics[width=0.2\textwidth]{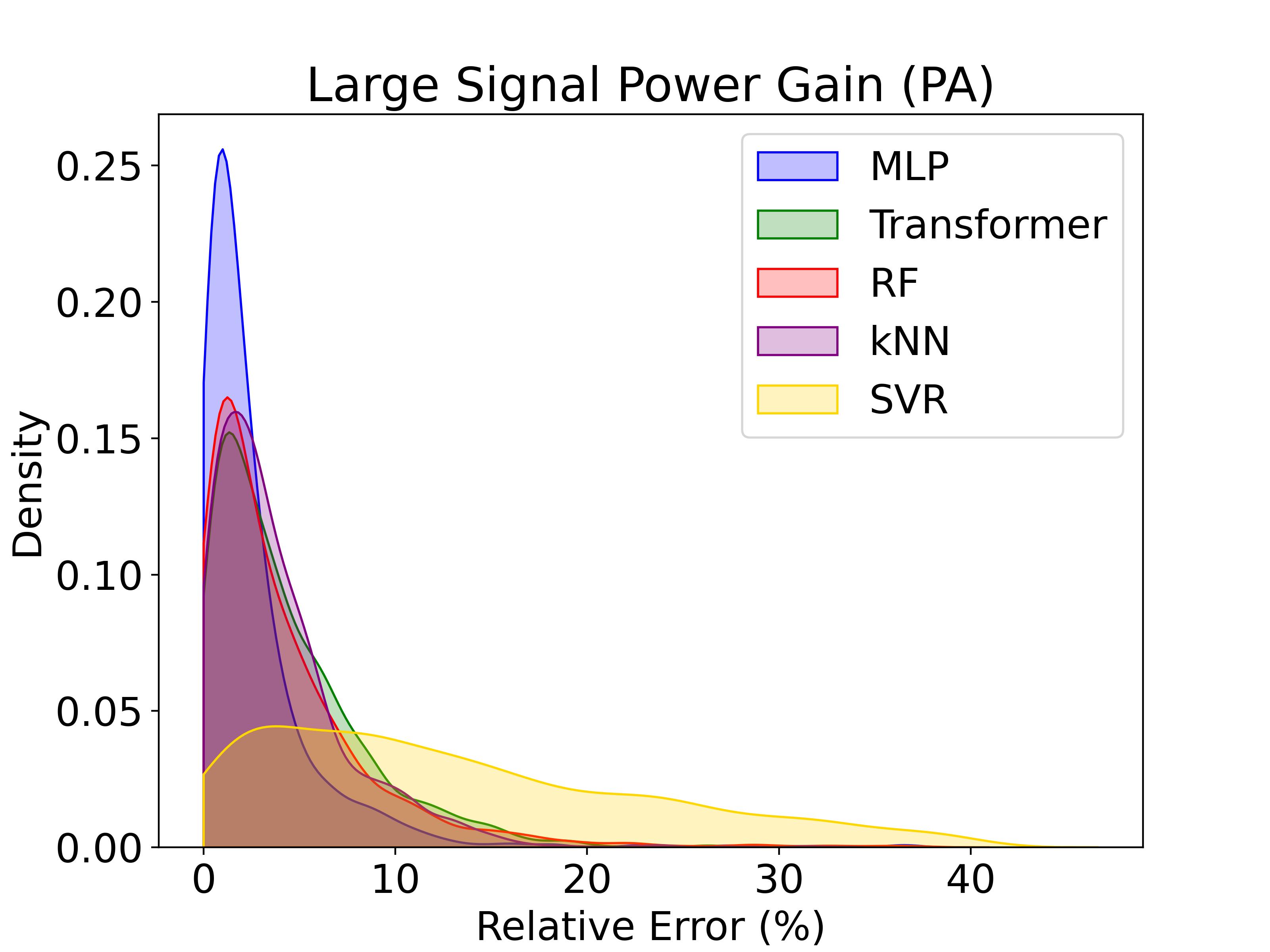}}
    \subfloat{\includegraphics[width=0.2\textwidth]{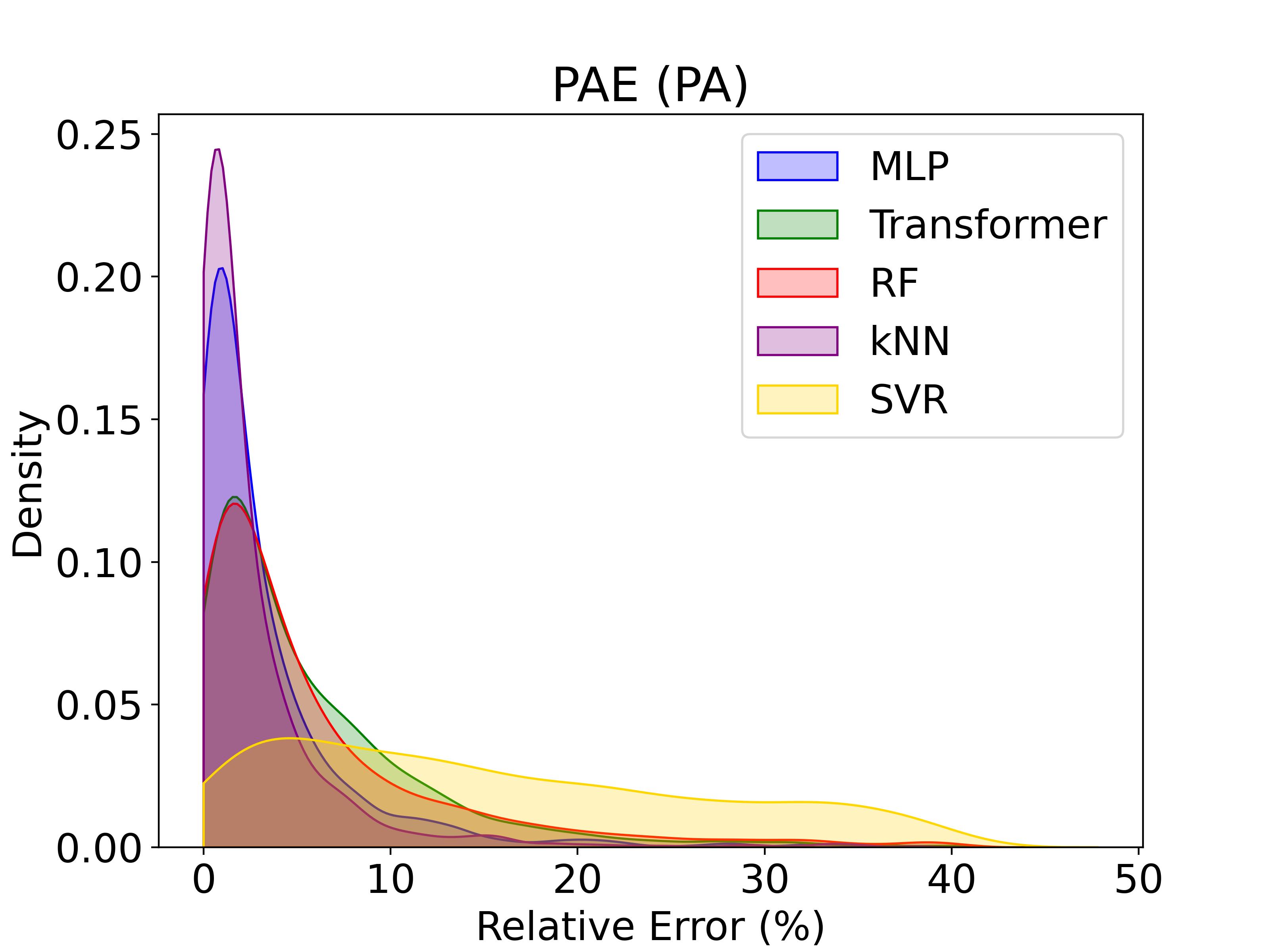}} 
    \subfloat{\includegraphics[width=0.2\textwidth]{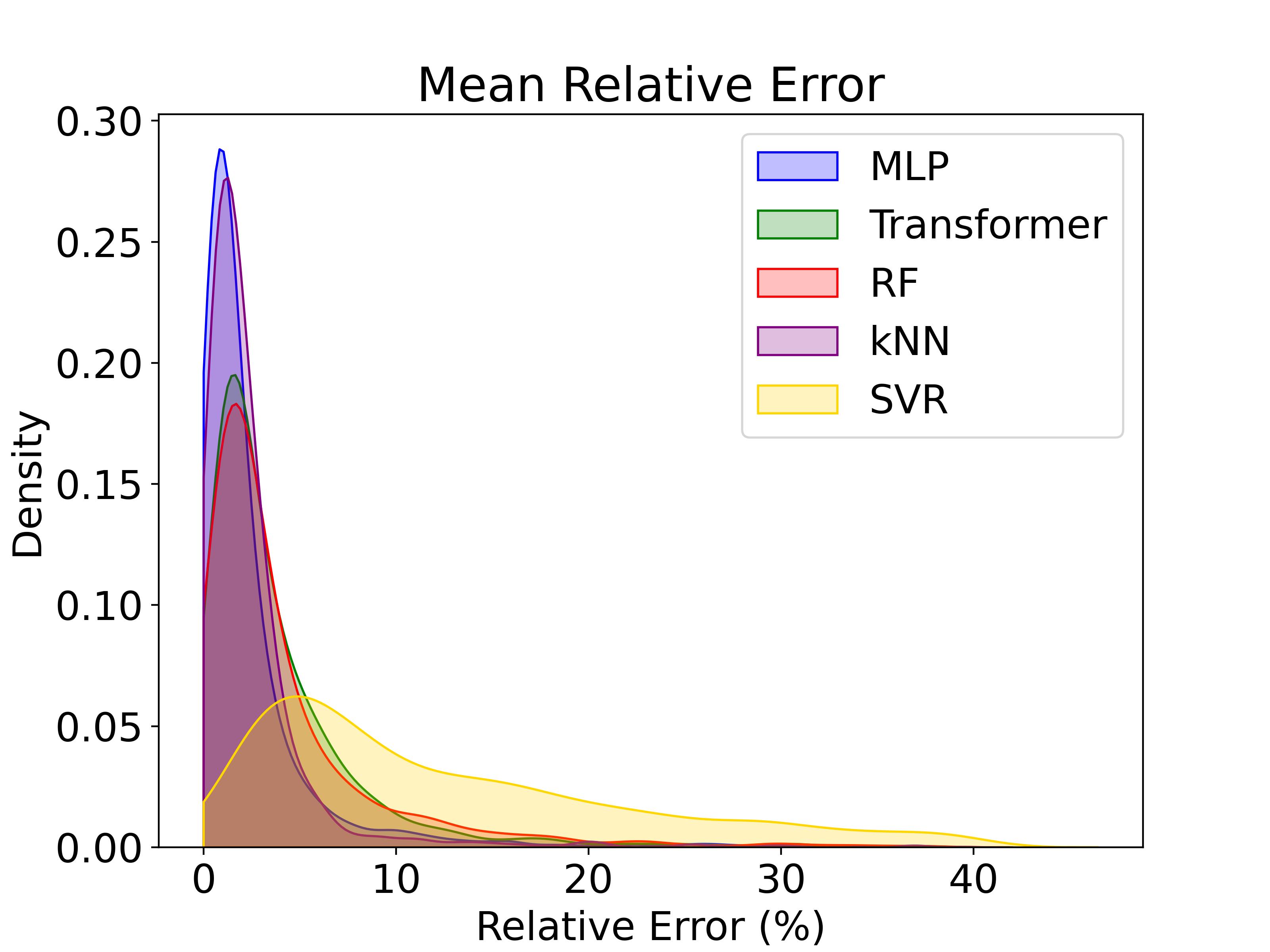}}
    \caption{\centering Relative error histogram of individual performance metrics and mean relative error histogram for Transmitter. The plots illustrate the relative error distributions for system performance metrics, such as power consumption, along with block-specific metrics like phase noise for VCO.}
    \label{fig:transmitter_plots}
\end{figure*}

\textbf{Transmitter.} The results for the Transmitter, shown in Table~\ref{tab:transmitter_summary} and Figure~\ref{fig:transmitter_plots}, highlight challenges in predicting performance metrics for complex systems. \textbf{MLP} achieves the lowest mean relative error and consistent accuracy across metrics. However, it shows slightly higher outlier errors compared to \textbf{kNN}. The \textbf{kNN} demonstrates strong performance in limiting large errors, achieving the lowest outlier percentage and competitive results in error percentiles. In contrast, the \textbf{Transformer} and \textbf{RF} models show increased error variability, with higher error spreads. \textbf{SVR} struggles significantly, exhibiting the largest mean error and the highest percentage of outliers, emphasizing its limitations in handling the Transmitter's complexity.
The results demonstrate that both \textbf{MLP} and \textbf{kNN} are effective for this circuit type, with \textbf{MLP} excelling in overall mean performance and \textbf{kNN} minimizing large errors.

\begin{figure*}[!htb]
    \centering
    \vspace{-4mm}
    \subfloat{\includegraphics[width=0.2\textwidth]{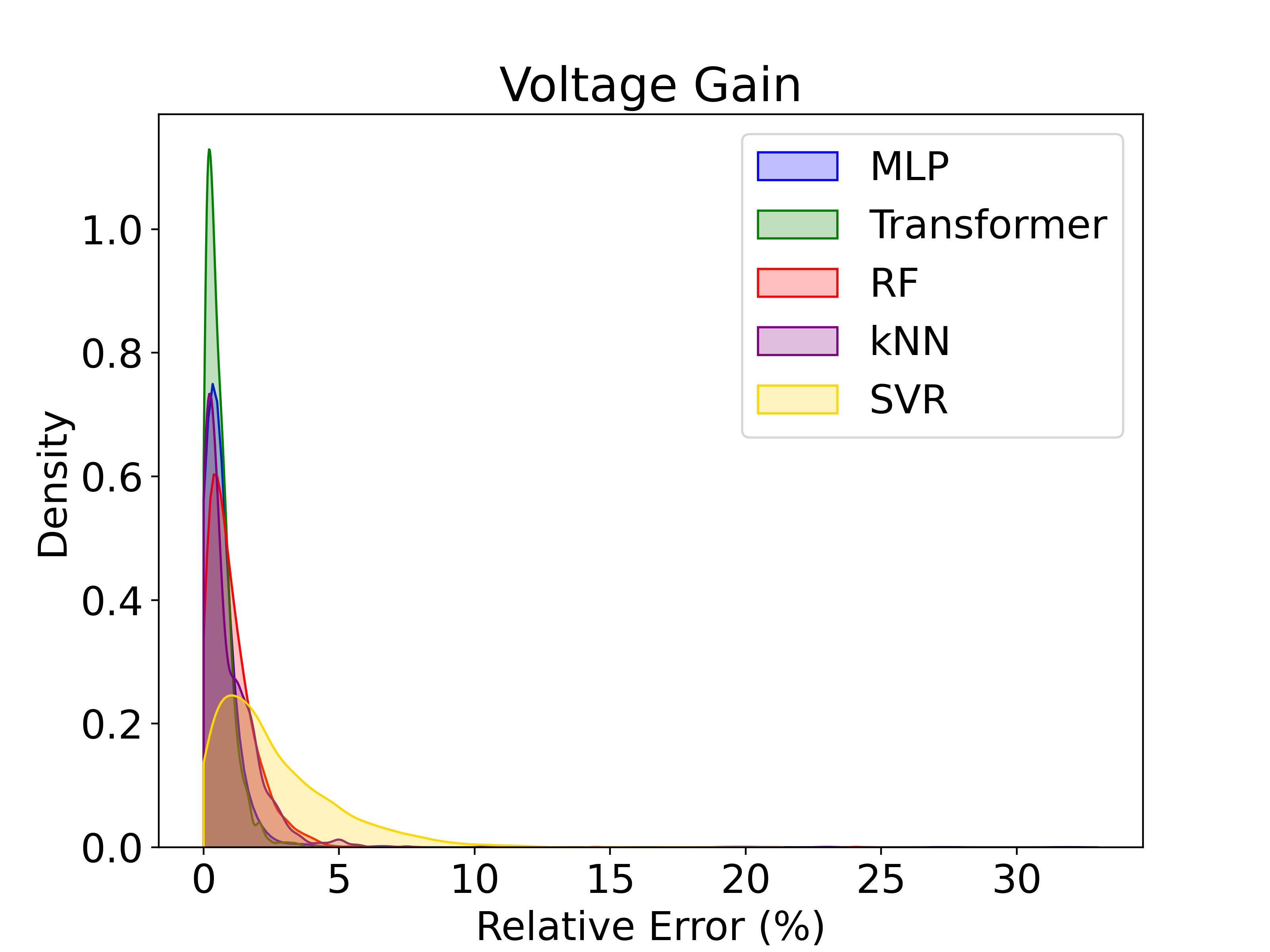}}
    \subfloat{\includegraphics[width=0.2\textwidth]{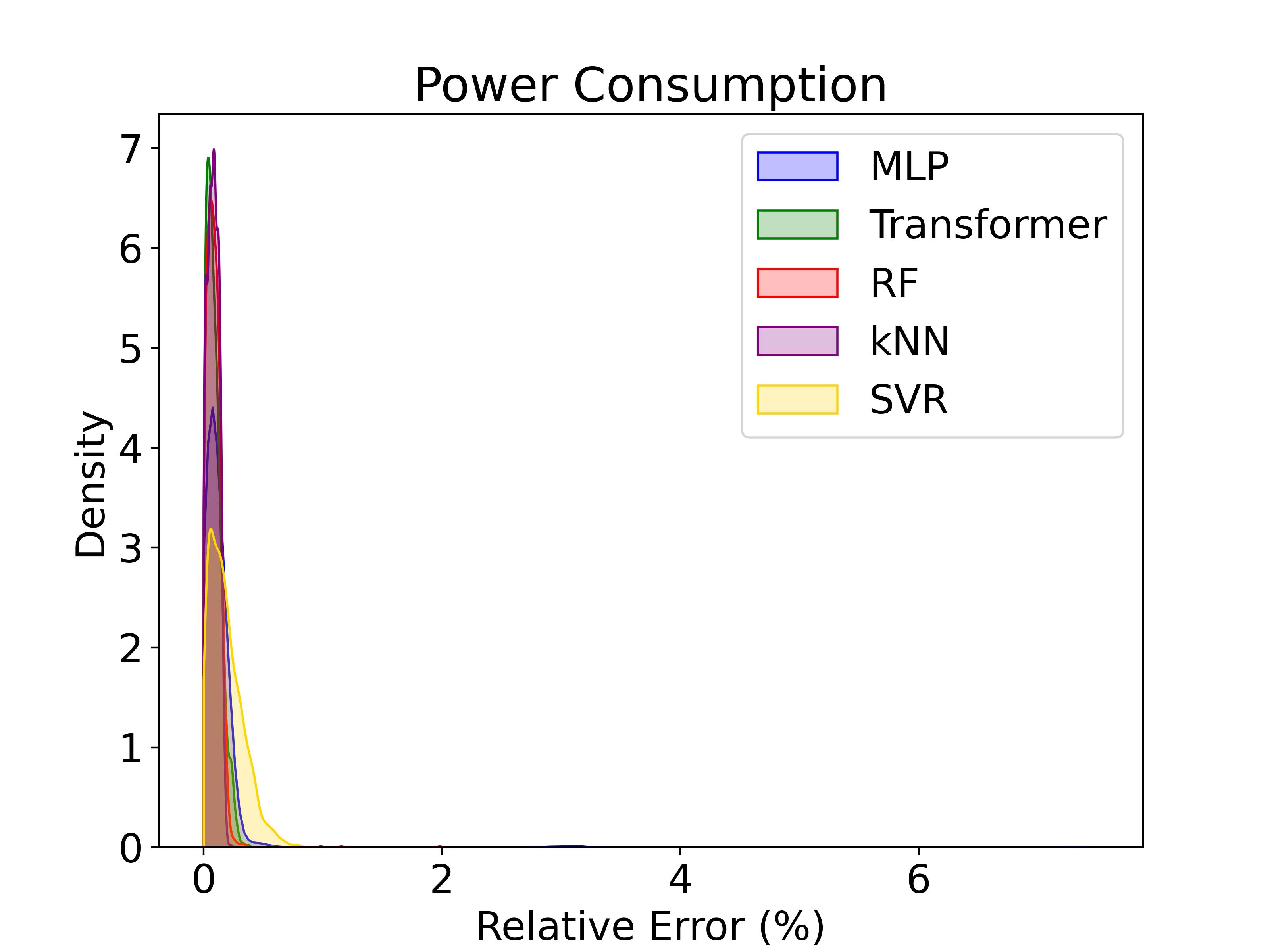}} 
    \subfloat{\includegraphics[width=0.2\textwidth]{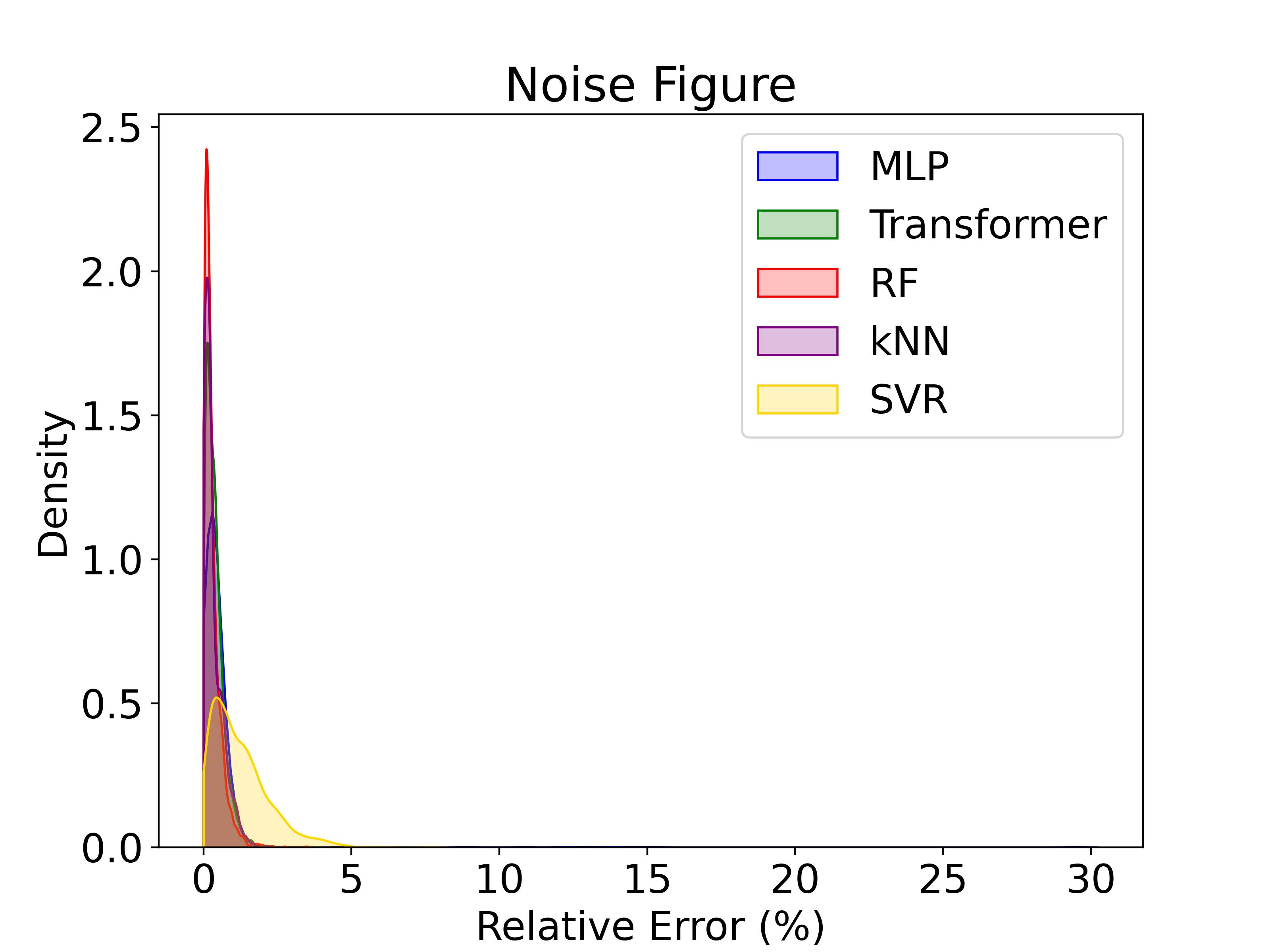}} 
    \subfloat{\includegraphics[width=0.2\textwidth]{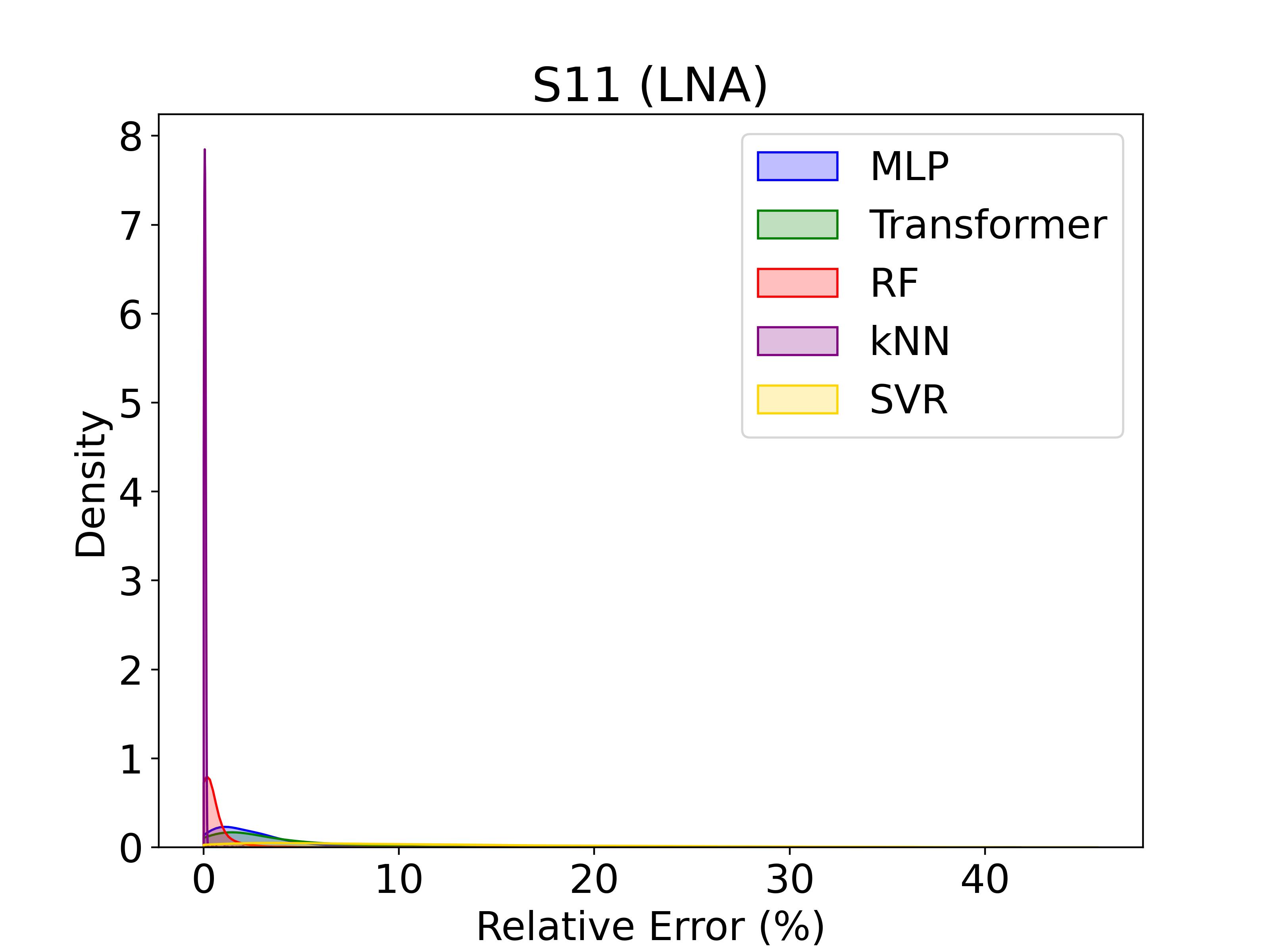}}
    \subfloat{\includegraphics[width=0.2\textwidth]{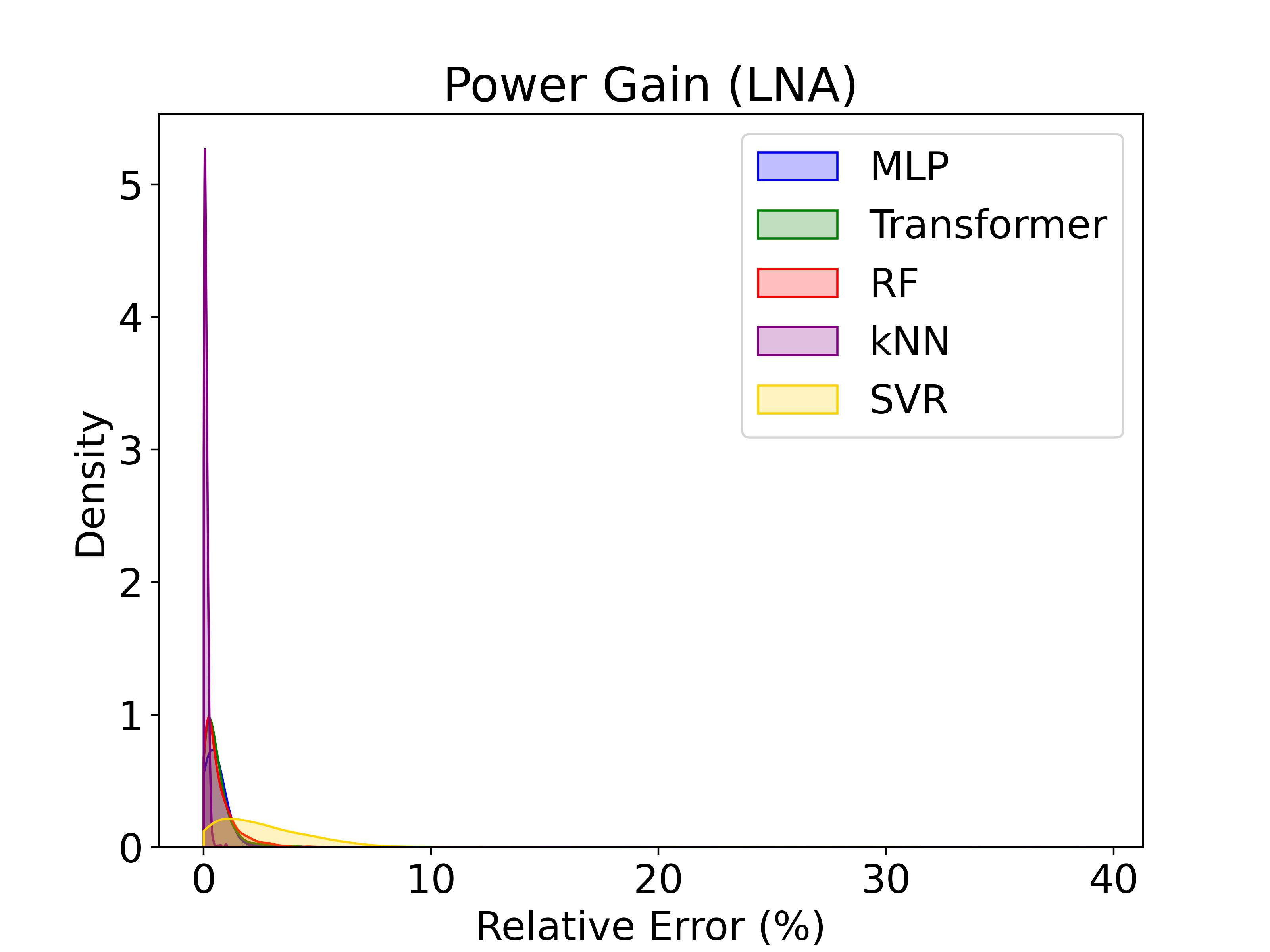}} \\ [-1.5ex]
    \subfloat{\includegraphics[width=0.2\textwidth]{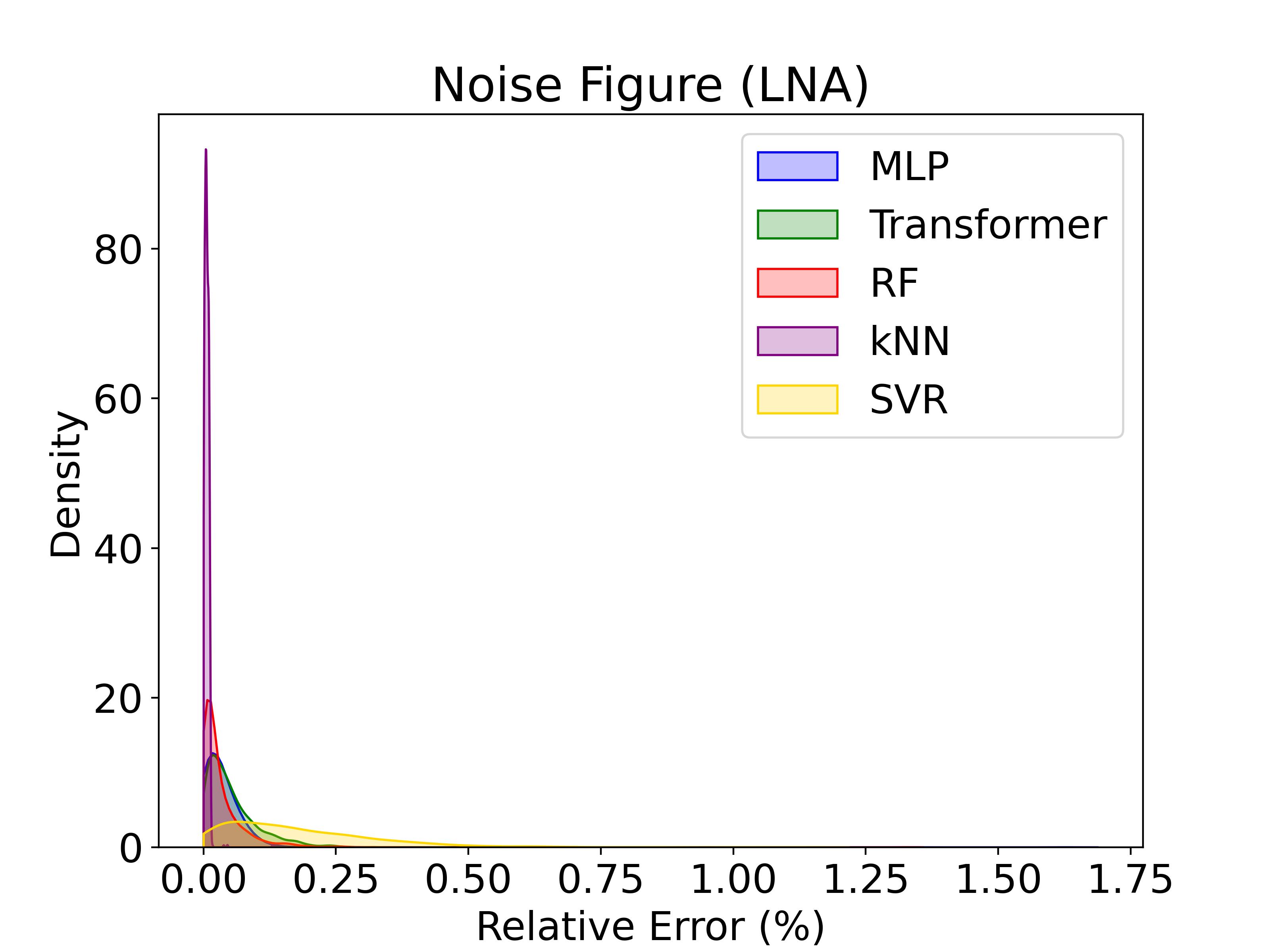}} 
    \subfloat{\includegraphics[width=0.2\textwidth]{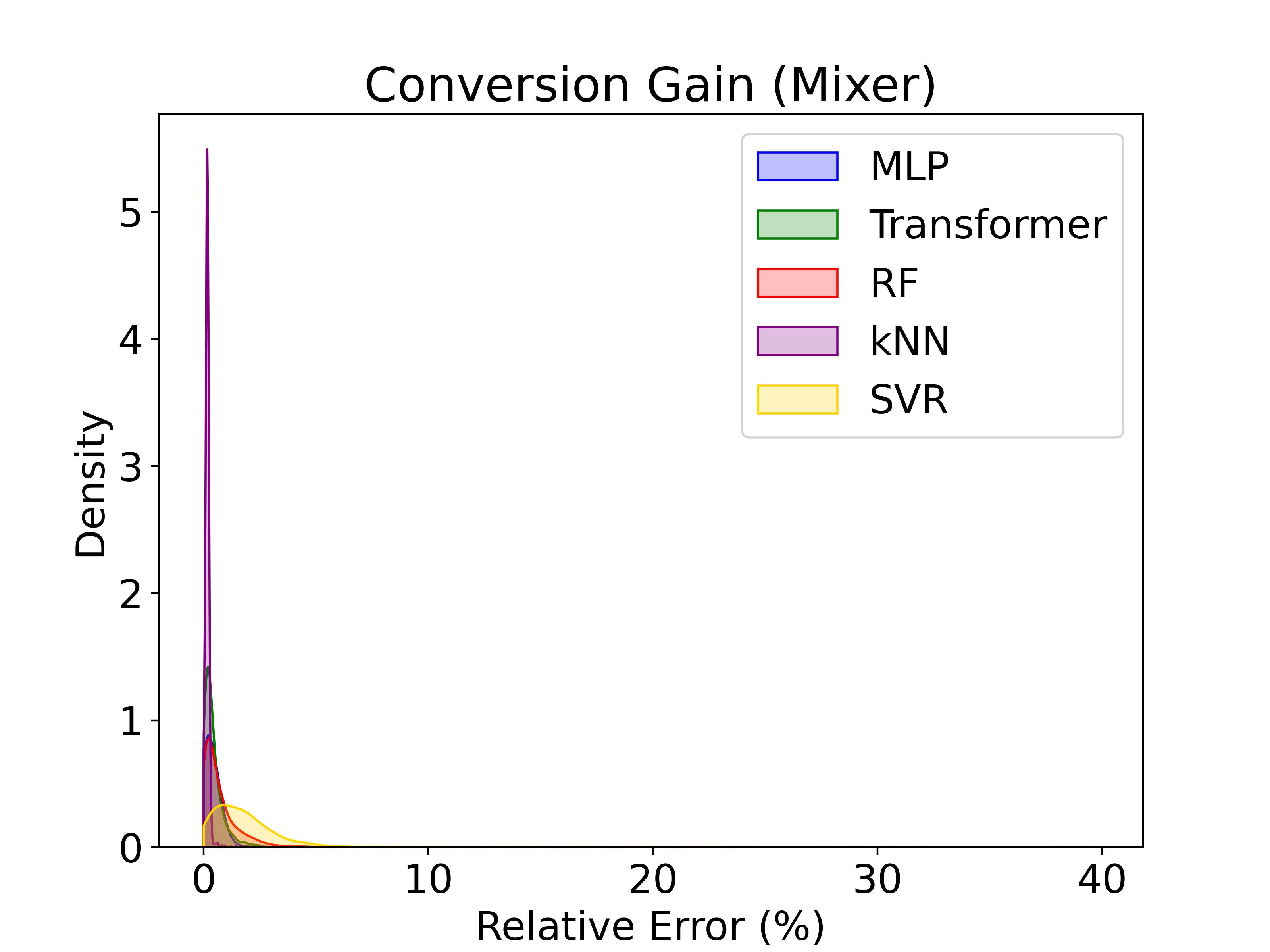}} 
    \subfloat{\includegraphics[width=0.2\textwidth]{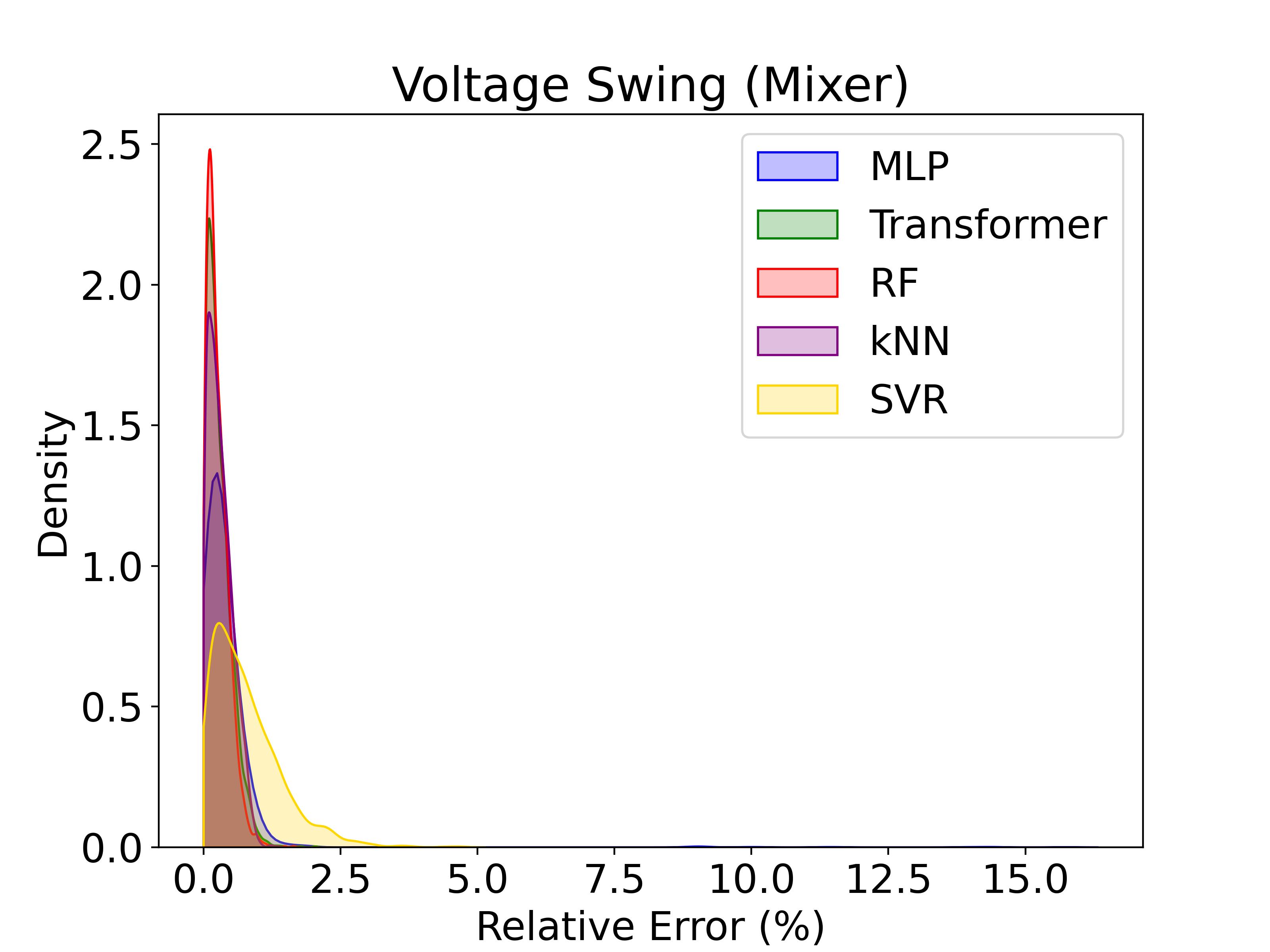}}
    \subfloat{\includegraphics[width=0.2\textwidth]{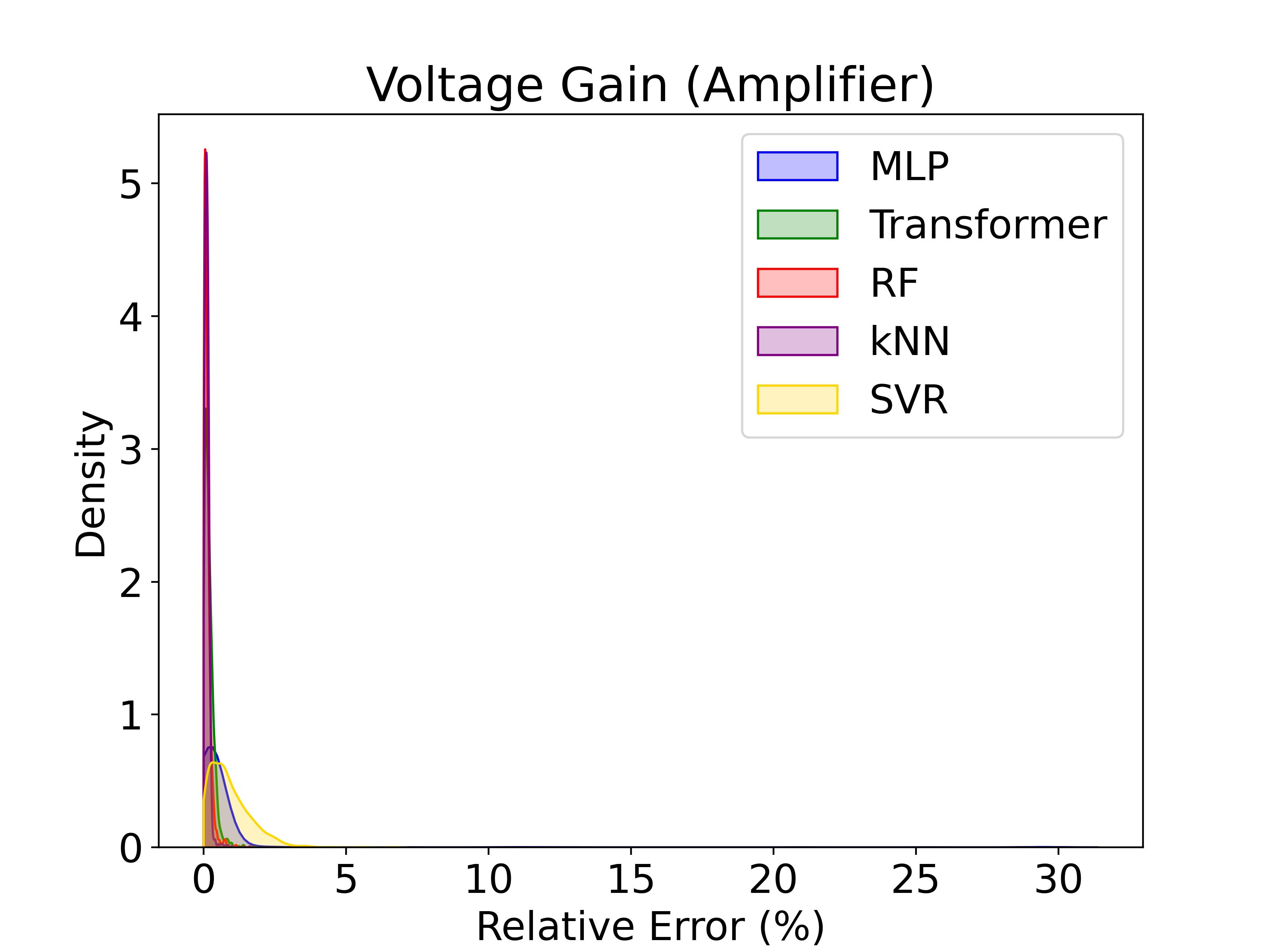}} 
    \subfloat{\includegraphics[width=0.2\textwidth]{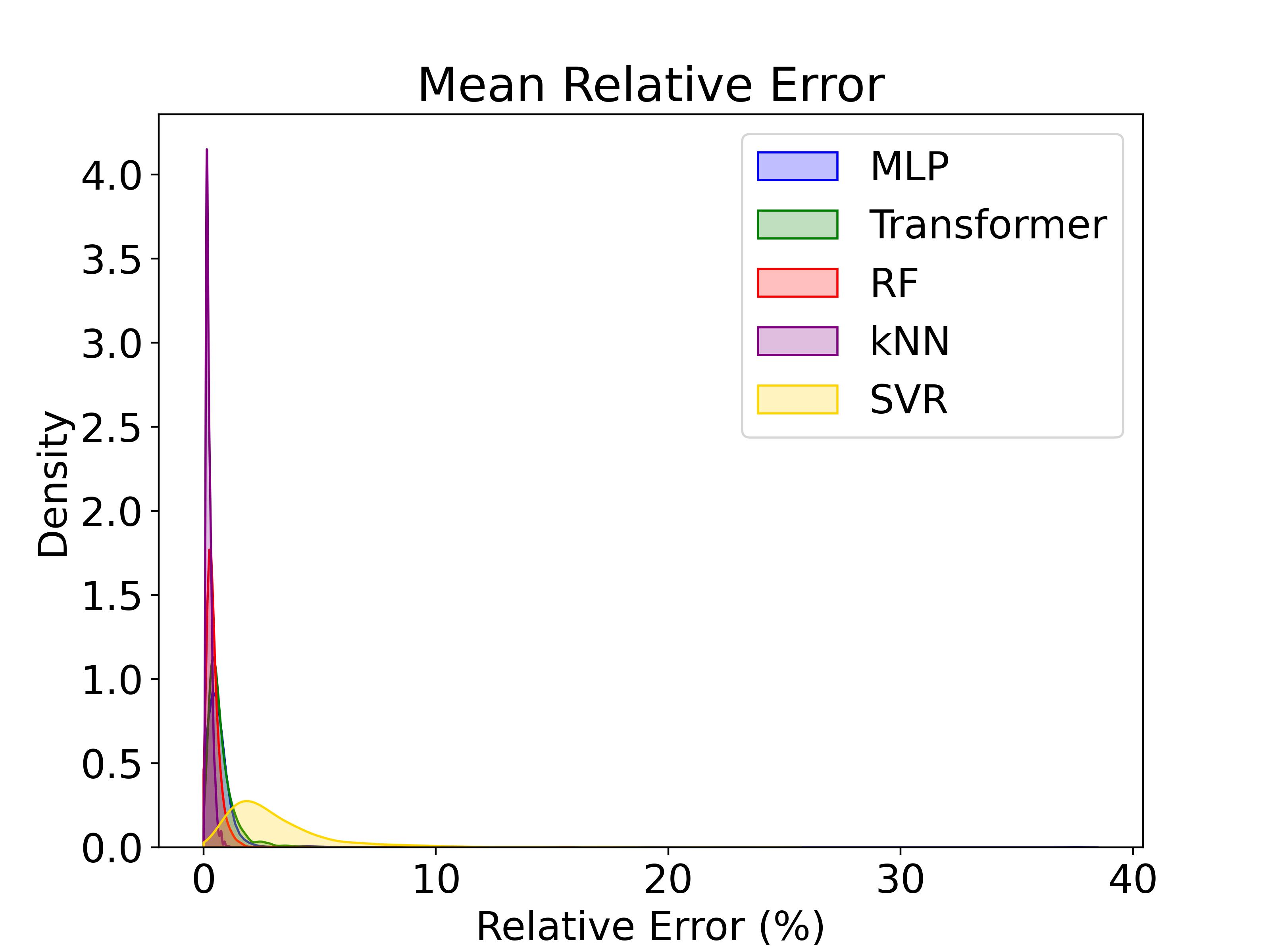}}
    \caption{\centering Relative error histogram of individual performance metrics and mean relative error histogram for Receiver. The plots depict the relative error distributions for system-level performance metrics, such as power consumption, along with block-specific metrics like noise figure for the LNA.}
\label{fig:receiver_plots}
\end{figure*}

\begin{table}[ht!]
\vspace{-2mm}
\caption{Statistical summary of mean relative error for Transmitter}
\label{tab:transmitter_summary}
\centering
\renewcommand{\arraystretch}{0.8} 
\resizebox{0.48\textwidth}{!}{
\begin{tabular}{cccccccc}
\toprule
\raisebox{0.5em}{\textbf{Model}} & 
\raisebox{0.5em}{\textbf{Mean}} & 
\raisebox{0.5em}{\textbf{Std}} & 
\raisebox{0.5em}{\textbf{P75}} & 
\raisebox{0.5em}{\textbf{P90}} & 
\textbf{\shortstack{\% Errors \\ $<$ 2\% }} & 
\textbf{\shortstack{\% Errors \\ $<$ 5\% }} & 
\textbf{\shortstack{\% Outlier \\ ($>$ 20\%)}} \\

\midrule
\textbf{MLP} & \textbf{4.02} & \textbf{38.51} & \textbf{2.51} & 5.36 & \textbf{67.80} & 88.65 & 2.10 \\

\midrule
\textbf{Transformer} & 9.52 & 102.83 & 4.98 & 8.91 & 42.97 & 75.09 & 3.15 \\

\midrule
\textbf{RF} & 9.60 & 97.08 & 5.11 & 11.44 & 41.90 & 74.50 & 4.60 \\

\midrule
\textbf{kNN} & 4.73 & 62.08 & 2.76 & \textbf{4.67} & 60.45 & \textbf{91.20} & \textbf{1.85} \\
        
\midrule
\textbf{SVR} & 38.28 & 240.91 & 26.25 & 56.94 & 3.25 & 22.32 & 32.68 \\
\bottomrule
\end{tabular}
}
\end{table}

\textbf{Receiver.}
The Receiver circuit achieves excellent prediction accuracy across all models, as shown by the narrow distributions in the relative error histograms (Figure~\ref{fig:receiver_plots}) and the statistical summary in Table~\ref{tab:receiver_summary}. The \textbf{kNN} model demonstrates the best overall performance, with the lowest mean error, smallest standard deviation, and the highest percentage of predictions under 2\% and 5\% errors. Furthermore, \textbf{kNN} produces no outlier predictions. \textbf{RF} also performs exceptionally well, closely matching \textbf{kNN}, with similarly small mean and P75 errors. \textbf{MLP} and \textbf{Transformer} models exhibit competitive results but with slightly larger error spreads, indicating less consistency. \textbf{SVR}, on the other hand, lags behind, showing higher mean errors and more significant relative errors. Overall, the receiver system results highlight the ability of \textbf{kNN} and \textbf{RF} to deliver highly reliable predictions with minimal variability.

\begin{table}[!ht]
\vspace{-2mm}
\caption{Statistical summary of mean relative error for Receiver}
\label{tab:receiver_summary}
\centering
\renewcommand{\arraystretch}{0.8} 
\resizebox{0.48\textwidth}{!}{
\begin{tabular}{cccccccc}
\toprule
\raisebox{0.5em}{\textbf{Model}} & 
\raisebox{0.5em}{\textbf{Mean}} & 
\raisebox{0.5em}{\textbf{Std}} & 
\raisebox{0.5em}{\textbf{P75}} & 
\raisebox{0.5em}{\textbf{P90}} & 
\textbf{\shortstack{\% Errors \\ $<$ 2\% }} & 
\textbf{\shortstack{\% Errors \\ $<$ 5\% }} & 
\textbf{\shortstack{\% Outlier \\ ($>$ 20\%)}} \\

\midrule
\textbf{MLP} & 0.75 & 2.15 & 0.68 & 1.03 & 97.35 & 98.90 & 0.15 \\

\midrule
\textbf{Transformer} & 0.75 & 0.71 & 0.91 & 1.41 & 95.50 & 99.65 & \textbf{0.00} \\

\midrule
\textbf{RF} & 0.44 & 0.54 & 0.52 & 0.77 & 99.35 & 99.85 & \textbf{0.00} \\

\midrule
\textbf{kNN} & \textbf{0.23} & \textbf{0.14} & \textbf{0.30} & \textbf{0.40} & \textbf{100.00} & \textbf{100.00} & \textbf{0.00} \\
        
\midrule
\textbf{SVR} & 3.08 & 2.32 & 3.79 & 5.73 & 37.50 & 86.30 & 0.10 \\
\bottomrule
\end{tabular}
}
\end{table}

\section{Discussion}\label{sec:discussion}

The performance of ML models in predicting circuit parameters reflects a balance between circuit complexity and model capability. Simpler circuits with linear parameter-performance relationships tend to yield lower errors, whereas complex circuits with non-linear interactions and intricate trade-offs present greater challenges. Aligning model selection with the unique characteristics of each circuit and understanding their structural differences are crucial for achieving robust and reliable design automation in analog and RF circuits.

\begin{table*}[!t]
\centering
\vspace{-5mm}
\caption{Summary of Best Models and The Statistics of Mean Relative Error for Each Circuit}
\label{tab:best_model_performance}
\resizebox{0.8\textwidth}{!}{
\begin{tabular}{ccccccccc}
\toprule
\textbf{Circuit} & \textbf{Best Model} & \textbf{Mean} & \textbf{Std} & \textbf{P75} & \textbf{P90} & \textbf{\% Errors $<$ 2\%} & \textbf{\% Errors $<$ 5\%} & \textbf{\% Outlier $>$ 20\%} \\
\midrule
\textbf{CSVA} & Transformer & 5.00 & 6.13 & 5.84 & 9.58 & 26.4 & 67.4 & 2.4  \\   
\midrule
\textbf{CVA} & Transformer & 2.34 & 3.74 & 2.08 & 3.29 & 72.6 & 93.4 & 1.0 \\
\midrule
\textbf{TSVA} & RF & 3.30 & 10.74 & 1.60 & 4.66 & 80.0 & 90.6 & 4.6 \\
\midrule
\textbf{LNA} & MLP & 0.30 & 0.21 & 0.38 & 0.54 & 100.0 & 100.0 & 0.0 \\
\midrule
\textbf{Mixer} & Transformer & 3.22 & 2.53 & 4.04 & 8.25 & 37.0 & 81.2 & 0.0  \\
\midrule
\textbf{VCO} & RF & 6.95 & 52.06 & 3.43 & 5.24 & 30.6 & 89.4 & 2.0 \\  
\midrule
\textbf{PA} & MLP & 19.98 & 144.99 & 10.08 & 19.33 & 17.2 & 51.4 & 9.2 \\     
\midrule
\textbf{Transmitter} & MLP & 4.02 & 38.51 & 2.51 & 5.36 & 67.8 & 88.65 & 2.1 \\   
\midrule
\textbf{Receiver} & kNN & 0.23 & 0.14 & 0.30 & 0.40 & 100.0 & 100.0 & 0.0 \\
\bottomrule
\end{tabular}
}
\end{table*}

\begin{table*}[!htbp]
\centering
\caption{Comparison of Scenarios and Models}
\label{tab:table_COMP}
\renewcommand{\arraystretch}{1.1}
\resizebox{0.7\textwidth}{!}{
\begin{tabular}{l|ccc}
\toprule
\textbf{Scenario} & \textbf{Complexity} & \textbf{Data} & \textbf{Models} \\ 
\midrule
Homogeneous & Single $\mathcal{M}:\mathbb{R}^{N}\to\mathbb{R}^{D}$ & Fewer samples & MLP, Transformer (best), RF
\\
Heterogeneous & Coupled $\mathcal{M}_i$'s; Larger $N$ & More samples & MLP, kNN (best) \\ 
\bottomrule
\end{tabular}
}
\vspace{-2mm}
\end{table*}

The performance trends summarized in Table~\ref{tab:best_model_performance} highlight the relationship between circuit complexity and model accuracy. Among \textbf{homogeneous circuits}, more linear designs like the LNA exhibit the lowest mean relative errors, attributed to their straightforward physical structures and linear parameter-performance relationships. For example, the LNA's linear amplification characteristics enable models like MLP to accurately map performance metrics to parameters, achieving near-perfect alignment with training goals. 

Conversely, complex circuits like the PA and VCO exhibit higher mean relative errors due to their intricate trade-offs and non-linear behaviors. The PA's challenges arise from its complexity, with MLP emerging as the best model for its ability to capture non-linear interactions, albeit with higher variability indicated by a large standard deviation. For the VCO, the RF model excels by handling non-linear relationships, which proves effective despite the circuit's expansive design space and inherent randomness in metrics like phase noise.


These results highlight the importance of aligning model capabilities with the physical and functional characteristics of each circuit. The best-performing model is determined not only by statistical accuracy but also by the design complexities and non-linearities inherent in the circuit's structure.

For \textbf{heterogeneous circuits}, the complexity increases due to a larger number of parameters to predict and a corresponding rise in the number of performance metrics. This expansion results in a larger design space. To address this challenge and maintain high prediction accuracy, we increase the number of data points during dataset generation. The Receiver benefits from kNN’s ability to efficiently handle its moderately linear parameter space, made possible by the abundance of samples distributed across different regions of the space. Similarly, the Transmitter is best modeled by both MLP and kNN. The MLP excels in managing multi-block complexities, while kNN effectively captures the slight linearity introduced by the extensive sampling of the parameter space.

\begin{figure}[ht]
    \centering
    \vspace{-4mm}
    \subfloat[Transmitter]{\includegraphics[width=0.24\textwidth]{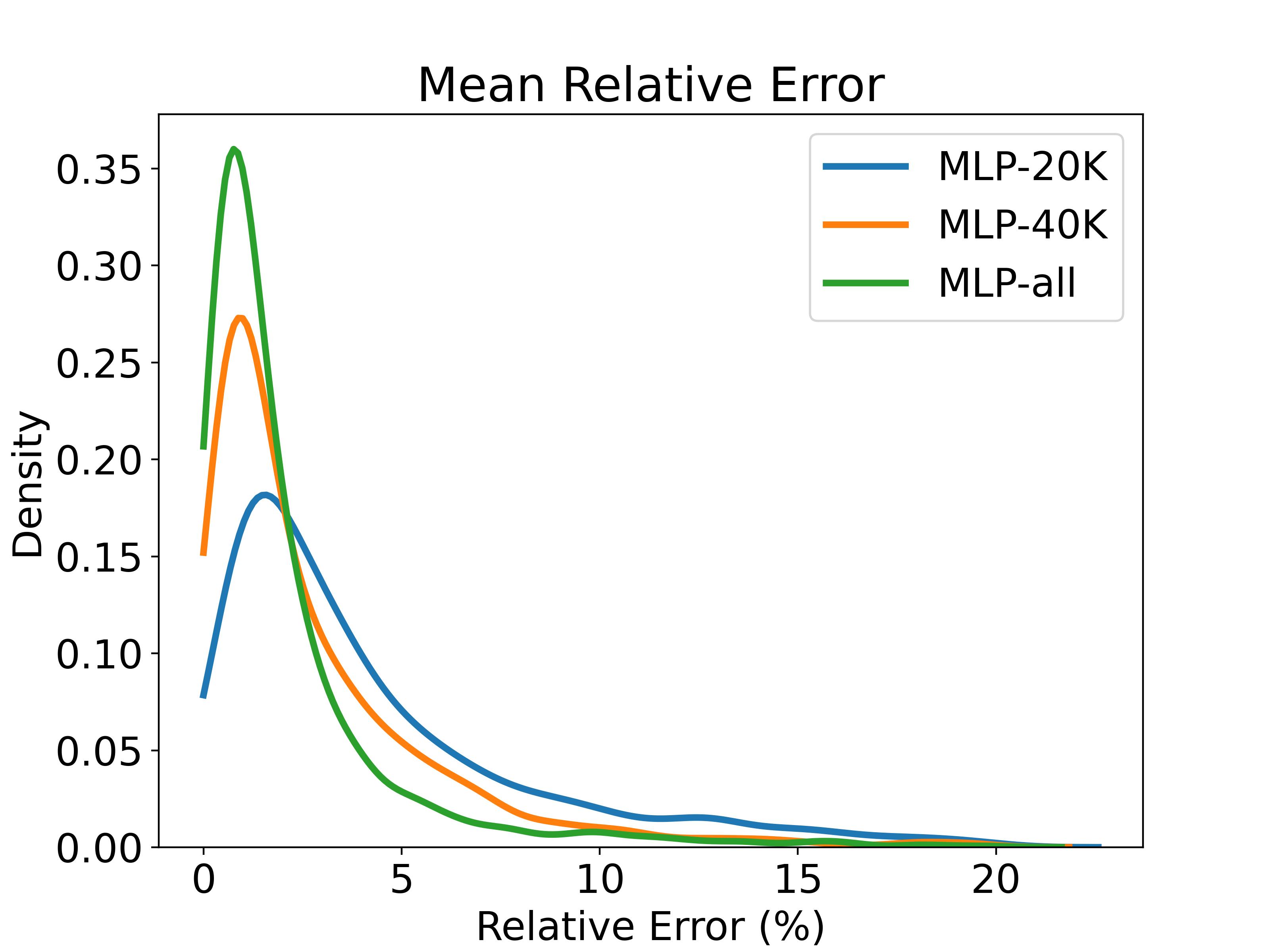}}
    \subfloat[Receiver]{\includegraphics[width=0.24\textwidth]{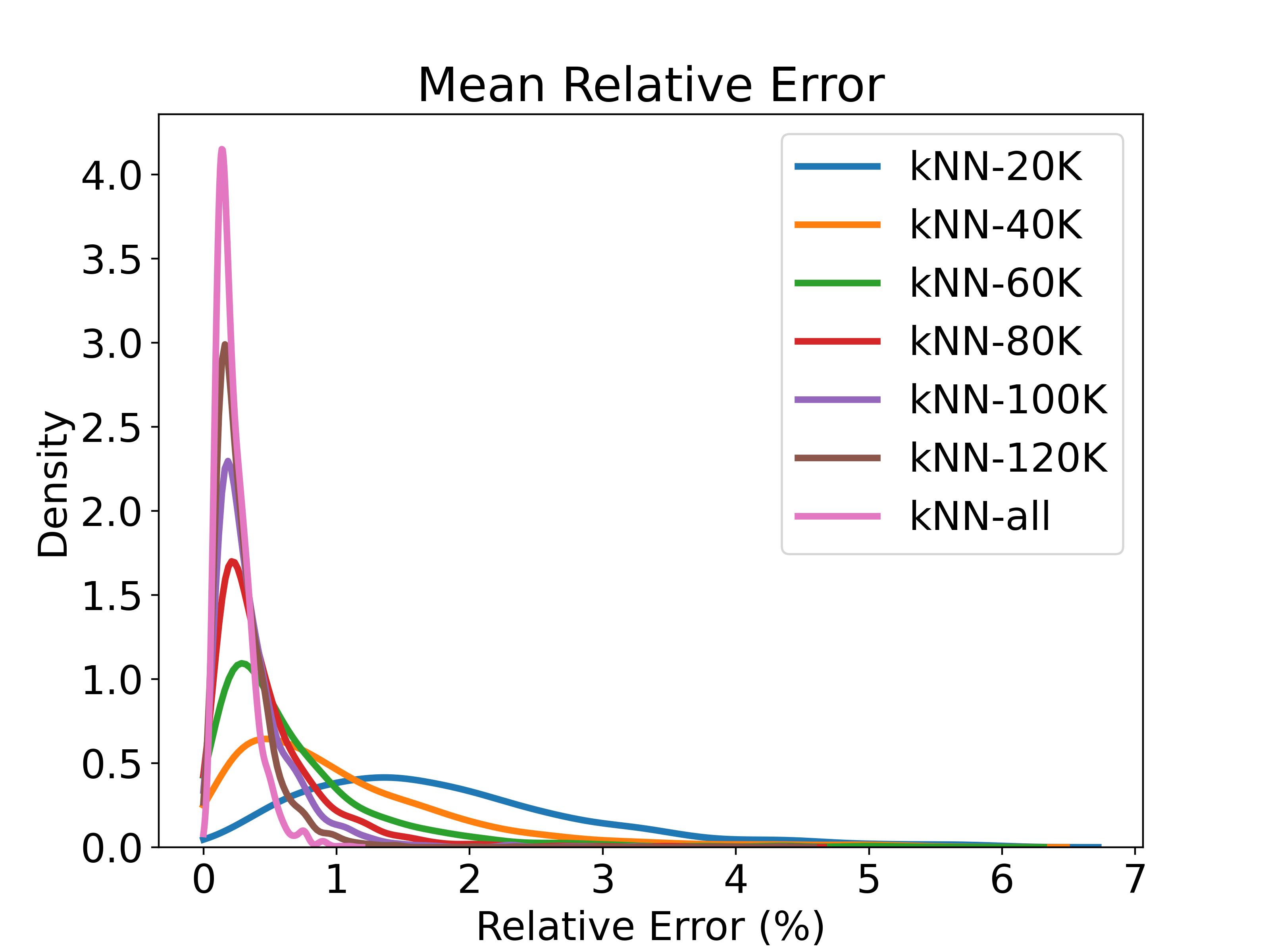}}
    \caption{\centering Mean relative error histograms for (a) Transmitter using the MLP model and (b) Receiver using the kNN model, evaluated across different training dataset sizes. The plots illustrate the scalability of both models, showing that increasing the training data leads to improved prediction accuracy for their respective circuits.}
    \label{fig:scalibility}
\end{figure}

To demonstrate the scalability of our approach, Figure~\ref{fig:scalibility} illustrates the mean relative error for the Transmitter and Receiver as the size of the training dataset increases. As shown, accuracy improves with more data points, emphasizing the scalability of our methodology. Importantly, the dataset was carefully designed to ensure an adequate number of data points, achieving accuracy levels that are satisfactory for practical purposes.

Building on the discussion of scalability, the analytical formulation provides insight into the structural differences between homogeneous and heterogeneous circuits, shedding light on their respective data and modeling requirements.

A \textbf{homogeneous circuit} implements a single non-linear mapping:
\begin{equation}
\mathcal{M}:\mathbb{R}^{N}\to\mathbb{R}^{D}, \quad x \mapsto y,
\end{equation}
where \(x\) and \(y\) represent input and output vectors, respectively. Under supervised learning, given samples \((x,y)\sim P(x,y)\), the training objective follows (\ref{eq:supervised_obj}). Due to the single-stage mapping and moderate dimensionality, fewer samples are typically required for accurate predictions.
  
In contrast, a \textbf{heterogeneous circuit} consists of multiple coupled mappings, for example, three sub-blocks with mappings \(\mathcal{M}_1\), \(\mathcal{M}_2\), \(\mathcal{M}_3\), each defined as:
\begin{equation}
\mathcal{M}_i:\mathbb{R}^{n_i}\to\mathbb{R}^{d_i}, \quad i=1,2,3.
\end{equation}
These mappings combine into a total mapping:
\begin{equation}
\mathcal{M}_{\text{total}}(X) = f(\mathcal{M}_1(x_1), \mathcal{M}_2(x_2), \mathcal{M}_3(x_3)),
\end{equation}
where \(X=[x_1; x_2; x_3]\), with overall input dimensionality \(N=\sum n_i\) and output dimensionality \(D=\sum d_i\). Although the supervised learning objective remains as (\ref{eq:supervised_obj}), the increased complexity, non-linear interactions, and higher dimensionality \(D\) require larger datasets and more expressive models to achieve reliable predictions.


Table \ref{tab:table_COMP} summarizes the model-centric comparison for homogeneous and heterogeneous circuits. For \textbf{homogeneous circuits}, the single-stage mapping \(\mathcal{M}\) and moderate input dimensionality \(N\) enable effective training with fewer data samples. In contrast, for \textbf{heterogeneous circuits}, the multi-stage structure, higher dimensionality \(N\), and intricate interdependencies necessitate larger datasets or more expressive non-linear models to accurately capture the complex interactions.

\section{Conclusion}\label{sec:conclusion}

This work introduces an ML-based framework for predicting circuit parameters from performance specifications, addressing both homogeneous and heterogeneous analog and RF circuits. The study highlights the impact of model selection on prediction accuracy and robustness, with Transformers and MLPs demonstrating strong performance in capturing complex relationships. Moreover, evaluation metrics, such as mean relative error, effectively measured prediction accuracy, while results emphasized the importance of aligning models with circuit complexities and operational characteristics.

Future directions include extending the framework to incorporate \textit{layout optimization}, integrating \textit{uncertainty quantification} for robust predictions, and exploring \textit{generative models} to generate diverse parameter sets that meet performance specifications. This research lays the groundwork for advancing ML-driven circuit design, streamlining workflows, and addressing the growing complexities of modern electronic systems.

\bibliographystyle{IEEEtran}
\bibliography{references}

\begin{thebibliography}{10}
\providecommand{\url}[1]{#1}
\csname url@samestyle\endcsname
\providecommand{\newblock}{\relax}
\providecommand{\bibinfo}[2]{#2}
\providecommand{\BIBentrySTDinterwordspacing}{\spaceskip=0pt\relax}
\providecommand{\BIBentryALTinterwordstretchfactor}{4}
\providecommand{\BIBentryALTinterwordspacing}{\spaceskip=\fontdimen2\font plus
\BIBentryALTinterwordstretchfactor\fontdimen3\font minus \fontdimen4\font\relax}
\providecommand{\BIBforeignlanguage}[2]{{%
\expandafter\ifx\csname l@#1\endcsname\relax
\typeout{** WARNING: IEEEtran.bst: No hyphenation pattern has been}%
\typeout{** loaded for the language `#1'. Using the pattern for}%
\typeout{** the default language instead.}%
\else
\language=\csname l@#1\endcsname
\fi
#2}}
\providecommand{\BIBdecl}{\relax}
\BIBdecl

\bibitem{Hong2021Role}
W.~Hong, Z.~H. Jiang, C.~Yu, D.~Hou, H.~Wang, C.~Guo, Y.~Hu, L.~Kuai, Y.~Yu, Z.~Jiang, Z.~Chen, J.~Chen, Z.~Yu, J.~Zhai, N.~Zhang, L.~Tian, F.~Wu, G.~Yang, Z.-C. Hao, and J.~Y. Zhou, ``The role of millimeter-wave technologies in 5g/6g wireless communications,'' \emph{IEEE Journal of Microwaves}, vol.~1, no.~1, pp. 101--122, 2021.

\bibitem{Dokhanchi2019AutomotiveRadar}
S.~H. Dokhanchi, B.~S. Mysore, K.~V. Mishra, and B.~Ottersten, ``A mmwave automotive joint radar-communications system,'' \emph{IEEE Transactions on Aerospace and Electronic Systems}, vol.~55, no.~3, pp. 1241--1260, 2019.

\bibitem{SubsurfaceImaging}
M.~Alesheikh, R.~Feghhi, F.~M. Sabzevari, A.~Karimov, M.~Hossain, and K.~Rambabu, ``Design of a high-power gaussian pulse transmitter for sensing and imaging of buried objects,'' \emph{IEEE Sensors Journal}, vol.~22, no.~1, pp. 279--287, 2022.

\bibitem{IoTDevices}
P.~Tseng, W.~Yang, M.~Wu, L.~Jin, D.~Li, E.~Low, C.~Hsiao, H.~Lin, K.~Yang, S.~Shen, C.~Kuo, C.~Heng, and G.~Dehng, ``A 55nm saw-less nb-iot cmos transceiver in an rf-soc with phase coherent rx and polar modulation tx,'' in \emph{2019 IEEE Radio Frequency Integrated Circuits Symposium (RFIC)}, 2019, pp. 267--270.

\bibitem{Quantum}
Y.~Guo, Q.~Liu, T.~Li, N.~Deng, Z.~Wang, H.~Jiang, and Y.~Zheng, ``Cryogenic cmos rf circuits: A promising approach for large-scale quantum computing,'' \emph{IEEE Transactions on Circuits and Systems II: Express Briefs}, vol.~71, no.~3, pp. 1619--1625, 2024.

\bibitem{THzSpectroscopy}
R.~Han and E.~Afshari, ``A cmos high-power broadband 260-ghz radiator array for spectroscopy,'' \emph{IEEE Journal of Solid-State Circuits}, vol.~48, no.~12, pp. 3090--3104, 2013.

\bibitem{IEEE97_Moore}
R.~R. Schaller, ``Moore's law: past, present and future,'' \emph{IEEE spectrum}, vol.~34, no.~6, pp. 52--59, 1997.

\bibitem{10031431}
H.~Bong, K.~Cho, and Y.~Seo, ``Automation framework for digital circuit design and verification,'' in \emph{2022 19th International SoC Design Conference (ISOCC)}, 2022, pp. 263--264.

\bibitem{869353}
V.~Vassilev, D.~Job, and J.~Miller, ``Towards the automatic design of more efficient digital circuits,'' in \emph{Proceedings. The Second NASA/DoD Workshop on Evolvable Hardware}, 2000, pp. 151--160.

\bibitem{elfadel2019machine}
I.~M. Elfadel, D.~S. Boning, and X.~Li, \emph{Machine learning in VLSI computer-aided design}.\hskip 1em plus 0.5em minus 0.4em\relax Springer, 2019.

\bibitem{gielen2000computer}
G.~Gielen and R.~Rutenbar, ``Computer-aided design of analog and mixed-signal integrated circuits,'' \emph{Proceedings of the IEEE}, vol.~88, no.~12, pp. 1825--1854, 2000.

\bibitem{Els21_Analog}
P.~E. Allen and D.~R. Holberg, \emph{CMOS analog circuit design}.\hskip 1em plus 0.5em minus 0.4em\relax Elsevier, 2011.

\bibitem{razavi_RFIC}
B.~Razavi, \emph{RF Microelectronics}, 2nd~ed.\hskip 1em plus 0.5em minus 0.4em\relax Prentice Hall Press, 2011.

\bibitem{ICCAD19_BagNet}
K.~Hakhamaneshi, N.~Werblun, P.~Abbeel, and V.~Stojanovi{\'c}, ``Bagnet: Berkeley analog generator with layout optimizer boosted with deep neural networks,'' in \emph{2019 IEEE/ACM International Conference on Computer-Aided Design (ICCAD)}.\hskip 1em plus 0.5em minus 0.4em\relax IEEE, 2019, pp. 1--8.

\bibitem{DATE_AutoCkt}
K.~Settaluri, A.~Haj-Ali, Q.~Huang, K.~Hakhamaneshi, and B.~Nikolic, ``Autockt: deep reinforcement learning of analog circuit designs,'' in \emph{Proceedings of the 23rd Conference on Design, Automation and Test in Europe}, ser. DATE '20.\hskip 1em plus 0.5em minus 0.4em\relax San Jose, CA, USA: EDA Consortium, 2020, p. 490–495.

\bibitem{IEEE_Angel}
M.~Fayazi, M.~T. Taba, E.~Afshari, and R.~Dreslinski, ``Angel: Fully-automated analog circuit generator using a neural network assisted semi-supervised learning approach,'' \emph{IEEE Transactions on Circuits and Systems I: Regular Papers}, 2023.

\bibitem{DAC_GCNRL}
H.~Wang, K.~Wang, J.~Yang, L.~Shen, N.~Sun, H.-S. Lee, and S.~Han, ``Gcn-rl circuit designer: Transferable transistor sizing with graph neural networks and reinforcement learning,'' in \emph{2020 57th ACM/IEEE Design Automation Conference (DAC)}.\hskip 1em plus 0.5em minus 0.4em\relax IEEE, 2020, pp. 1--6.

\bibitem{ICML_Analog}
D.~Krylov, P.~Khajeh, J.~Ouyang, T.~Reeves, T.~Liu, H.~Ajmal, H.~Aghasi, and R.~Fox, ``Learning to design analog circuits to meet threshold specifications,'' in \emph{Proceedings of the 40th International Conference on Machine Learning}, ser. ICML'23.\hskip 1em plus 0.5em minus 0.4em\relax JMLR.org, 2023.

\bibitem{Mehradfar2024AICircuit}
A.~Mehradfar, X.~Zhao, Y.~Niu, S.~Babakniya, M.~Alesheikh, H.~Aghasi, and S.~Avestimehr, ``{AICircuit: A Multi-Level Dataset and Benchmark for AI-Driven Analog Integrated Circuit Design},'' \emph{Machine Learning and the Physical Sciences Workshop @ NeurIPS}, 2024.

\bibitem{bayesian2020}
S.~A. Abdelaal, A.~Hussein, and H.~Mostafa, ``A bayesian optimization framework for analog circuits optimization,'' in \emph{2020 15th International Conference on Computer Engineering and Systems (ICCES)}, 2020.

\bibitem{bayesianZhang}
S.~Zhang, F.~Yang, C.~Yan, D.~Zhou, and X.~Zeng, ``An efficient batch-constrained bayesian optimization approach for analog circuit synthesis via multiobjective acquisition ensemble,'' \emph{IEEE Transactions on Computer-Aided Design of Integrated Circuits and Systems}, vol.~41, no.~1, pp. 1--14, 2022.

\bibitem{rlzhao}
Z.~Zhao and L.~Zhang, ``Deep reinforcement learning for analog circuit sizing,'' in \emph{2020 IEEE International Symposium on Circuits and Systems (ISCAS)}, 2020.

\bibitem{razavi_design_2017}
B.~Razavi, \emph{Design of analog {CMOS} integrated circuits}.\hskip 1em plus 0.5em minus 0.4em\relax New York, NY: McGraw-Hill Education, 2017.

\bibitem{wireless_systems}
A.~Abidi, ``Cmos wireless transceivers: the new wave,'' \emph{IEEE Communications Magazine}, vol.~37, no.~8, pp. 119--124, 1999.

\bibitem{Millimiter_wave_circuit}
X.~Liu, M.~H. Maktoomi, M.~Alesheikh, P.~Heydari, and H.~Aghasi, ``A 49-63 ghz phase-locked fmcw radar transceiver for high resolution applications,'' in \emph{ESSCIRC 2023- IEEE 49th European Solid State Circuits Conference (ESSCIRC)}, 2023, pp. 509--512.

\bibitem{li2002multi}
\BIBentryALTinterwordspacing
X.~Li and M.~Ismail, \emph{Multi-Standard CMOS Wireless Receivers: Analysis and Design}.\hskip 1em plus 0.5em minus 0.4em\relax Springer US, 2002. [Online]. Available: \url{https://books.google.com/books?id=OkSi8gSf-4IC}
\BIBentrySTDinterwordspacing

\bibitem{Cadence02}
A.~J.~L. Martin, ``Cadence design environment,'' \emph{New Mexico State University, Tutorial paper}, vol.~35, 2002.

\bibitem{NIPS23_Attention}
A.~Vaswani, N.~Shazeer, and et~al, ``Attention is all you need,'' \emph{Advances in neural information processing systems}, vol.~30, 2017.

\bibitem{arXiv19_Bert}
J.~Devlin, M.-W. Chang, K.~Lee, and K.~Toutanova, ``Bert: Pre-training of deep bidirectional transformers for language understanding,'' 2019.

\bibitem{ML01_RandomForest}
L.~Breiman, ``Random forests,'' \emph{Machine learning}, vol.~45, pp. 5--32, 2001.

\bibitem{kramer2013k}
O.~Kramer, ``K-nearest neighbors,'' \emph{Dimensionality reduction with unsupervised nearest neighbors}, pp. 13--23, 2013.

\bibitem{awad2015support}
M.~Awad, R.~Khanna, M.~Awad, and R.~Khanna, ``Support vector regression,'' \emph{Efficient learning machines: Theories, concepts, and applications for engineers and system designers}, pp. 67--80, 2015.

\bibitem{MIT04_Kernel}
J.-P. Vert, K.~Tsuda, and B.~Sch{\"o}lkopf, ``2 a primer on kernel methods,'' \emph{Kernel Methods in Computational Biology}, p.~35, 2004.

\bibitem{arXiv17_Adam}
D.~P. Kingma and J.~Ba, ``Adam: A method for stochastic optimization,'' 2017.

\end{thebibliography}

\end{document}